\documentclass[a4paper,12pt,twoside]{report} 
\pdfoutput=1
\usepackage[top=3cm,bottom=2cm,left=2cm,right=2cm,twoside,includefoot, heightrounded]{geometry}
\usepackage[final]{pdfpages}
\usepackage{fancyhdr}
\usepackage{subfigure}
\usepackage{pifont}
\usepackage{color}
\usepackage{enumerate}
\usepackage{amssymb,amsmath, mathtools}
\usepackage[mathscr]{eucal}
\usepackage{multicol}
\usepackage[titles,subfigure]{tocloft}
\usepackage{cite}
\usepackage{url}
\usepackage{amsfonts}
\usepackage{amsbsy}
\usepackage{amscd}
\usepackage{morefloats}
\usepackage{multirow}
\usepackage{epigraph}
\usepackage{array}
\usepackage{relsize}
\usepackage{makecell}

\newtheorem{myremark}{Remark}[section]
\newtheorem{mydef}{Definition}[section]
\newtheorem{myassump}{Assumption}[section]
\newtheorem{mytheorem}{Theorem}[section]
\newtheorem{mylemma}{Lemma}[section]
\newlength{\mylength} 

\begin{document}

\pagenumbering{roman}

\begin{titlepage}
\pagestyle{empty}
\begin{center}
\vfill
{\huge {\bf \LARGE Decentralized Autonomous Navigation Strategies for Multi-Robot Search and Rescue\\}}
\vspace{2cm}

by\\
\vspace{1cm}
{\Large {\bf \Large Ahmad Baranzadeh}}\\

\vspace{2cm}

\vspace{2cm}

\vspace{8cm}

2016
\end{center}

\end{titlepage}

\chapter*{Abstract}

Use of multi-robot systems has many advantages over single robot systems in various applications. However, it comes with its own complexity and challenges. In this report, we try to improve the performance of existing approaches for “search” operations in multi-robot context. We propose three novel algorithms that are using a triangular grid pattern, i.e., robots certainly go through the vertices of a triangular grid during the search procedure. The main advantage of using a triangular grid pattern is that it is asymptotically optimal in terms of the minimum number of robots required for the complete coverage of an arbitrary bounded area. Therefore, using the vertices of this triangular grid coverage guarantees complete search of a region as well as shorter searching time. We use a new topological map which is made and shared by robots during the search operation. We consider an area that is unknown to the robots a priori with an arbitrary shape, containing some obstacles. Unlike many current heuristic algorithms, we give mathematically rigorous proofs of convergence with probability 1 of the algorithms. The computer simulation results for the proposed algorithms are presented using a simulator of real robots and environment. We evaluate the performance of the algorithms via experiments with real Pioneer 3DX mobile robots. We compare the performance of our own algorithms with three existing algorithms from other researchers. The results demonstrate the merits of our proposed solution.

A further study on formation building with obstacle avoidance for a team of mobile robots is presented in this report. We propose a robust decentralized formation building with obstacle avoidance algorithm for a group of mobile robots to move in a defined geometric configuration. Furthermore, we consider a more complicated formation problem with a group of anonymous robots; these robots are not aware of their position in the final configuration and need to reach a consensus during the formation process. We propose a randomized algorithm for the anonymous robots that achieves the convergence to a desired configuration with probability 1. We also propose a novel obstacle avoidance rule, used in the formation building algorithm. A mathematically rigorous proof of the proposed algorithm is given. The performance and applicability of the proposed algorithm are confirmed by the computer simulation results.


\tableofcontents


\pagenumbering{arabic}

\chapter{Introduction}\label{chap:introduction}

The past three decades have seen increasingly rapid advances  in robotics. Also, mobile robots area; usually referred to unmanned aerial vehicles (UAVs), unmanned ground vehicles (UGVs), unmanned underwater vehicles, driverless cars, etc., has been a major area of interest within the field of robotics. Using mobile robots has recently become  very popular  in various tasks. Mobile robots are employed to achieve from very simple home tasks like vacuum cleaning \cite{viet2013ba, hess2014probabilistic} to some important and complicated jobs like finding a dangerous odour source in a warehouse \cite{marjovi2011multi}. They have been used to help or replace human in various applications such as unmanned aircraft systems (UAS) \cite{kendoul2012survey, MAM:MAM12046, zhang2012application}, factory automation \cite{dang2014scheduling, guizzo2012rise}, mining \cite{chi2012automatic, dunn2015control, subhan2014study}, home and office assistance \cite{reiser2013care, wang2015intelligent}, interactive guide systems \cite{satake2015field}, search and rescue operations \cite{guarnieri2009helios, kruijff2014experience}, search and exploration in hazardous environments \cite{marjovi2010multi}, and sensor networks \cite{tuna2014autonomous, liu2013dynamic}.

A single-robot system refers to only one individual robot which can model itself, the environment and their interaction \cite{stone2000multiagent}. Use of a single robot in applications mentioned above has been studied by many researchers, and this field of research is well developed. NASA  Mars Pathfinder \cite{nasaMarsPathfinder}, Sony  Aibo, a robotic dog capable of seeing and walking \cite{sonyAibo}, and Boston Dynamics’ BigDog that is able to walk on icy terrain and recover its balance when kicked from the side \cite{BigDog}, are famous examples of mobile robots. Despite advances in  single-robot systems that enable them to overcome many obstacles and achieve a lot of operations, there are still some tasks that are inherently impossible or too complicated to be done by an individual robot. Furthermore, there are also some tasks that might not be timely or economically efficient to be done by a single robot. For instance,  two keys in different parts of an environment must be activated concurrently, cannot be accomplished by a single robot \cite{dudek1996taxonomy}.

\section{Multi-Robot Systems}
On the other side, using a team consisting of some  autonomous mobile robots instead of only a single robot for aforementioned tasks, is a newer approach that has attracted many researchers in recent years \cite{zadorozhny2013information, pereira2015coordination, carvalho2013multi, sharma2015multi}. Though, using multi-robot systems may cause some new challenging problems in contrast to single-robot systems, they have a lot of advantages that  assure the researchers  to have an increased interest in this topic. Furthermore, recent developments in the fields of microelectronics, measurement and  sensors, wireless communication networks, and computer hardware and software,  have led to a renewed interest in multi-robot systems. Compared to single-robot systems, multi-robot systems have several potential advantages which are \cite{yan2013survey, roumeliotis2002distributed, fox2000probabilistic, wawerla2010fast, prorok2012low}:

\begin{itemize}
  \item More reliable; as a result of information sharing among robots, any failure in one robot would not cause a failure of the whole system.
  \item Robust and fault-tolerant;  using multi-robot systems can reduce errors in odometry, communication, sensing and so on.
  \item Scalable; with growing amount of work, the algorithm applied to multi-robot systems still works properly.
  \item Flexible; they can adopt themselves with any changes in the environment.
  \item Lower cost; using some simple robots is cheaper than a single complex and powerful robot.
  \item A better spatial distribution.
  \item A better overall system performance  in terms of  time or the energy consumption required completing a task.
\end{itemize}

Despite aforementioned advantages, multi-robot systems consist of many different and yet related research topics and problems. Communication and coordination of the robots during the operation, localization, and mapping are some of the most significant problems of multi-robot systems which have been  discussed in the next sections.

\section{Multi-Robot Search and Rescue}
In the field of multi-robot systems, search and rescue, exploration, foraging and patrolling are some different terms that are closely related. A considerable amount of literature has been published on these topics \cite{agmon2008multi, chaimowicz2005deploying, wurm2008coordinated, baranzadehdecentralized, nazarzehidecentralized, nazarzehi2015distributedRobio}. Although these topics may seem distinct, their related problems in multi-robot systems are almost similar so that they can be classified in the same category. All of these problems can be done by a single robot, and there are many articles have been published about that. But as outlined before, using a team of robots has many advantages over a single robot that make it powerful so that it can accomplish a search task more efficient in terms of time and cost. Multi-robot exploring systems have numerous applications such as exploring an entire  region for unknown number of targets \cite{baranzadeh2015distributed, baranzadeh2016distributed}, search and rescue operations \cite{sugiyama2009integrated}, surveyance a building to detect potential intruders (e.g., museums and laboratories) \cite{portugal2011survey}, patrolling in outdoor environments (e.g., a country borders) \cite{matveev2011method}, intruder detection and perimeter protection \cite{howard2006experiments, fagiolini2008consensus}.

 This report intends to present some algorithms which can be applied in the search and rescue operations (SAR) \cite{tadokoro2009rescue, murphy2009mobile} as well as other exploring tasks. Clearly, it is significant to save human  lives in a disaster such as an earthquake, a plane crash, a missed boat in the ocean and a fire in an industrial warehouse. In most of such cases, search and rescue operations are usually difficult or even impossible to be performed by humans. Therefore, using a team of robots would be helpful to carry out the operation to increase the survival rate and to decrease the risk of the operation. Besides, in a search and rescue operation, the search area must be completely explored ensuring all possible targets are detected.  Our suggested algorithms thoroughly  cover this goal in the sense that they guarantee exploring all points of the search area. In the following, we mention some of the studies in this field carried out in recent years by other researchers.

Beard and McLain \cite{beard2003multiple} proposed a  dynamic programming approach  for a team of cooperating UAVs to explore a region as much as possible while avoiding regions of hazards. Vincent and Rubin \cite{vincent2004framework} used predefined swarm patterns to address the problem of cooperative search strategies for UAVs searching for moving targets in  hazardous environments. Yang et al. \cite{yang2007multi} used distributed reinforcement learning for multi-agent search wherein the robots learn the environment online and store the information as a map and utilise that to compute their route.

Baxter et al. \cite{baxter2007multi} described two implementations of a potential field sharing multi-robot system; pessimistic and optimistic. They considered that robots perform no reasoning and are purely reactive in nature. They pointed out that potential field sharing has a positive impact on robots involved in
a search and rescue problem. Marjovi et al. \cite{marjovi2009multi} presented a method for fire searching by a multi-robot team. They proposed a decentralized frontier based exploration by which the robots explore an unknown environment to detect fire sources. In \cite{guarnieri2009helios}, Guarnieri et al. presented a search and rescue system named HELIOS team, consisting of five tracked robots for urban search And rescue. Two units are equipped with manipulators for the accomplishment of particular tasks, such as the handling of objects and opening doors; the other three units, equipped with cameras and laser range finders, are utilized to create virtual 3D maps of the explored environment. The three units can move autonomously while collecting the data by using a collaborative positioning system (CPS).

Sugiyama et al. \cite{sugiyama2010autonomous} investigated the autonomous chain network formation by multi-robot rescue systems. According to Sugiyama et al. \cite{sugiyama2010autonomous}, chain networks connecting a base station and rescue robots are essential to reconnoiter distant spaces in disaster areas, and the chains must be formed to assure communications among them and must be transformed if the target of exploration changes. They adopted autonomous classification of robots into search robots and relay robots so that the robots act according to the behavior algorithms of each class of robot to form chain network threading the path to the distant spaces. In \cite{luo2011multi}, Luo et al. proposed an approach that employs a team consisting of  ground and aerial vehicles simultaneously. The ground vehicle is used for the purpose of environment mapping, the micro aerial vehicle is used simultaneously for the purpose of search and localization with a vertical camera and a horizontal camera, and two other micro ground vehicles with sonar, compass and colour sensor used as the back-up team.

Lewis and Sycara \cite{lewis2011network} demonstrated the evolution of an experimental human-multi-robot system for urban search and rescue  in which operators and robots collaborate  to search for victims. Macwan and Nejat \cite{macwan2011target} presented a modular methodology for predicting a lost person's behavior (motion) for autonomous coordinated multi-robot wilderness search and rescue. They asserted introducing  new concept of isoprobability curves, which represents a unique mechanism for identifying the target’s probable location at any given time within the search area while accounting for influences such as terrain topology, target physiology and psychology, clues found, etc. Mobedi and Nejat \cite{mobedi20123} developed an active 3-D sensory system that can be used in robotic rescue missions to map these unknown cluttered urban disaster environments and determine the locations of victims.

Sugiyama et al. \cite{sugiyama2013real} proposed a system procedure for a multi-robot rescue system that performs real-time exploration over disaster areas. The proposed system procedure consists of the autonomous classification of robots into search and relay types, and behavior algorithms for each class of robot. Searching robots explore the areas, and relay robots act as relay terminals between searching robots and the base station. The rule of the classification and the behavior algorithm refer to the forwarding table of each robot constructed for ad hoc networking. The table construction is based on DSDV (destination-sequenced distance vector) routing that informs each robot of its topological position in the network and other essentials. In \cite{liu2013learning},  Liu et al.  proposed a hierarchical reinforcement learning (HRL) based semi-autonomous control architecture for rescue robot teams to enable cooperative learning between the robot team members. They claimed that the HRL-based control architecture allows a multi-robot rescue team to collectively make decisions regarding which rescue tasks need to be carried out at a given time, and which team member should execute them to achieve optimal performance in exploration and victim identification.

In \cite{cipolleschi2013semantically}, Cipolleschi et al. proposed a system that exploits semantic information to push robots to explore areas that are relevant; according to a priori information provided by human users. That semantic information, which embedded in a semantic map, associates spatial concepts (like ‘rooms’ and ‘corridors’) with metric entities, showing its effectiveness to improve total explored area.

Two main issues of multi-robot exploration are the exploration strategy employed to select the most convenient observation locations the robots should reach in a partially known environment and the coordination method employed to manage the interferences between the actions performed by robots \cite{amigoni2013much}. To determine the effect of each issues, Amigoni et al. \cite{amigoni2013much}  studied a search and rescue setting in which different coordination methods and exploration strategies are implemented and their contributions to an efficient exploration of indoor environments are comparatively evaluated. Their results showed that the role of exploration strategies dominates that of coordination methods in determining the performance of an exploring multi-robot system in a highly structured indoor environment, while the situation is reversed in a less structured indoor environment.

To our knowledge, there is not a lot of publications about the grid-based search by multi-robot systems.  A preliminary related work on this topic was undertaken by Spires et al. \cite{spires1998exhaustive} wherein they  proposed  space filling curves such as Hilbert curves for geographical search for targets by multiple robots. Enns et al. \cite{enns2002guidance} addressed the problem of searching  a wide area for targets using multiple autonomous air vehicles. They considered a scenario where viewing from two perpendicular directions is needed to confirm a target. They employed a simple rule wherein the unmanned air vehicles move in lanes to detect the  targets.

 In recent years, there has been an increasing amount of literature on using the Voronoi diagram for multi-robot systems \cite{bash2007exact, wu2007voronoi, bullo2009distributed}. Wurm et al. \cite{wurm2008coordinated}  addressed the problem of exploring an unknown environment with a team of mobile robots. They proposed an approach to distribute the robots over the environment by partitioning the space into segments using Voronoi diagram. Haumann et al. \cite{haumann2010discoverage} proposed a frontier based approach for multi-robot exploration wherein  using a Voronoi partition of the environment, each robot autonomously creates and optimizes the objective function to obtain a collision-free path in a distributed fashion. In \cite{cowley2011rapid}, authors used Voronoi decomposition of the map to build a connected graph of a place for generating the set of exploration goals. Bhattacharya et al. \cite{bhattacharya2013distributed} presented a distributed algorithm that computes the generalized Voronoi tessellation of non-convex environments in real-time for use in feedback control laws for cooperative coverage control in unknown non-convex environments. Yang et al. \cite{yang2015self}  proposed a decentralized control algorithm of swarm robots for target search and trapping inspired by bacteria chemotaxis; by dividing  the target area into Voronoi cells.  Guruprasad and Ghose \cite{guruprasad2011automated, guruprasad2013performance, guruprasad2013heterogeneous} studied employing Voronoi diagram for multi-robot deploy and search.

However, such approaches that use Voronoi diagram for multi-robot search have some weakness. For example, they need a lot of resources for computation to obtain a reasonable performance that depends on the number of robots. Also, they are inefficient and quiet complex to implement in practice \cite{yan2013survey}. This study makes a major contribution to research on multi-robot systems  by demonstrating some new methods of grid-based search to fill such gaps. In particular, our proposed methods do not need much computational resources, and are feasible to implement  in real systems.

 Regarding the shape of search area, there are a lot of research conducted in multi-robot systems with various assumptions about the search area  \cite{baxter2007multi, zlot2002multi, guruprasad2013performance, fink2008multi}. However, few of them propose a comprehensive solution for the shape of the environment and obstacles. For instance, in \cite{baxter2007multi} a structured environment has been assumed  as the search area whereas an unstructured environment has been considered in \cite{zlot2002multi}. In  \cite{guruprasad2013performance},  the search area has been assumed a  region without any obstacles while in \cite{fink2008multi}, the operation has been done in an environment with obstacles.

\section{Communication and Coordination}
It is apparent  that the strength of multi-robot systems rises when the robots cooperate to complete a task. Communication and coordination, as a means of interaction among robots, is essential to the accomplishment of teamwork effectively \cite{derr2009multi}. The  primary objective of coordination is to have a team of autonomous robots work together efficiently to achieve a collective team behavior through local interaction. Recent advances in science and technology in the fields of microelectronics, computing, communication and control systems have made it feasible to deploy a large number of autonomous robots to work cooperatively to accomplish civilian and military missions with the capability to significantly improve the operational effectiveness, reduce the costs, and provide additional degrees of redundancy. Coverage control, flocking, formation control and consensus are some of the problems in cooperative control of multi-robot systems \cite{li2014cooperative}.

Multi-robot coordination has been primarily inspired from nature as there are various coordination behaviours in animals. For instance, flocking in a school of  fish to avoid predators and obstacles, flocking in a group of migrating birds, or a swarm of insects looking for food are some famous examples of coordinating behaviour of animals \cite{okubo1986dynamical}. A seminal study in this area is the work of Reynolds in the 1980s, wherein the flocking behaviour of animals has been simulated \cite{reynolds1987flocks}, which shows that a mathematical study of the animals flocking can be used as a framework for cooperative control of multi-agent systems.

Today we have a large number of industrial systems which are good examples  of multi-agent coordination.  An example is the deployment of sensors in an unknown environment \cite{howard2002mobile} or clock synchronization \cite{sivrikaya2004time} in wireless sensor networks. Formation flying of satellites is another example, which can be employed, e.g., for space interferometers and military surveillance \cite{beard2001coordination, aung2004overview}. Multi-robot teams competitions, e.g., Robocup, are further useful  samples of cooperation control of  multi-agent systems \cite{candea2001coordination}.

\begin{figure}[!htb]
  \centering
 \includegraphics[natwidth=8.5 cm, natheight=5.5 cm]{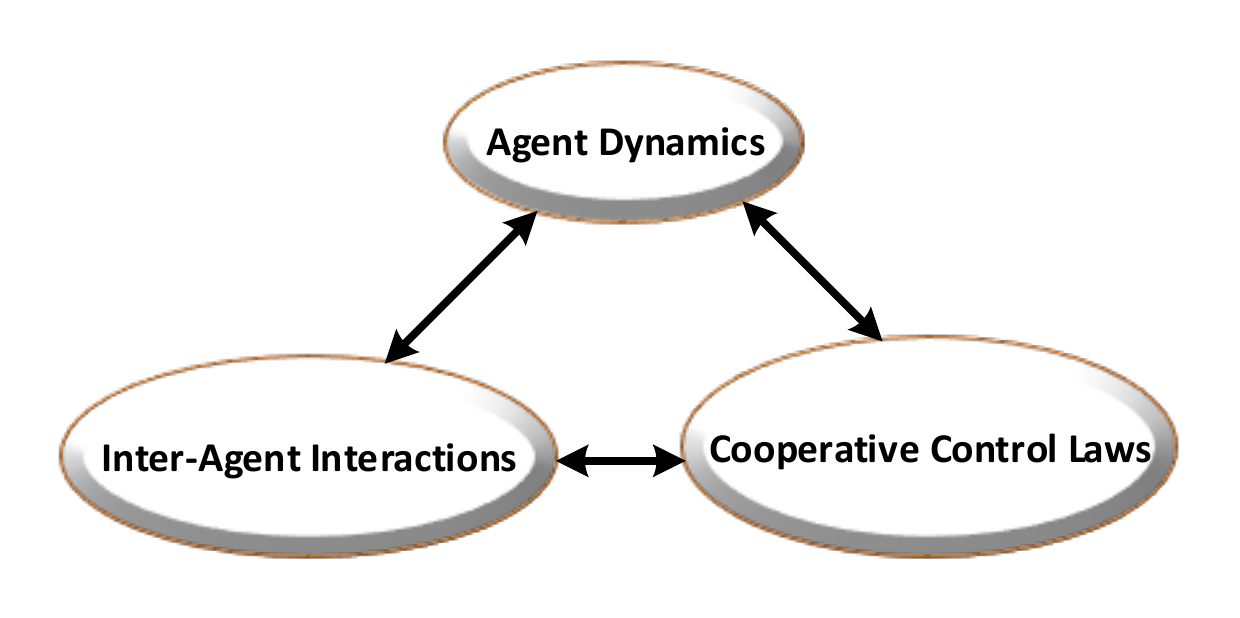}\\
  \caption{Configuration of multi-agent systems}\label{Intro:multiRobotConfig}
\end{figure}

Fig. \ref{Intro:multiRobotConfig} shows the configuration of multi-robot systems \cite{li2014cooperative}. As depicted in this figure, there are three main components in multi-robot systems, which are: agent dynamics, inter-agent interactions and cooperative control laws. Describing  agents dynamics depends on the complexity of the agents, and in most cases, a first-order or second-order dynamics is used to model the agents. In some circumstances, the agents are described  by nonlinear models, and some  uncertainties or disturbances are taken into account.

Note that the agents in a multi-agent system are dynamically decoupled; therefore, they must be coupled through another way enabling them to cooperate. Exchanging information through a communication network  is a method by which the agents can interact with each other. The information which is exchanged among the agents can be their odometry data,  the data of the environment they explore, e.g., maps, and the information about the targets they detect. Direct sensing is another way by which the agents can communicate, for example, a robot can detect the other robots' positions by its sensors such as sonars, lasers or cameras.  It is very beneficial to  represent the topology of the communication between the robots themselves and  with a center in a multi-robot system by a graph \cite{mesbahi2010graph}.  To put it simply, the agents of a multi-robot system are equal to the nodes in the related graph, and an edge between two nodes in the graph implies that there is a communication between two related robots. As the communication between robots can be one-way or two-way, the related graph is directed or undirected. Furthermore, if the robots are always connected via a communication network, the related graph is static meaning that its edges are not time varying. On the other hand, due to the limited communication range of the robots, they sometimes get disconnected. Thus, the corresponding graph will be dynamic, i.e., the edges of the graph vary over time.

In multi-robot systems, cooperative control laws are applied via a control center or distributed among all agents; in agreement with the agent dynamics and the interaction topology. Therefore, coordination control in a multi-robot system is classified into two main categories: centralized and decentralized which will be discussed in the next section. Also, we describe some related problems of cooperative control such as formation control, consensus and flocking later in this chapter. For more detail on these topics and issues of multi-robot cooperation control, see \cite{lewis2013cooperative, bai2011cooperative, cao2013overview, antonelli2013interconnected, jia2013survey} and the references therein.

Networked control systems are another increasingly important area in multi-robot systems \cite{hespanha2007survey}. It has already been common to consider the control and communication as two separate parts in a system. Usually in control theory, we consider that dynamical systems are connected via ideal channels, whereas  communication theory studies data transmission over non-ideal channels. Actually, networked control systems are the combination of these two theories. Since the control signals in multi-robot systems are transmitted through imperfect communication channels with a limited capacity, it is more practical to study them as networked control systems \cite{matveev2009estimation, matveev2001optimal, matveev2003problem, matveev2005comments, savkin2006analysis}. In this report, we consider a multi-robot system as a network, and a decentralized coordination control is applied by which the robots share their information via wireless communication; therefore, our multi-robot system is an example of a networked control system with various communication limitations.

\section{Centralized vs. Decentralized}

For a cooperative control problem, designing an  appropriate controller to carry out the desired team goal is the primary task \cite{cao2013overview}. Centralized approach and decentralized (or distributed) approach  are two major ways to control  multi-robot systems \cite{sarker2014local, luna2011efficient, cheng2011decentralized, pini2011task}. In centralized control systems , there is no communication  between  robots  at  all, and the robots are controlled by a central unit which is responsible for the team coordination. Although multi-robot centralized systems are easier to implement and apply, decentralized control has important advantages so that it has recently brought researchers to pay more attention to it. Also in some applications,  it is hard or even impossible to use a centralized control; then, using a decentralized system is inevitable. Several constraints are imposed by a centralized control such as  limited wireless communication range,  the short sensing capability of sensors, high cost, limited resources and energy, and large size of vehicles to manage and control or even infeasible to implement \cite{cao2013overview}.

In decentralized control, there is not any control center;  thus, the algorithm is such that  the control procedure is distributed among the agents. In other words, all the robots in a decentralized control system are autonomous robots which are controlled by relatively simple control rules \cite{choi2010adaptive, hou2012dynamic}. Thus, decentralized control only relies on local information of a robot itself and the information that it receives from the other robots of the team; usually the information from its neighbours. The decentralized control approach has many advantages in achieving cooperative team performances, specifically with low operational costs, less system requirements, high robustness, strong adaptivity, and flexible scalability; therefore, has been widely recognized and appreciated \cite{cao2013overview}.

In decentralized control of multi-robot systems, the primary  goal typically is to have the whole team working in a cooperative way throughout a distributed law. Here, cooperative means an interaction among all the robots in the team via sharing their  information. Therefore, one of the main issues in  decentralized control systems is how the robots share their information among themselves. When the robots of a team collaborate to achieve a goal based on an algorithm, they need to share their information. That information can be  their motion's  data like position and velocity, the maps they build during the search procedure or the information about the targets they detect. Moreover, in multi-robot systems, a control algorithm must consider some practical limitations  such as  memory usage \cite{chand2012two}, processing speed \cite{chand2013mapping} and communication bandwidth \cite{savkin2010decentralized}. One shortcoming of decentralized control is that some robots cannot predict the team behavior based only on the available local information from their neighbours; so a team behavior cannot be controlled. In addition, in a complicated  task where the goal is to optimize a global cost,  the decentralized control may suffer from the lack of competence. Therefore, it would be interesting to assess a  balance between decentralization and centralization so as to achieve further improvement in overall performance.

\section{Consensus Variables Rule}
Consensus problem in networks is a cooperative  behavior used in many fields of studies such as sensor networks \cite{savkin2015decentrCoverage, cheng2009distributed, cheng2013decentralized}, mobile communication systems, unmanned vehicles, distributed computing  as well as multi-robot systems \cite{Cheng2011497}. The first serious discussions and analysis of consensus problems started with distributed computing in computer science \cite{borkar1982asymptotic, tsitsiklis1984distributed}. Vicsek presented a model for studying collective behavior of interacting self-propelled particles, which is a basis for modeling of many animals behaviours such as a school of fish, flock of birds and swarm of insects \cite{vicsek1995novel, savkin2004coordinated}. Later,  a mathematical  justification for the Vicsek model  was proposed in \cite{jadbabaie2003coordination},  and a general framework of the consensus problem for networks of integrators was provided in \cite{olfati2004consensus}. There is a large volume of published studies describing the role of consensus problems; see, e.g., \cite{ren2005survey, cao2013overview, antonelli2013interconnected, olfati2007consensus} and the references therein.

In the case of  multi-robot systems, the consensus is an important and fundamental problem that means reaching a consensus on an amount for one or more variables. For example, the robots communicate with each other to agree on a particular value for their speed and direction. They share their information and also exchange their calculations based on a rule called consensus algorithm \cite{stanoev2013consensus}. Consensus algorithms are widely used in multi-robot systems such as formation control \cite{savkin2016distributed}, flocking \cite{blondel2005convergence}, and coverage control \cite{savkin2011method, nazarzehi2015distributed}.

In this report, we apply an algorithm based on consensus variables, by which the robots eventually reach an agreement on a common triangular grid. Hence, they can be located on some  vertices of that grid and begin the search operation by moving through the vertices of the grid. Furthermore, in another algorithm proposed for formation building, a control rule based on the consensus variables rule is employed.

\section{Mapping}

Mapping and map merging are two other challenging problems in multi-robot systems. In  single robot systems, a robot makes its map and utilizes  it only by itself of course since there is nobody else to share with. On the other hand, in multi-robot systems, each robot makes its map which is used by the other robots, getting  better performance in an operation. Therefore, map merging  is essential for multi-robot systems; however, it is a  challenging issue due to the different coordinate systems of the robots. Researchers proposed  several methods for mapping and map merging in multi-robot systems. One solution is  based on a global or local positioning system \cite{parker2002distributed} which is not available in many conditions. Another approach assumes that all robots know their positions in a shared map \cite{marjovi2010multi}. There are also methods that merge maps of different robots to build a shared map of the environment \cite{cunningham2012fully, erinc2014anytime}.

Geometric grid based \cite{vu2011grid} and topological maps \cite{marinakis2010pure} are two main methods for map making of mobile robots. Although geometric grid based method is more accurate for the case where the grid size is small enough, it needs considerable memory and takes a remarkable time to process; especially for map merging in multi-robot systems. Thus, in multi-robot systems, topological maps are more efficient and feasible. In this report, we use a kind of topological map, described in detail in Chapter \ref{chap:Consensus}.

\section{Formation Building}
In the field of multi-robot systems, formation control has been the subject of many classic studies in the past two decades. The purpose of formation control  is to drive a group of agents to some desired states; for example, to build a  geometric pattern. There are some multi-robot applications such as remote sensing, patrolling, search and rescue wherein it is helpful if the robots move with a desired geometric formation. Therefore, the research about distribution formation control of a team of mobile robots has become a new challenging area for researchers in recent years; see, e.g., \cite{baranzadehdistributed2015, baranzadeh2015decentralized, olfati2002graph, farrokhsiar2012unscented, turpin2012trajectory} and the references therein. The notable  difference between general distributed control approach and the approaches that are used for a  team of mobile robots is that in the second case, there is no  dynamic coupling among the robots; meaning that the robots do not directly affect each other. Based on the distributed control approach, each robot uses the information provided by its nearest neighbouring robots in order to update its linear and angular velocities at discrete time instants.  There are some articles on this topic that  proposed a distributed control algorithm by which the robots will eventually move with the same heading and speed; see, e.g., \cite{hong2006tracking, yu2008coordinated, liu2013distributed}. The more challenging problem is to apply a distributed control algorithm to force the robots to move so that they finally build a desired geometric pattern. Furthermore, formation building in the existence of obstacles is even a more difficult problem.

Many of the articles presented in this area consider a simple  linear model for the motion of the robots without  constraints on the control inputs; see, e.g., \cite{dong2011robust, kwon2012hierarchical, guo2010adaptive}. In particular, these simple models do not consider the essential standard constraints on the angular and linear velocities. Indeed, all actual vehicles such as Unmanned ground vehicles (UGVs) and  unmanned aerial vehicles (UAVs) have standard hard constraints on their  angular and linear velocities \cite{low2007biologically}. For example, with no constraints on the angular velocity, it may become a large value  which  results in a small turning radius for the robots that is impossible to obtain with actual robots. Therefore, the  linear system approaches considered without constraints on the control inputs  are not applicable.  In \cite{savkin2010decentralized}, an algorithm of flocking for a group of wheeled robots described by the unicycle model with hard constraints on angular and linear velocities was proposed; however, the much more challenging  problem of formation building was not considered. We consider a more difficult problem of formation building where a nonlinear model with hard constraints on the  angular and linear velocities describing the robots.

Communication between the robots is another issue that was considered in many  of the papers in this area. However, most of such publications consider leader-follower in the team; therefore, they must consider  quite restrictive classes of robot communication graphs; see, e.g., \cite{consolini2012class, defoort2008sliding}. In some other papers,  robots communication graph is assumed to be minimally rigid \cite{krick2009stabilisation, wang2012minimally} or time-invariant and connected \cite{mehrjerdi2011nonlinear}  which is also quite restrictive.

 The existence of   obstacles in the environment is the other problem that has been considered  in a lot of recent research on mobile robots control methods; see, e.g.,  \cite{hoy2015algorithms, teimoori2010biologically, matveev2011method, matveev2012real, savkin2014seeking, savkinsimple, hoy2012collision, savkin2013reactive} and the references therein.  The problem of existence of  obstacles in the environment is not considered in most references mentioned above  for formation building methods \cite{savkin2013method}. However, some papers assume obstacles in the environment and include an obstacle avoidance method in their proposed algorithms; see, e.g., \cite{liang2006decentralized,  de2008dynamic, rezaee2014decentralized} and the references therein. In \cite{liang2006decentralized}, it was assumed that the shape of the obstacle is convex and known to the robots. Obstacle avoidance strategy  in \cite{de2008dynamic}  is based on the concept of impedance with fictitious forces. The  technique for obstacle avoidance of mobile robots in  \cite{rezaee2014decentralized} relies on a rotational potential field.

 In this report, we consider the problem of distributed control of a  team of autonomous mobile robots in which the robots finally move with the same direction and speed in a desired geometric pattern  while avoiding the obstacles. We propose a distributed motion coordination control algorithm  so that the robots collectively move in a desired geometric pattern from any initial position while  avoiding the obstacles on their routes. Furthermore, we present the algorithm of formation building with anonymous robots meaning that the robots do not know their final position in the desired geometric configuration at the beginning; but, using a randomized algorithm, they will eventually reach a consensus on their positions. Also, we propose a new obstacle avoidance method by which the robots maintain a given distance to the obstacles as well as mathematical justification of the proposed method.  There are various applications for the suggested  formation control algorithm such as  sweep coverage \cite{cheng2011decentralized, Cheng2011497, choset2001coverage}, border patrolling \cite{kumar2005barrier}, mine sweeping \cite{cassinis1999strategies},  ocean floor monitoring \cite{jeremic1998design}, and exploration in a sea floor \cite{borhaug2007straight}.

\section{Contributions of This Report}

In this report, we propose some new methods for distributed control of a multi-robot team which its duty is to search all or a part of an unknown region. The aim of the search can be either finding some targets in the region, patrolling the region frequently or putting some signs on particular  points in the region. The suggested algorithms use a triangular grid pattern, i.e., robots certainly go through the vertices of a triangular grid during the search operation. To ensure that all the vertices of the region's covering grid are visited by the robots, they must have a common triangular grid. In order to achieve that, we use a two-stage algorithm. In the first stage, robots apply an algorithm according to the consensus variables rule to deploy themselves on the vertices of a common triangular grid which covers the region. In the second stage, they begin searching the area by moving between the vertices of the common triangular grid. There are various scenarios can be used in the second stage. We propose three different methods for the second stage, namely, random triangular grid-based search algorithm, semi-random triangular grid-based search algorithm and modified triangular-grid-based search algorithm, which are presented in Chapters \ref{chap:RandomSearch}, \ref{chap:SemiRandomSearch} and \ref{chap:ModSearch}, respectively.

The proposed  algorithms are partly based on ideas from \cite{savkin2012optimal} where a  distributed random algorithm for self-deployment of a network of mobile sensors has been presented. In \cite{savkin2012optimal}, authors have proposed a method to deploy a team of mobile sensors on the vertices of a triangular grid that covers the entire area of interest. Indeed, in the proposed search algorithms, we benefit from the triangular grid coverage to cover the entire area. It is clear that any triangular grid coverage of a region is a complete blanket coverage of that region. The main advantage of deploying robots in a triangular grid pattern is that it is asymptotically optimal regarding the minimum number of robots required for the complete coverage of an arbitrary bounded area \cite{kershner1939number}. Therefore, using the vertices of this triangular grid coverage, which is applied in this report,  guarantees search of the whole region.

One of the advantages of the proposed algorithms is  the method by which the robots make and share their maps and use them for exploration. We use a kind of topological map, described in detail in the next chapters. Furthermore, we consider a region with an arbitrary shape that contains some obstacles. Also we assume the area is unknown to the robots a priori. In addition, mathematically  rigorous proofs of convergence with probability 1 of the algorithms are given. Moreover, our algorithms are implemented and tested using Mobilesim, a simulator of the real robots and environment. Mobilesim is a powerful simulator which considers robots' real circumstances such as dynamics of motion, encoders, sonar and also  borders and obstacles of the environment. We also test one of the algorithms via experiments by real robots. In fact, we confirm the performance of  the  proposed algorithm with  experiments with Adept Pioneer 3DX wheeled mobile robots in a real world environment.

A further study on networked multi-robot formation building algorithm is presented in this report. We consider the problem of distributed control of a  team of autonomous mobile robots in which the robots finally move with the same direction and speed in a desired geometric pattern  while avoiding the obstacles. We propose a distributed motion coordination control algorithm  so that the robots collectively move in a desired geometric pattern from any initial position while  avoiding the obstacles on their routes.  In the proposed method, the robots have no information on the shape and position of the obstacles and only use range sensors to obtain the information. We use  standard kinematic equations for the robots with hard constraints on the linear and angular velocities.  There is no leader in the team, and the robots apply a  distributed control algorithm  based on the local information they obtain from their nearest neighbours. We take the advantage of using the  consensus variables approach that is a known rule in multi-agent systems. Also, an obstacle avoidance technique based on the information from the range sensors is used. Indeed, we propose a new obstacle avoidance method by which the robots maintain a given distance to the obstacles. We consider quite general class of robot communication graphs which are not assumed to be time-invariant or always connected.  Furthermore, we present the algorithm of formation building with obstacle avoidance with anonymous robots meaning that the robots do not know their final position in the desired geometric configuration at the beginning; but, using a randomized algorithm, they eventually reach a consensus on their positions. Mathematically rigorous proofs of the proposed control algorithms are given, and  the effectiveness of the algorithms are illustrated via computer simulations.

\subsection{Main Contribution Highlights}
The main contributions of this report are summarized as follows:

\begin{itemize}
  \item Three novel algorithms are proposed for search with multi-robot systems. The detailed descriptions of the proposed algorithms are presented.

 \item  The suggested algorithms use a triangular grid pattern, i.e., robots certainly go through the vertices of a triangular grid during the search procedure. The main advantage of using a triangular grid pattern is that it is asymptotically optimal in terms of the minimum number of robots required for the complete coverage of an arbitrary bounded area. Therefore, using the vertices of this triangular grid coverage, what is applied in this report,  guarantees complete search of all the region as well as better performance in terms of search time.

 \item   We use a new kind of topological map which robots make and share during the search operation.

   \item We consider a region with an arbitrary shape that contains some obstacles; also, we assume the area is unknown to the robots a priori.

 \item Unlike many existing heuristic methods, we give mathematically  rigorous proofs of convergence with probability 1 of the proposed algorithms.

  \item We present an extensive simulation study for the proposed algorithms using a powerful simulator of real robots and environment. The results confirm the effectiveness and applicability of the proposed algorithms.

  \item To evaluate the performance of the algorithms,  the experiment results with real Pioneer 3DX mobile robots are presented for one of the search algorithms with detailed descriptions and explanations. The results demonstrate the features of the proposed algorithms and their performance with  real systems.

  \item We compare the  proposed search algorithms with each other and also with three algorithms from other researchers. The comparison shows the strength of our algorithms over the other existing algorithms.

  \item The problem of formation building for a group of mobile robots is considered. A robust decentralized formation building with obstacle avoidance algorithm for a group of mobile robots to move in a defined geometric configuration is proposed. Furthermore, we consider a more complicated formation problem with a group of anonymous robots where the robots are not aware of their position in the final configuration, and have to reach a consensus during the formation process. We propose a randomized algorithm for the anonymous robots which achieves the convergence to the desired configuration with probability 1.

  \item We give mathematically  rigorous proofs of convergence of the proposed algorithms for formation building and also  computer simulation results to confirm the performance and applicability of our proposed solution.

  \item A novel obstacle avoidance rule is proposed which is used in the formation building algorithms.

\end{itemize}

\section{Report Outline}
The remainder of this report is organised as follows:

In Chapter \ref{chap:Consensus}, we describe the problem statement, defining the search problem in details and shedding some lights on terms, assumptions and definitions used in this report. Then, the first stage of the proposed search algorithms is presented that is the consensus variables locating algorithm  by which the robots will be located on the vertices of a common triangular grid. A mathematically proof for the presented algorithm is given as well as some simulation results. Whenever all the robots are located on  the vertices of a common  triangular grid, the search operation starts by moving the robots between the vertices of the common triangular  grid. The first method of search that we propose is the random triangular grid-based search algorithm  presented in Chapter \ref{chap:RandomSearch}. Using this algorithm, the robots randomly move through the vertices of the common triangular grid during the search operation. Therefore, a complete search of the whole area is guaranteed. We give a mathematically rigorous proof of convergence of the presented algorithm as well as the computer simulation results to demonstrate that the algorithm is effective and practicable. In a similar way, in Chapter \ref{chap:SemiRandomSearch}, the semi-random triangular grid-based search algorithm is presented along with a mathematically rigorous proof of the algorithm and  computer simulation results. In Chapter \ref{chap:ModSearch}, the modified triangular grid-based search algorithm is given, and a mathematically rigorous proof of the algorithm along with  computer simulation results are presented. In addition, the experiment results with Pioneer 3DX wheeled mobile robots are presented to confirm the performance of our suggested algorithm.  Finally, a comparison between  the proposed triangular grid-based search algorithms  in  Chapters \ref{chap:RandomSearch}, \ref{chap:SemiRandomSearch} and  \ref{chap:ModSearch} against three other algorithms are given.  Chapter \ref{chap:Formation}, describes a distributed motion coordination control algorithm  so that the robots collectively move in a desired geometric pattern from any initial position while  avoiding the obstacles on their way. Also, a randomized algorithm for the anonymous robots which achieves the convergence to the desired configuration  is presented. Mathematically rigorous proofs of the proposed control algorithms are  given, and the  effectiveness of the algorithms are confirmed  via computer simulations. Chapter \ref{chap:Conclusions} summarizes the work that is presented and discusses possible future research projects as an extension of the presented work.

\chapter{Consensus Variables Locating Algorithm}\label{chap:Consensus}

Consensus problem in networks is a cooperative  behavior used in many fields of studies such as sensor networks, mobile communication systems, unmanned vehicles, distributed computing  as well as multi-robot systems. In the case of  multi-robot systems, it means reaching a consensus on an amount for one or more variables. For example, the robots communicate with each other to agree on a specific value for their speed and direction. They share their information and also exchange their calculations based on a rule called consensus algorithm \cite{stanoev2013consensus}.

In the next three chapters, we present some search algorithms regarding that an area is explored by a team of mobile robots to explore the entire area or to find some targets. The presented algorithms are grid-based search algorithms meaning that the robots certainly pass through the vertices of a grid; a triangular grid in our algorithms. Therefore, for the first step, all the robots must be located on a common triangular grid to start the exploring operation. To accomplish that, we apply an algorithm based on consensus variables, by which the robots eventually reach an agreement on a common triangular grid. Hence, they can be located on some  vertices of that grid and begin the search operation by moving through the vertices of the grid.

In this chapter, the problem statement is described firstly, which defines the problem of search  in details. We also give some terms, assumptions and definitions used  in this chapter and Chapters \ref{chap:RandomSearch}, \ref{chap:SemiRandomSearch} and \ref{chap:ModSearch}. Then, the first stage of the proposed search algorithms is presented that is  "consensus variables locating algorithm"  by which the robots will be located on the vertices of a common triangular grid. A mathematically rigorous proof of the presented algorithm is given as well as some simulation results.

\section{Problem statement}

\begin{figure}[bt]
  \centering
  \includegraphics[width=8.5 cm, height=6 cm]{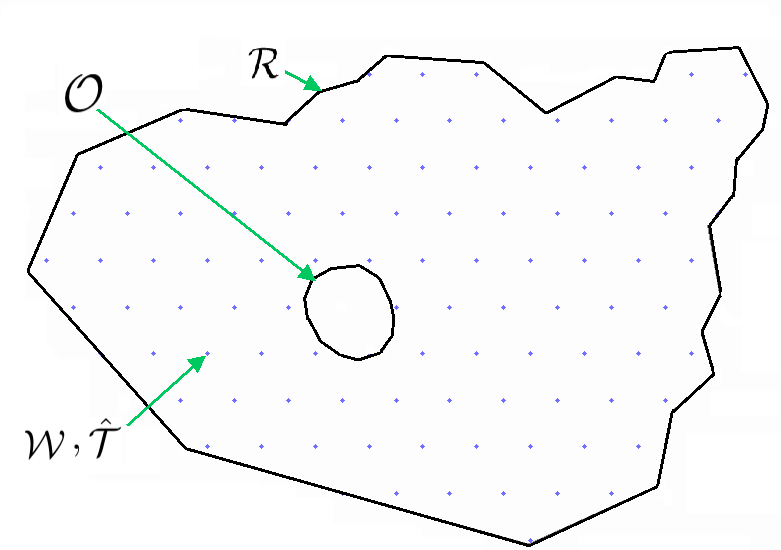}\\
  \caption{A triangular grid (dotted area) covers the search area}\label{CV:area}
\end{figure}

\begin{figure}[bt]
  \centering
  \includegraphics[width=13 cm, height=10 cm]{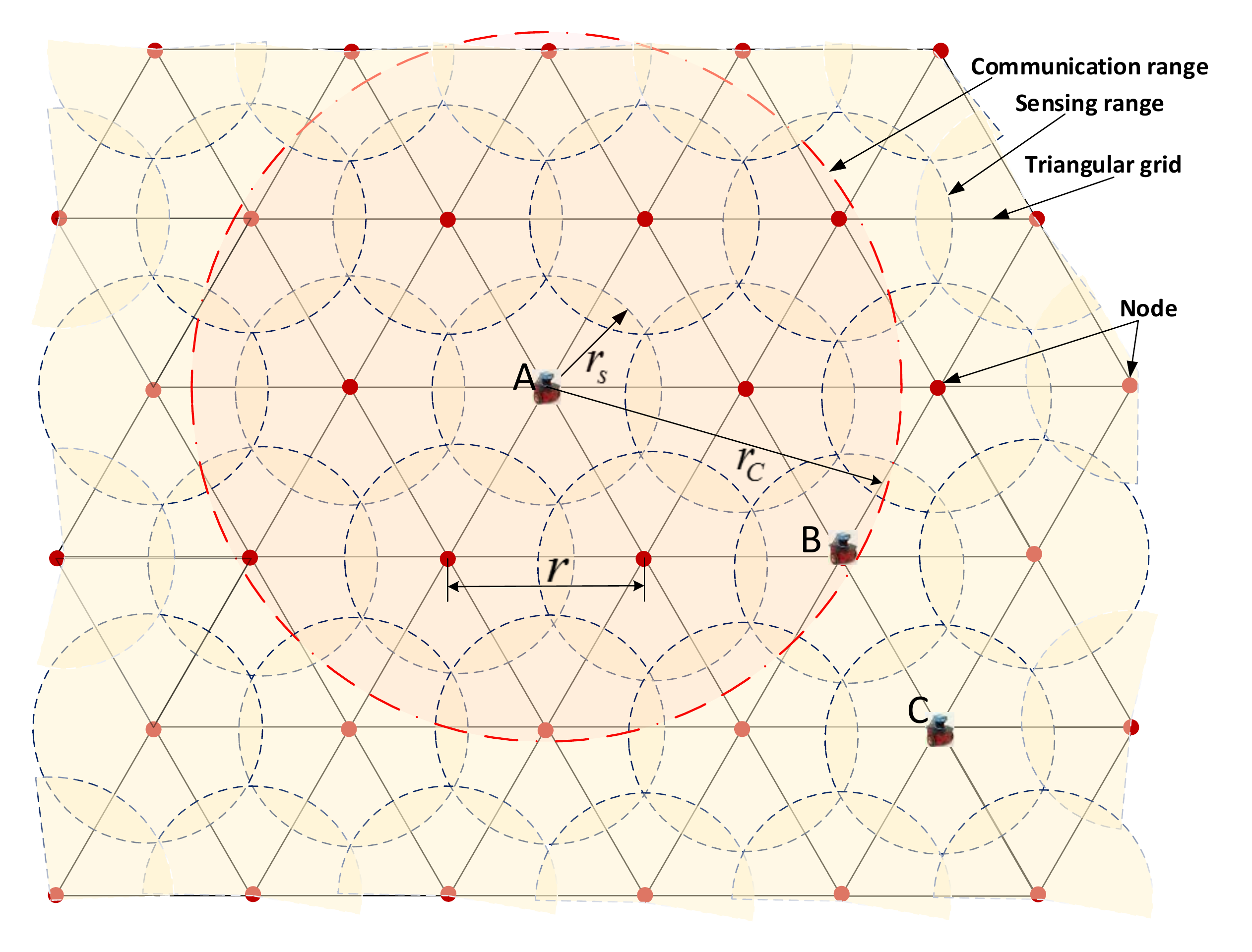}\\
  \caption{Robots on vertices of a triangular grid; small circles are robots' sensing ranges and big circle is the communication range of robot A}\label{CV:TriangleGrid}
\end{figure}

 Consider a planar and  bounded area $\mathcal{R}$. Also, consider a few number of  obstacles $\mathcal{O}$$_{1}$,$\mathcal{O}$$_{2}$,\ldots $\mathcal{O}$$_{m}$ inside the area $\mathcal{R}$ (see Fig. \ref{CV:area}). The goal is to search the whole area by a few  autonomous mobile robots in order to find some targets. The number of targets may be known or unknown to the robots. We use a distributed algorithm to drive the robots inside the search area as well as  avoiding the obstacles and borders. The algorithm is such that the robots follow a pattern to search. The proposed pattern is a triangular grid so that the robots search the area by moving through the vertices of that triangular grid. The grid consists of equilateral triangles with sides $r$. Fig. \ref{CV:TriangleGrid} shows this triangular grid that covers the searching area.

It is assumed that the robots are equipped with sensors to detect the targets, and  these sensors  have a circular sensing area with radius $r_{s}$. As shown in Fig. \ref{CV:TriangleGrid}, the small circles of radius $r_{s}$ with centers  located on the vertices of the triangular grid are the sensing ranges of the robots. It means that robot $i$ can gather information of or detect a target if it is located in a disk of center $p_{i}(k)$ with radius $r_{s}$ defined by $D_{i,r_{s}}(k) :=\{p\in \mathbb{R}$$^{2}:\Vert p-p_{i}(k)\Vert\leq r_{s}\}$, where $k$ indicates discrete time instances; $\ k=0,1,2, \ldots$, $p_{i}(k)\in \mathbb{R}$$^{2}$ denotes the Cartesian coordinates of robot $i$ at time $k$ and  $\Vert \cdot\Vert$ denotes the Euclidean norm. As  demonstrated in Fig. \ref{CV:TriangleGrid}, to have an optimal search operation, the common areas of the robots' sensing regions (small circles)  must be minimum. To achieve this goal, we assume  $r=\sqrt{3}r_s$. However, this is for an  ideal and optimal case but in practice where there is sensor noise and/or position uncertainty, the triangles sides $r$ can be chosen a little smaller than $\sqrt{3}r_s$ that reduces the possibility of unexplored areas.

One of the main benefits of multi-robot systems is that robots share their information in order to improve the strength of the search algorithm in terms of time and cost. The information includes  the position of robots, maps, explored areas and detected targets.   We assume that the robots share their information via a wireless communication so that a robot always sends new information it obtains to the other robots and also listens to receive new information from them. Due to the limited communication range of  the robots, we assume  $r_{c}$ as the communication range  that is the same for all the mobile  robots. It means that a robot can only receive information from the robots which are located not further than $r_{c}$. Therefore, the range of communication for a robot can be defined as the disk of $D_{i,r_{c}}(k) :=\{p\in \mathbb{R}$$^{2}:\Vert p-p_{i}(k)\Vert\leq r_{c}\}$. In Fig. \ref{CV:TriangleGrid}, the big circle of radius $r_{c}$ is the communication range of robot A, meaning that robot A can communicate with robot B but not with robot C. The multi-robot system under consideration is an example of networked control systems in which coordinates and heading of neighbouring robots can be estimated from distance based measurements
using robust Kalman state estimation and model validation techniques via limited communication with each robot’s neighbours; see, e.g., \cite{savkin1996model, moheimani1998robust, petersen1999robust, pathirana2005node}.

\begin{mydef}\label{CV:d1}
Robot $j$ is a neighbour of robot $i$ at time $k$ if it is located on the disk $D_{i,r_{c}}(k)$. So, $\mathcal{N}$$_{i}(k)=\{j:p_{j}\in D_{i,r_{c}}(k), j\in \{1,2,\ \ldots,n\},j\neq i\}$ is the set of all neighbours of robot $i$ at time $k$.  Also, $|\mathcal{N}$$_{i}(k)|$ denotes the number of its neighbours.
\end{mydef}

It is essential for the robots in any search operation to make the map of the searching area when it has not  been available already to the robots. That helps the robots to keep the information of the search operation in their memory for future use or to send it to a center or to the other robots of the search team. The problem of mapping is  a challenging problem  especially when there is a team of robots instead of only one robot. The main problem is merging the maps which are made by different robots. It is much easier for robots to make maps and merge them in the case that there is a central control station or  when robots can use a  central positioning  system like GPS. On the other hand, without a central positioning system, robots have to make their map themselves in their own coordinate systems. Furthermore, the information of maps needs lots of memory also processing of maps for merging is a time consuming task. In such cases , i.e., distributed systems, it is useful to employ topological maps instead of grid geometry maps that need a considerable amount of memory.

In this report, we take advantage of using a kind of topological map that robots make and share among themselves. It is also assumed that the searching area is unknown to all the robots a priori. Therefore, robots have to make the map of the area during the search operation.

 To define the problem, we utilize definitions of \cite{savkin2011method}. Also, to clarify  some terms in the rest of this report, and to state our theoretical results, we introduce a number of assumptions and definitions.

\begin{myassump}\label{CV:a1}
The area $\mathcal{R}$ is bounded and  connected; also, the obstacles $\mathcal{O}$$_{n}$ are non-overlapping, closed, bounded, and linearly connected sets for any $n\geq 0$.
\end{myassump}

\begin{mydef}\label{CV:d2}
Let $\mathcal{O}:=\cup\mathcal{O}$$_{n}$ for all $n\geq 0$; then, introduce $\mathcal{W}:=\{$ $p\in\mathcal{R}$$:p\notin$ $\mathcal{O}\}$.
\end{mydef}

As mentioned before, the robots can detect borders of the area $\mathcal{R}$ and all the obstacles therein. Accordingly, $\mathcal{W}$ is actually those points in the area $\mathcal{R}$ that the robots can go there during the search operation. Also, our search procedure is so that robots go through vertices of a triangular grid that covers this area and cuts the plane into equilateral triangles. A triangular grid can be determined by  its vertices; therefore, when we say a triangular grid, it means the set of all the vertices of it. It is obvious that there exists an infinite number of triangular grids.

\begin{mydef}\label{CV:d3}
Consider $\mathcal{T}$, one of the all possible triangle grids that covers the area $\mathcal{R}$. $\hat{\mathcal{T}} :=\mathcal{T} \cap \mathcal{W}$ is called a triangular grid set in $\mathcal{W}$ (see Fig. \ref{CV:area}).
\end{mydef}\label{CV:d4}
Simply, $\hat{\mathcal{T}}$  is the set of all vertices of the covering triangular grid which robots must visit during the search operation. Since we use  topological maps, $\hat{\mathcal{T}}$ is actually the map of $\mathcal{W}$.

\begin{mydef}\label{CV:d6}
A set  consisting of   all the vertices of a triangular grid in an area is called a map of that area.
\end{mydef}

In fact, what the robots actually save in their memory as maps, are the coordinates of the vertices. A robot can also put some tags on these vertices to assign some attributes like visited, unvisited, occupied, etc. The robots  share their maps with their neighbours too.

 The relationships between robots can be defined by an undirected graph $G(k)$. We assume that any robot of the multi-robot team is a node of the graph $G(k)$ at time $k$, i.e., $i$ in $V_{G}=\{1,2,\ldots,n\}$ , the node set of $G(k)$, is related to robot $i$.  In addition, robot $i$ is a neighbour of  robot $j$  at time $k$ if and only if  there is an edge between the nodes $i$ and $j$ of graph $G(k)$ where $i\neq j$. Therefore, the problem of communication among the team of robots equals the problem of the connectivity of the related graph. It is undeniable that it does not need for  robot $i$  to be the neighbour of robot $j$  to get the information from it. The information can be transferred through the other robots which connect these robots in the related graph. Fig. \ref{CV:TriangleGrid} shows this condition in which robot A cannot directly communicate with robot C but  can do it through robot B.
To guarantee  the  connectivity of the graph, we accept the following assumption \cite{jadbabaie2003coordination}.
\begin{myassump}\label{CV:a2}
There exists an infinite sequence of contiguous, non-empty, bounded, time-intervals $[k_{j},k_{j+1})$, $j= 0,1,2, \ldots$, starting at $k_{0}=0$, such that across each $[k_{j},k_{j+1})$, the union of the collection $\{G(k):k\in[k_{j},k_{j+1})\}$ is a connected graph.
\end{myassump}

Since the main goal of the proposed algorithm is to search an  area in such a way that robots certainly go through the vertices of a triangular grid, firstly, robots have to be located on some vertices of that grid. Thus, we divide our algorithm into two stages. In the first stage, robots make a common triangular grid in order to be finally  located on some vertices of it. To achieve this goal, we apply consensus variables method which is a known distributed method for multi-robot systems \cite{ren2005survey}.
In the second stage, robots start to search the area based on moving between  the vertices of the created grid which is common among all the members of the team. In this chapter, we introduce the first stage of our search algorithms named consensus variables locating algorithm.  Then, we propose three different algorithms for the second stage of our algorithms in Chapters \ref{chap:RandomSearch}, \ref{chap:SemiRandomSearch} and \ref{chap:ModSearch}.

\section{Consensus Variables Locating Algorithm}
We propose some two-stage search algorithms so that in the first stage, using a consensus variables rule, robots make a triangular grid that is common among  all the members of the team \cite{savkin2011method}.  In the beginning, robots are located anywhere in the area $\mathcal{W}$. Each robot can assign its position and heading angle respect to its own coordinate system. The center of a robot at  the starting point is assumed to be  the origin of its coordinate system, and its heading vector  as the x-axis. To define a unique triangular grid with equilateral triangles in a plane, we only need a point and an angle. Thus, the point $q$ which is  any vertex of the grid together with  the angle  $\theta$ that is the angle of the grid, uniquely   defines the triangular grid $\hat{\mathcal{T}}[q,\ \theta]$ that covers the area $\mathcal{W}$ (see Fig. \ref{CV:area}). We consider that at the beginning, the triangular grid that a robot makes for itself at each vertex, is based on the  position and heading of the robot that are $q$ and  $\theta$, respectively. Accordingly, each robot has its own grid that is different from the other robots' grids. To  combine these different grids to a unique grid which will be common among all robots, we apply the consensus variables approach.

We assume that  at any time $k$, robot $i$ has two consensus variables; $q_{i}(k)$ and $\theta_{i}(k)$ on which  it builds  its triangular grid $\hat{\mathcal{T}}[q_{i}(k),\ \theta_{i}(k)]$. At first, these consensus variables are not the same for different robots, so their triangular grids are not the same as well. Using the proposed algorithm will bring the consensus variables $q_{i}(k)$ and $\theta_{i}(k)$ from different values of $q_{i}(0)$ and $\theta_{i}(0)$ to  the same values of $q_0$ and $\theta_0$ for all the robots. That is, a common triangular grid for all the members of the team is built based on  $q_0$ and $\theta_0$ which are the same for all.

\begin{myassump}\label{CV:a3}
The initial values of the consensus variables $\theta_{i}$ satisfy $\theta_{i}(0)\in[0,\ \pi)$ for all $i=1,2, \ldots,\ n$.
\end{myassump}

\begin{mydef}\label{CV:d7}
Consider $p$ be a point on the plane; also, $q$ and $\theta$ are a vertex and an angle that build the triangular grid $\hat{\mathcal{T}}[q,\ \theta]$; then,  $C{[q, \theta]}(p)$ will be the closest vertex of $\hat{\mathcal{T}}[q,\ \theta]$ to $p$. If there is more than one vertex, any of them can be chosen.
\end{mydef}

Now, we propose the following rules as the consensus variables locating algorithm:

\begin{equation*}
\theta_{i}(k+1)=\frac{\theta_{i}(k)+\Sigma_{j\in{\mathcal{N}}_{i}(k)}\theta_{j}(k)}{1+|{\mathcal{N}}_{i}(k)|};
\end{equation*}

\begin{equation}\label{CV:consAlg1}
q_{i}(k+1)=\frac{q_{i}(k)+\Sigma_{j\in{\mathcal{N}}_{i}(k)}q_{j}(k)}{1+|{\mathcal{N}}_{i}(k)|}
\end{equation}

\begin{equation}\label{CV:consAlg2}
p_{i}(k+1)=C[q_{i}(k),\ \theta_{i}(k)](p_{i}(k))
\end{equation}

\begin{figure}[!htbp]
\centering
\subfigure[]{\includegraphics[width=10 cm, height=8 cm]{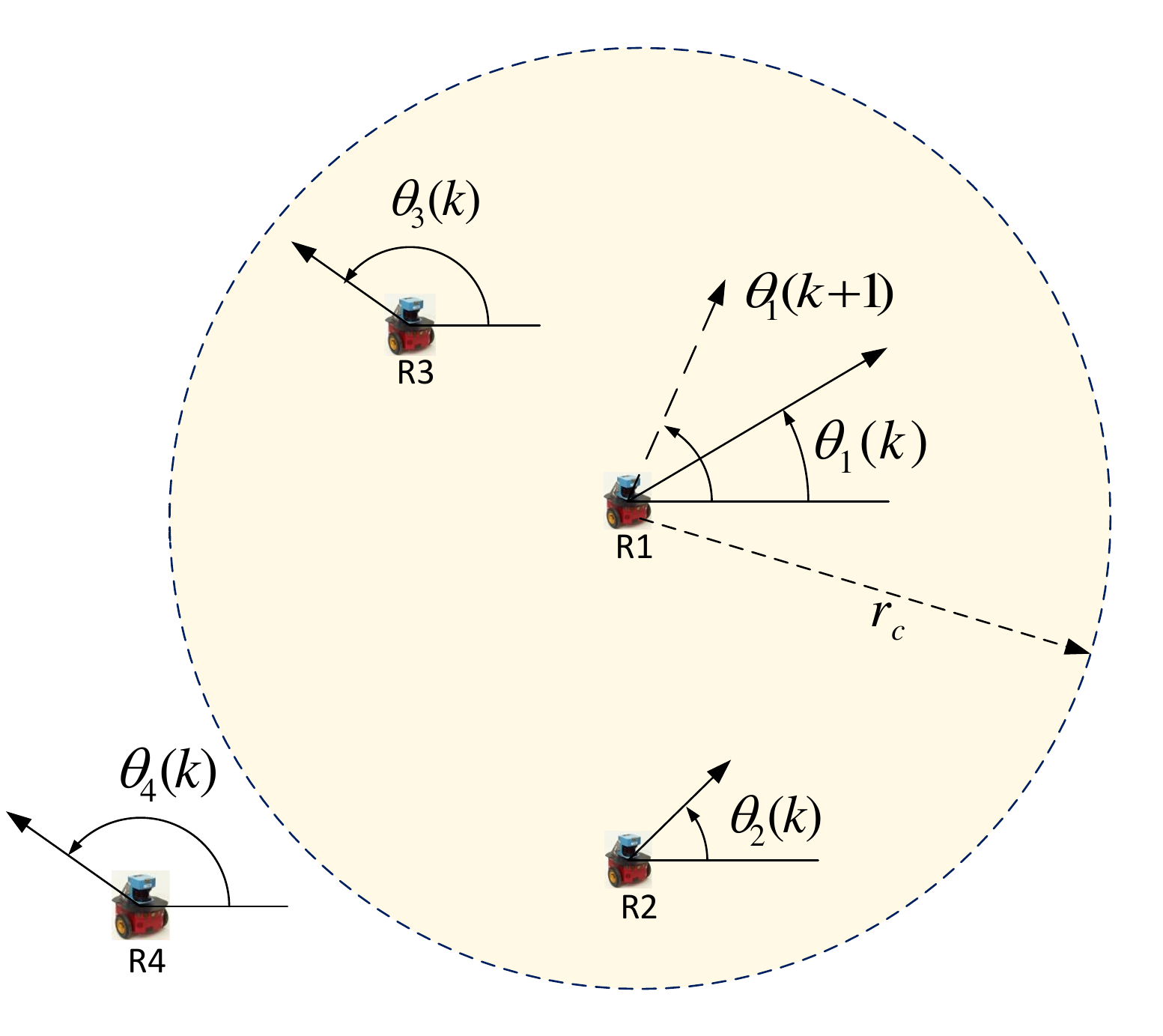}}
\subfigure[]{\includegraphics[width=10 cm, height=8 cm]{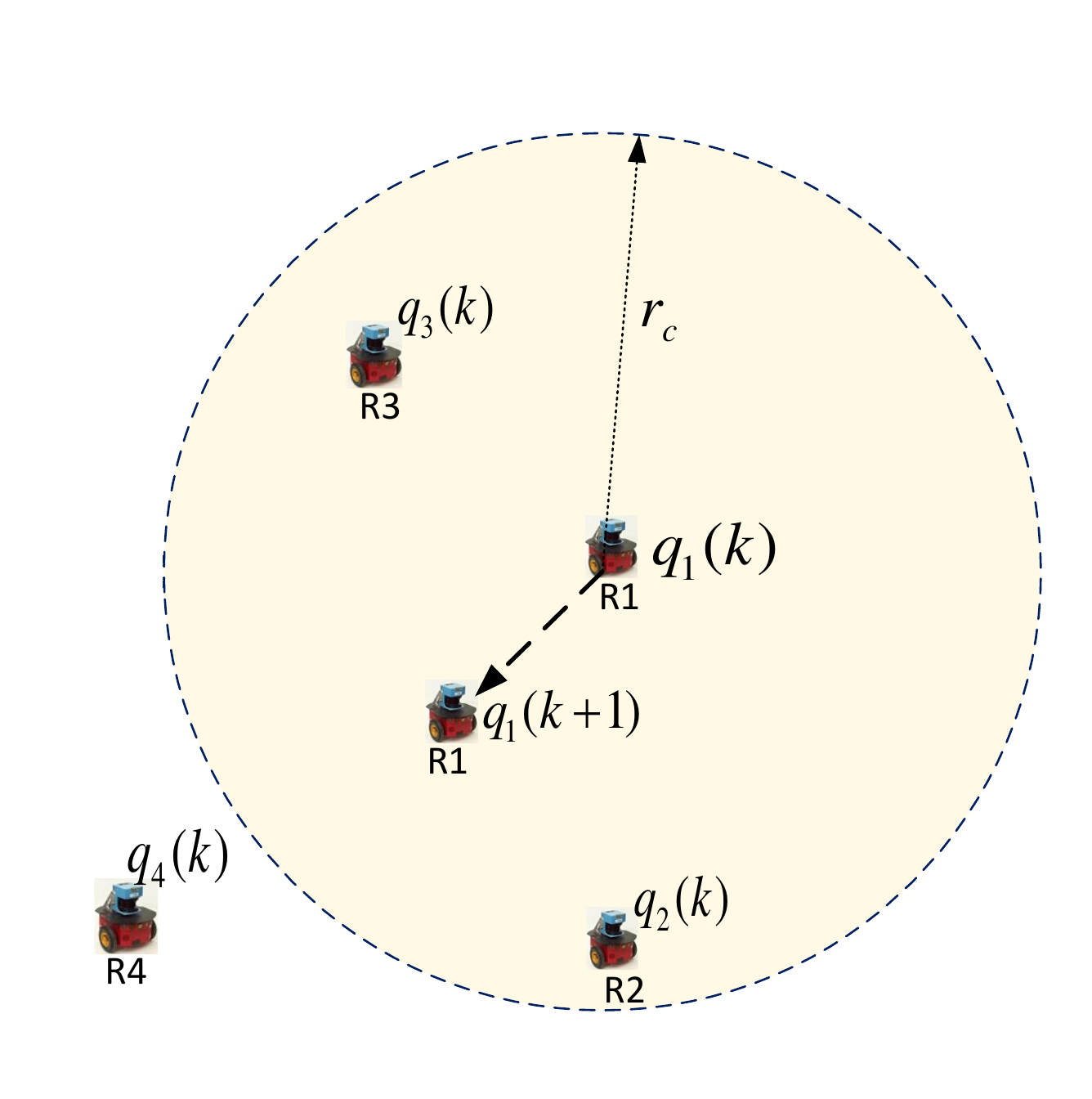}}
\caption{Updating consensus variables $\theta$ and $q$ using the information from the neighbours} \label{CV:consensusFigure}
\end{figure}

The rule (\ref{CV:consAlg1}) causes that the robots reach consensus on heading using $\theta_i$ and gain consensus on phase shift using $q_i$. Based on $q$ and $\theta$ calculated by each robot using rule (\ref{CV:consAlg1}), a robot makes a triangular grid for itself, which is used in rule (\ref{CV:consAlg2}). Fig. \ref{CV:consensusFigure} illustrates the rule  (\ref{CV:consAlg1}). Suppose that there are four robots in the environment. At time $k$, robots R2 and R3 are located in the communication range of robot R1 but robot R4 is out of the range. Therefore, only robots  R2 and R3 are the neighbours of robot R1.  As a result, robot R1 updates variable $\theta_{1}$ using the information from R2 and R3, i.e.,  $\theta_{1}(k+1)$ will be the average of $\theta_{1}$, $\theta_{2}$ and  $\theta_{3}$ at time $k$ (see Fig. \ref{CV:consensusFigure}(a)). In a similar way, the variable $q$ is updated  as shown in Fig. \ref{CV:consensusFigure}(b).

\begin{figure}[!htb]
\centering
{\includegraphics[width=10 cm, height=9 cm]{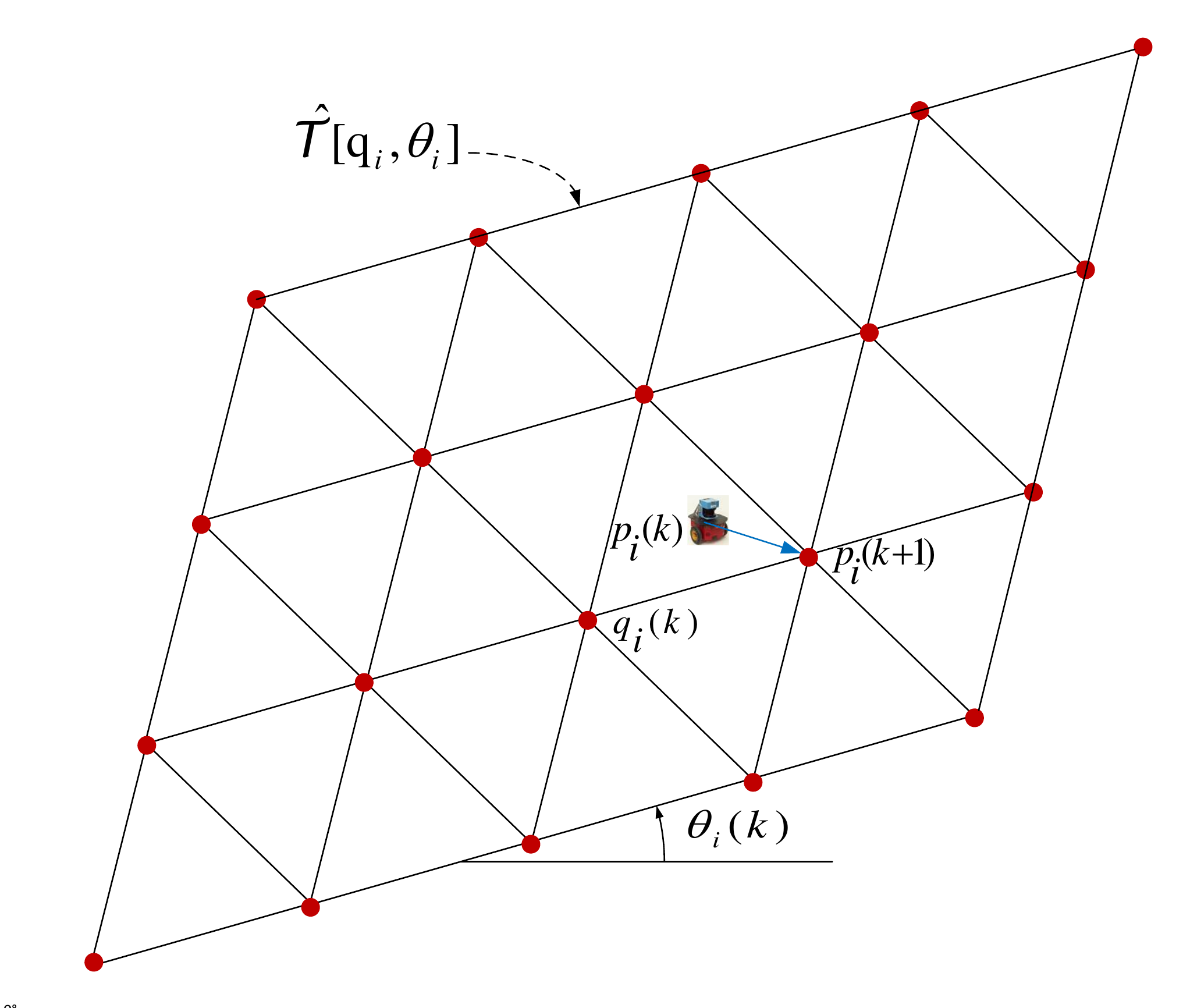}}
\caption{A robot moves to the nearest vertex of its triangular grid} \label{CV:consensusGrid}
\end{figure}

Rule (\ref{CV:consAlg2}) means that whenever a robot makes its triangular grid, it will move to the nearest vertex on it. Fig. \ref{CV:consensusGrid} demonstrates this stage in which  robot $i$ located at $p_{i}(k)$ at time k, makes the  triangular grid $\hat{\mathcal{T}}[q,\ \theta]$ using $\theta_{i}(k)$ and $q_{i}(k)$. Then, it moves to the nearest vertex of the  grid,  which will be  the position of robot $i$  at time $k+1$, i.e.,  $p_{i}(k+1)$. Since rule (\ref{CV:consAlg1}) brings the same $q$ and $\theta$ for all the robots, they eventually will build a common triangular grid, and they all  will be located on its vertices.

\begin{myremark}\label{CV:remark1}
 The robots initially do not have a common coordinate system; otherwise, the consensus building problem would be trivial. Therefore, each robot has consensus variables $\theta_{i}(k)$ and $q_{i}(k)$ in its own coordinate system. However, each robot knows the bearing and the distance to each of its neighbouring robots. Using this information, at any time instance $k$, robot $i$ sends to a neighbouring robot $j$ the consensus variables $\theta_{i}(k)$ and $q_{i}(k)$ re-calculated in the coordinate system with the line $(p_{i},p_{j})$ as the $x$-axis, $p_{j}$ as the origin, and the angle $\theta_{i}(k)$ is measured from this axis in counter-clockwise direction. Using this information, each robot can re-calculate the sums (\ref{CV:consAlg1}) at each time step in its own coordinate system \cite{savkin2011method}.
\end{myremark}

\begin{mytheorem}\label{CV:th1}
Suppose that Assumptions \ref{CV:a1}, \ref{CV:a2} and \ref{CV:a3} hold, and the mobile robots move according to the distributed control law (\ref{CV:consAlg1}), (\ref{CV:consAlg2}). Then, there exists a triangular grid set $\hat{\mathcal{T}}$ such that for any $i=1,2,\ \ldots,\ n$, there exists a $\tau\in\hat{\mathcal{T}}$ such that $\displaystyle \lim_{k\rightarrow\infty}p_{j}(k)=\tau$.
\end{mytheorem}

 {\bf Proof of Theorem \ref{CV:th1}}: Assumption \ref{CV:a1} and the update rule (\ref{CV:consAlg1}) guarantee that there exist a $\theta_{0}$ and $q_{0}$ such that
\begin{equation}\label{CV:eq3}
\theta_{i}(k)\rightarrow\theta_{0},\ q_{i}(k)\rightarrow q_{0}\ \forall i=1,2,\ \ldots,\ n\text{   }
\end{equation}
(see \cite{jadbabaie2003coordination}). Furthermore, the update rule (\ref{CV:consAlg2}) guarantees that $p_{i}(k+1)\in\hat{\mathcal{T}}[q_{i}(k),\ \theta_{i}(k)]$. Therefore, this and (\ref{CV:eq3}) guarantee that $\displaystyle \lim_{k\rightarrow\infty}p_{i}(k)=\tau$ where $\tau\in\hat{{\mathcal{T}}}[q_{0},\ \theta_{0}]$. This completes the proof of Theorem \ref{CV:th1}. $\square $

\section{Simulation Results}

To verify the suggested algorithm, computer simulations are employed. The region $\mathcal{W}$ is considered to be searched by a few robots (see Fig. \ref{CV:area}). We  suppose  a multi-robot team with three robots which are randomly located in the region $\mathcal{W}$ with random initial values of angles. The goal is locating the robots on the vertices of a triangular grid by applying algorithm (\ref{CV:consAlg1}),(\ref{CV:consAlg2}).

\begin{table}[!htb]\footnotesize
\Large
\centering
\caption{Simulation Parameters}
\resizebox{14cm}{!} {
\label{CV:simParameters}
\begin{tabular}{|l|l|l|}
\hline
Linear speed (default)                  & 0.4                          & m/s                                                                                                                       \\ \hline
Linear speed (Maximum)                  & 0.4                            & m/s                                                                                                                       \\ \hline
Angular speed (default)                 & 1.3                          & radian/s                                                                                                                  \\ \hline
Angular speed (Maximum)                 & 1.74                         & radian/s                                                                                                                  \\ \hline
Linear acceleration                     & 0.3                          & $\text{m/s}^{2}$                                                                                                    \\ \hline
Angular acceleration                    & 1.74                         & $\text{radian/s}^{2}$                                                                                                \\ \hline
Linear deceleration                     & 0.3                          & $\text{m/s}^{2}$                                                                                                     \\ \hline
Angular deceleration                    & 1.74                         & $\text{radian/s}^{2}$                                                                                               \\ \hline
\multicolumn{3}{|c|}{\textbf{Localization}}                                                                                                                                                        \\ \hline
Localization method                     & Odometry                     &                                                                                                                           \\ \hline
Localization origin                     & {[}0 0 0{]}                  & $x, y, \theta$                                                                                                               \\ \hline
Odometry error                          & {[} 0.0075 0.0075 0.0075 {]} & \begin{tabular}[c]{@{}l@{}}Slip in x, y and $\theta$ (Uniform random distribution), \\ proportional to velocity\end{tabular} \\ \hline
\multicolumn{3}{|c|}{\textbf{Sonar}}                                                                                                                                                               \\ \hline
Number of sonars                        & 16                           &                                                                                                                           \\ \hline
Minimum view                            & 0.1                          & meter                                                                                                                     \\ \hline
Maximum view                            & 5                            & meter                                                                                                                     \\ \hline
Field of view                           & 30                           & degree                                                                                                                    \\ \hline
Noise                                   & 0.0005                       & meter                                                                                                                     \\ \hline
\multirow{16}{*}{Position (x, y $\theta$)} & 0.069, 0.136, 90             & \multirow{16}{*}{m,m, degree}                                                                                             \\ \cline{2-2}
                                        & 0.114, 0.119, 50             &                                                                                                                           \\ \cline{2-2}
                                        & 0.148, 0.078, 30             &                                                                                                                           \\ \cline{2-2}
                                        & 0.166, 0.027, 10             &                                                                                                                           \\ \cline{2-2}
                                        & 0.166, -0.027, -10           &                                                                                                                           \\ \cline{2-2}
                                        & 0.148, -0.078, -30           &                                                                                                                           \\ \cline{2-2}
                                        & 0.114, -0.119, -50           &                                                                                                                           \\ \cline{2-2}
                                        & 0.069, -0.136, -90           &                                                                                                                           \\ \cline{2-2}
                                        & -0.157, -0.136, -90          &                                                                                                                           \\ \cline{2-2}
                                        & -0.203 -0.119 -130           &                                                                                                                           \\ \cline{2-2}
                                        & -0.237, -0.078, -150         &                                                                                                                           \\ \cline{2-2}
                                        & -0.255, -0.027, -170         &                                                                                                                           \\ \cline{2-2}
                                        & -0.255, 0.027, 170           &                                                                                                                           \\ \cline{2-2}
                                        & -0.237, 0.078, 150           &                                                                                                                           \\ \cline{2-2}
                                        & -0.203, 0.119, 130           &                                                                                                                           \\ \cline{2-2}
                                        & -0.157, 0.136, 90            &                                                                                                                           \\ \hline
\end{tabular}
}
\end{table}

\begin{figure}[!tb]
\centering
\subfigure[]{\includegraphics[width=9 cm, height=6 cm]{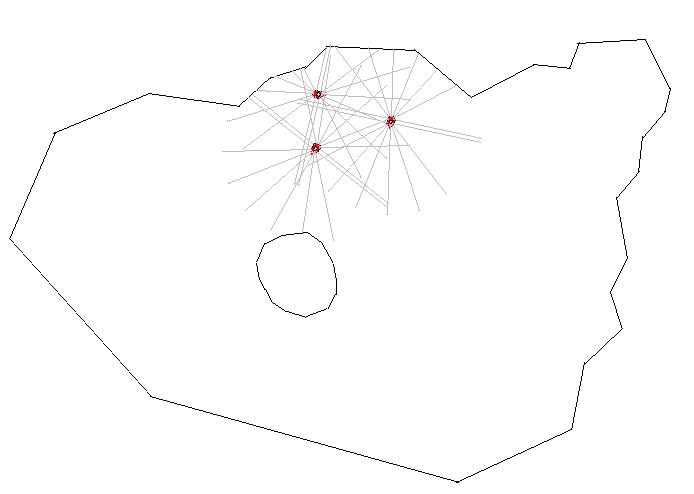}}
\subfigure[]{\includegraphics[width=9 cm, height=6 cm]{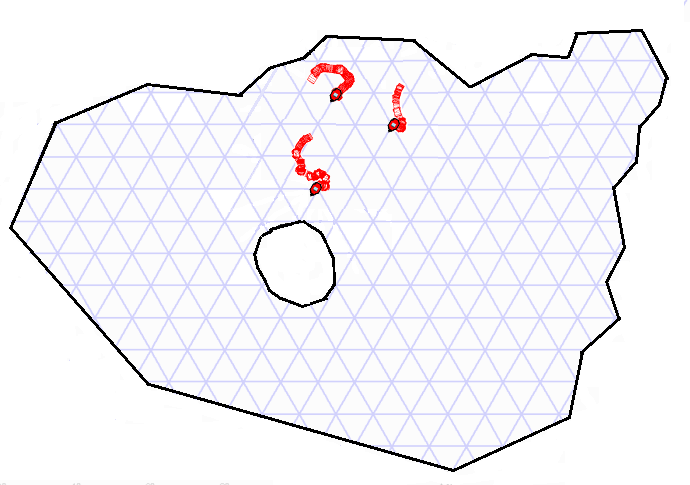}}
\caption{Robots' locations; (a) Initial locations of robots, (b) After applying the consensus variables locating algorithm } \label{CV:stage1SimResult}
\end{figure}

To simulate the algorithm, MobileSime, a simulator of mobile robots developed by  Adept MobileRobots, is used. We also use  Visual C++ for programming and ARIA, a C++ library that provides an interface and framework for controlling the robots. In addition,  Pioneer 3DX is selected as type of the robots, and simulation parameters are given in Table \ref{CV:simParameters}.  To prevent collisions between the robots and to avoid the obstacles and borders,  an obstacle avoidance algorithm is applied using functions provided in ARIA library. Furthermore, to avoid hitting and sticking to the borders, we assume a margin near the borders such that the robots do not pass it.

The simulation results are shown in Fig. \ref{CV:stage1SimResult}. Fig. \ref{CV:stage1SimResult}(a) displays  the initial position of the robots at $k=0$, which are randomly distributed in the area. Applying algorithm (\ref{CV:consAlg1}),(\ref{CV:consAlg2}) will result in locating the robots on the vertices of a common triangular grid. Routes  of the robots and their final position depicted in Fig. \ref{CV:stage1SimResult}(b) shows that they are eventually located at the desired places; on the vertices of a common triangular grid, at time 1m47s.

\section{Summary}

In this chapter, the problem of searching an area by a team of mobile robots using grid-based algorithms has been presented. Some terms, definitions and assumptions have been defined that are used in the proposed algorithms. It has been confirmed that for a grid-based search, the robots must be located on the vertices of a common grid, first. That is a consensus rule named consensus variables locating algorithm has been employed as the first stage of the search algorithms, which locates the robots on the vertices of a common grid among all the robots; a triangular grid in our proposed algorithm.  A mathematically rigorous proof of convergence of the presented algorithm has been demonstrated. Furthermore, the computer simulation results using MobileSim, a powerful simulator of real robots and environment, have been presented to confirm that the algorithm is effective and practicable. In the next three chapters, three grid-based search algorithms will be proposed. The locating algorithm  presented in this chapter will be employed as the first stage of those algorithms.

\chapter{Random Triangular Grid-Based Search Algorithm}\label{chap:RandomSearch}

As described in Chapter \ref{chap:Consensus}, applying consensus variables locating algorithm as the first stage of  a search algorithm, locates all the robots on the vertices of a common triangular grid. After that, the search operation is started by moving the robots between the vertices of the common triangular grid. The first method that we apply for search is "random triangular grid-based search algorithm". By this method, the robots randomly move between the vertices of the common triangular grid so that in each step they only move to the one of the six neighbouring vertices.  A mathematically  rigorous proof of convergence with probability 1 of the algorithm is given. Moreover, our algorithm is implemented and simulated using a simulator of the real robots and environment. The other methods will be presented  in Chapters \ref{chap:SemiRandomSearch} and \ref{chap:ModSearch}.

\section{Distributed Random Search Algorithm }
In order to search an area by a team of mobile robots using the vertices of a grid as the exploring points, we need to locate the searching mobile robots on the vertices of a common grid among the robots. Consensus variables locating algorithm (described in Chapter \ref{chap:Consensus}) can be the first stage of the suggested search algorithm, which  locates all the robots on the vertices of a common triangular grid  $\hat{\mathcal{T}}$ (see Fig. \ref{CV:area}). The next step will be search of the area $\mathcal{W}$ based on moving the robots between the vertices of the covering grid of the area. In this chapter, we propose a random triangular grid-based search algorithm. Suppose a robot located on a vertex of the common triangular grid.  Consequently, it can explore the surrounding  area using its sensors, and that depends on the sensing range of its sensors. After exploring that area,  the robot moves to another point which can be one of the six neighbouring vertices in the triangular grid. As the first method, we suppose that selecting the neighbouring vertex is random.  In this regard, there are a few scenarios  can be considered to search the area. The first scenario is exploring the whole area which  can be applicable when the robots are searching for an undetermined number of targets.  Therefore, to detect  all possible targets, the team of robots must search the whole area. Patrolling of the area is the  other application for this scenario where the robots should move continuously  to detect the possible intruders to the area. In the case of given number of targets which is our second scenario, the search operation should be stopped whenever all the targets are detected without searching  the whole area.

\subsection{Searching the Whole Area}\label{RS:Searchingthewholearea}
To make sure that the whole area is explored by the team of the robots, each vertex in  the triangular covering grid set of the area  $\mathcal{W}$ must be visited at least one time by a member of the team. Consider  $\hat{\mathcal{T}}$ is a triangular covering grid of  $\mathcal{W}$, and also each vertex of $\hat{\mathcal{T}}$ has been  visited at least one time by a robot of the team. This guarantees that the area $\mathcal{W}$ has completely been explored by the multi-robot team. Since the robots do not have any map at the beginning, they need to do map making during the search operation so that their maps will gradually be completed.

\begin{mydef}\label{RS:d8}
Let  $\hat{\mathcal{T}_i}(k)$ be the set of all the vertices of $\hat{\mathcal{T}}$  have been detected by robot $i$ at time $k$. Then,  $\hat{\mathcal{T}}(k)=\bigcup\hat{\mathcal{T}_i}(k)$ will be the map of the  area $\mathcal{W}$ detected by the team of the robots until time $k$.
\end{mydef}
Note that a detected vertex is different from an explored vertex. These terms are defined in detail in the  following definitions.
\begin{mydef}\label{RS:d9a}
A detected vertex means  that vertex is detected by a robot using map making, and it is in the map of that robot though it might be  visited or not  by the robots.
\end{mydef}
\begin{mydef}\label{RS:d9b}
An explored vertex is a vertex that is visited by, at least, one member of the team.
\end{mydef}
\begin{mydef}\label{RS:d10}
The map of robot $i$ at time $k$, ${\mathcal{M}_i}(k)$, is the set of the vertices in  $\hat{\mathcal{T}}$ detected by robot $i$ itself or received from other robots by which  they are detected until time $k$.
\end{mydef}

\begin{mydef}\label{RS:d11}
Suppose  a Boolean variable $V_{\tau}(k)$ which defines the state of vertex  $ \tau\in \hat{\mathcal{T}}(k)$ at time $k$. $V_{\tau}(k)=1$ if the vertex $\tau$ has already been  visited  by, at least, one of the robots, otherwise $V_{\tau}(k)=0$.
\end{mydef}

\begin{myassump}\label{RS:a4}
The triangular grid set $\hat{\mathcal{T}_i}(k)$; $k=0,1,...$ is a connected set. That means that if $ \tau\in \hat{\mathcal{T}_i}(k)$, then, at least, one of the six nearest neighbours of $\tau$ also belongs to $\hat{\mathcal{T}_i}(k)$.
\end{myassump}

Let $\aleph(p_{i}(k))$ be a set containing all the closest vertices to $p_{i}(k)$ on the triangular grid $\hat{\mathcal{T}}(k)$; also, consider $|\aleph(p_{i}(k))|$  as the number of elements in $\aleph(p_{i}(k))$. It is clear that $1\leq|\aleph(p_{i}(k))|\leq6$. In addition, assume $\nu$ be a randomly opted element of $\aleph(p_{i}(k))$.

Consider at time $k$ robot $i$ is located at point $p_{i}(k)$, and it wants to go to the next vertex. The following rule is proposed as the random triangular grid-based search algorithm:

\begin{equation}\label{RS:searchRule}
p_{i}(k+1)=
\begin{cases} \nu & \text{if } |\hat{\mathcal{M}_i}(k)|\neq0 \quad with\text{    } probability\quad \frac{1}{|\aleph(p_{i}(k))|}
\\
p_{i}(k) &\text{if } |\hat{\mathcal{M}_i}(k)|=0 \quad
\end{cases}
\end{equation}

where $\hat{\mathcal{M}_i}(k)=\{\mathfrak{m}\in{\mathcal{M}_i}(k);V_{\mathfrak{m}}(k)=0\}$ is the set of all elements of ${\mathcal{M}_i}(k)$ have not been visited before, and $|\hat{\mathcal{M}_i}(k)|$ denotes the number of elements in $\hat{\mathcal{M}_i}(k)$.

Applying the rule (\ref{RS:searchRule}) ensures that the area $\hat{\mathcal{T}}$ is completely explored and every vertex of it is visited at least one time by a robot of the team.

\begin{mytheorem}\label{RS:th2}
Suppose that all assumptions hold, and the mobile robots move according to the distributed control law (\ref{RS:searchRule}). Then, for any number of robots, with probability 1 there exists a time $k_{0}\geq 0$ such that $V_{\tau}(k_{0})=1 ; \quad \forall  \tau\in \hat{\mathcal{T}}$.
\end{mytheorem}

{\bf Proof of Theorem \ref{RS:th2}:} The algorithm \ref{RS:searchRule} defines an absorbing Markov chain which contains many transient states and a number of absorbing states  that are impossible to leave. Transient states are all the vertices of the triangular grid $\hat{\mathcal{T}}$ which have been visited by the robots during the search procedure. On the other hand, absorbing states are the vertices  where the robots stop at the end of the search operation. Using the algorithm \ref{RS:searchRule}, a robot goes to the vertices where may have not been visited yet. Therefore, the number of transient states will eventually decrease. This continues until the number of the robots is equal to the number of unvisited vertices which will be the absorbing states. It is also clear that these absorbing states can be reached from any initial states, with a non-zero probability. This implies that with probability 1, one of the absorbing states will be reached. This completes the proof of Theorem \ref{RS:th2}. $\square$

\begin{figure}[!hbt]
  \centering
  \includegraphics[width=11 cm, height=20 cm]{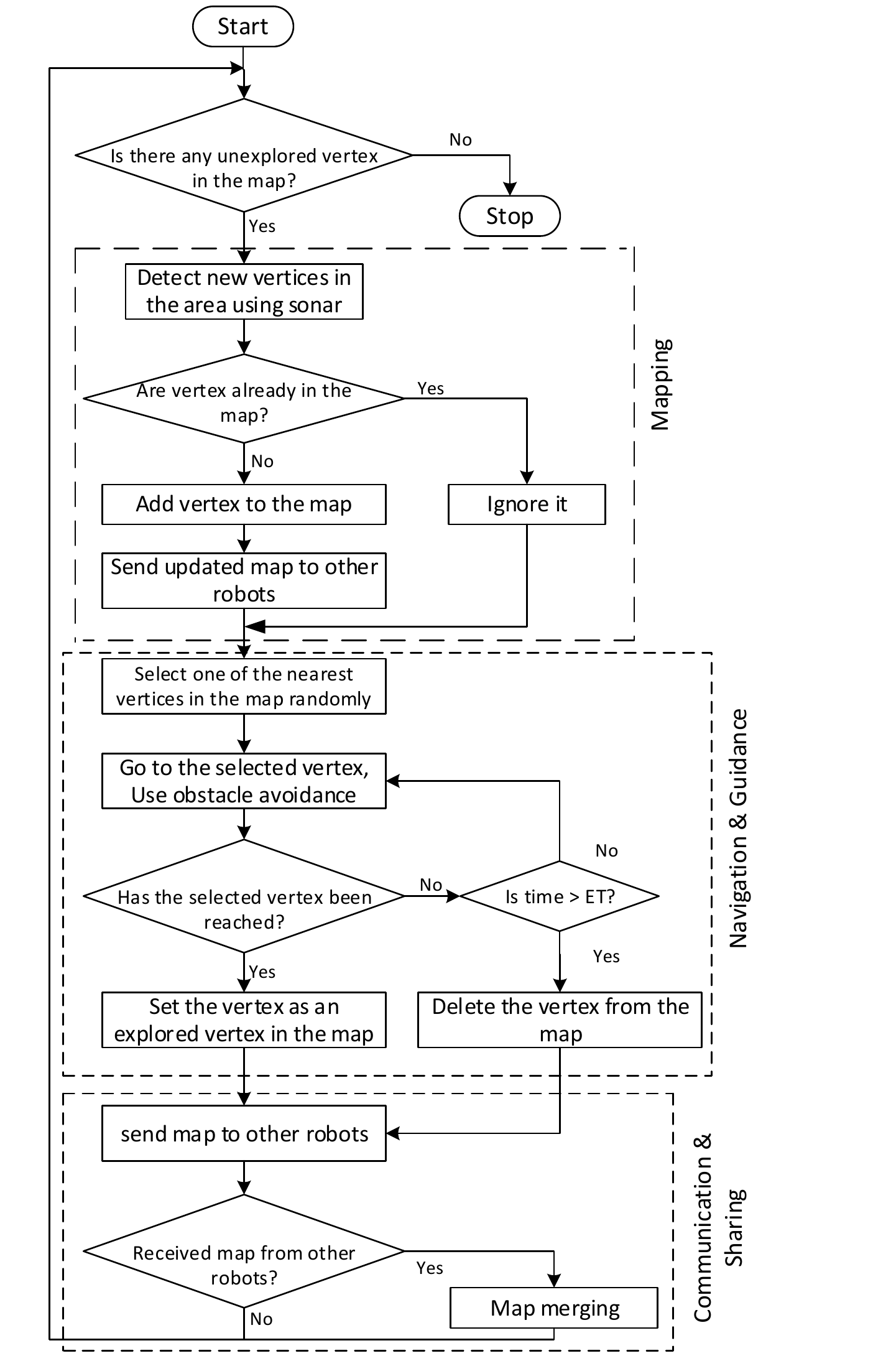}\\
  \caption{The flowchart of the suggested algorithm; searching the whole area}\label{RS:flowchart1}
\end{figure}

In Fig. \ref{RS:flowchart1}, the flowchart of the proposed algorithm is presented that  shows how our decision-making approach is implemented.  At the first step, robots start making their maps using their sonar. Each robot, based on the vertex on which it is located, assumes some probable neighbouring vertices on the common triangular grid. The number of these  probable neighbouring vertices and their distance to the robot depend on the robot's sonar range. Then, the robot uses the sonar to detect its surrounding environment including borders and obstacles. If any of those probable neighbouring vertices is located outside the borders or blocked by an obstacle, it will be ignored. The rest of those probable neighbouring vertices will be added to the map of the robot. This step is repeated every time that the robot occupies a vertex. In order to avoid sticking in borders or obstacles, we consider a margin near the borders and obstacles that depends on the size of the robot. If a vertex  on the map is closer to the borders or obstacles less than the margin, it will be eliminated from the robot's map.

 Whenever a robot makes any changes to its map, it sends the new map to the other robots by transmitting packets. On the other side, whenever a robot receives a packet of data, it extracts  the new vertices from the received map and adds them to its map. Since the communication range of the robots is limited, if two robots are far from each other, they cannot directly communicate but can do it via other robots. In other words, since there is a connected network of robots, each robot has the  role of a hub in the network in order to share the maps among the robots. Therefore, all the robots that are connected and make a network  have a common map.  In the second phase of the algorithm when the robots have a common triangular grid map, a robot can go far from the other robots and be disconnected from the team  for a while. In this case, sharing maps between disconnected robot and the others is paused until the robot returns back to the communication range of the team again.

 In the next step, each robot randomly chooses one of the nearest neighbouring vertices in the map and goes there.  Since we assume that the map of a robot is a connected set, there exists always at least one neighbouring vertex, and at most six vertices. If the robot reaches the target vertex, it marks that vertex as an explored vertex in its map and sends it to the other neighbouring robots as well. However, because of some practical issues, maybe it is impossible to reach the target vertex at a limited time or even maybe the target vertex is fake  that has been wrongly created during mapping. To avoid such problems, we include a factor of time. Since the robot knows its location and the location of the target, it can estimate the time needed to achieve  the goal based on the distance to the target and velocity of the robot. Here, we consider the parameter ET as the expected time to reach the target, that is a factor of the estimated time. This coefficient that has  a value greater than one, actually reflects the effects of a non-straight route because of  the shape of the search area and also  existing obstacles or other robots on the robot's path.  If the travelling time  were more than ET, the robot would ignore that target vertex, delete it from its map, and send the updated map to the other robots.

 Since the area has borders and obstacles, the robots should avoid them while they are searching for  the targets. That is why we have to use an obstacle avoidance in our algorithm. In addition, because of practical problems such as sonar and encoders accuracy and slipping of the robots, there  might be the difference between the actual position of a robot and the coordinates of the vertices stored in its map. Therefore, if a robot is closer to a target vertex than a specified distance, we consider the goal has been achieved.

\subsection{Searching for Targets}
When the robots are looking for some targets in the area, they should continue the search operation till all the targets are detected. If the number of the targets is not specified, they have to search the entire area like what described in Section \ref{RS:Searchingthewholearea}. When robots know the number of the  targets, they do not need to explore the entire area but until all the  targets are detected.
Suppose $\mathfrak{T}=\{\mathfrak{T}_1,\mathfrak{T}_2,\dots, \mathfrak{T}_{n_t} \}$ be the set of $n_t$ static targets should be detected by the robots. As it was stated before, we assume the robots equipped with sensors by which the targets can be detected whenever they are close enough to the robots. The distance by which  the robots must be close to the targets to be able to detect them is the sensing range of the robots ($r_s)$.

\begin{mydef}\label{RS:d12}
Suppose  a Boolean variable $V_{\mathfrak{T}_j}(k)$ which defines the state of target  ${\mathfrak{T}_j}$ at time $k$.  $V_{\mathfrak{T}_j}(k)=1$  if the target   ${\mathfrak{T}_j}$ has been detected by at least one of the robots, otherwise  ${\mathfrak{T}_j}(k)=0$.
\end{mydef}

 Then, we modify the rule (\ref{RS:searchRule}) as the following rule to ensure that the search operation stops after finding all the targets.

\begin{equation}\label{RS:eq5}
p_{i}(k+1)=
\begin{cases} \nu & \text{if } \exists {\mathfrak{T}_j} \in \mathfrak{T} ; V_{\mathfrak{T}_j}(k)=0 \quad
\\
p_{i}(k) &\text{if } \forall {\mathfrak{T}_j} \in \mathfrak{T} ; V_{\mathfrak{T}_j}(k)=1 \quad
\end{cases}
\end{equation}

\begin{mytheorem}\label{RS:th3}
Suppose that all assumptions hold, and the mobile robots move according to  distributed control law (\ref{RS:eq5}). Then, for any number of robots and any number of targets, with probability 1 there exists a time $k_{0}\geq 0$ such that $\forall j$ ; $V_{\mathfrak{T}_j}(k_0)=1$.
\end{mytheorem}

{\bf Proof of Theorem \ref{RS:th3}}: Proof is similar to the  proof of Theorem \ref{RS:th2}.

\begin{figure}[!hbt]
  \centering
  \includegraphics[width=8 cm, height=9 cm]{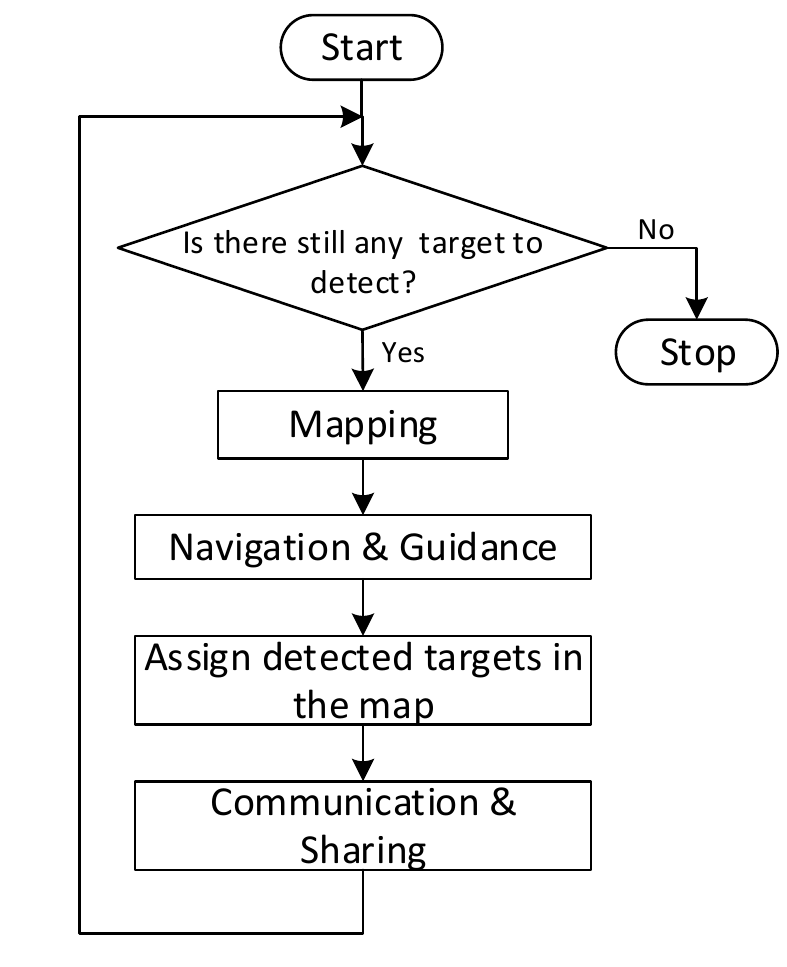}\\
  \caption{The flowchart of the suggested algorithm; searching for targets}\label{RS:flowchart2}
\end{figure}

The flowchart in Fig. \ref{RS:flowchart1} can also be used to describe this operation. Fig. \ref{RS:flowchart2} depicts the procedure we use to implement this algorithm. Most procedures are the same as the previous one; therefore, we ignore their description. We only need to change the condition that stops the operation in the algorithm. The operation will be stopped if all the targets are detected. Also, the robots send  the information about the detected targets  to the other members of the team.

\subsection{Patrolling}

 The above algorithms are appropriate for the cases when the robots should break the search operation; for instance, when the aim is finding some predetermined objects or labeling  some vertices of a  grid. In cases where a permanent search is needed; for example, in applications such as continuously patrolling or surveying a region, the algorithm should be modified such that the search procedure maintains. To achieve that, we modify control law \ref{RS:searchRule} as follows:

 \begin{equation}\label{RS:patrolRule}
p_{i}(k+1)=\nu  \text{ }  with\text{    } probability\quad \frac{1}{|\aleph(p_{i}(k))|};
\end{equation}

 hence, the robots do not stop and continuously move between the vertices of the grid.

\subsection{Robots' Motion}
To have a more real estimation of time that the robots spend to search an area to detect the targets, we apply motion dynamics in our simulations. Besides, since the environment is unknown to the team, it is essential to use an obstacle avoidance in the low-level control of the robots. As a result, we use a method of reactive potential field control in order to avoid obstacles \cite{khatib1986real}.

\section{Simulation Results}

To verify the suggested algorithm, computer simulations are employed. The region $\mathcal{W}$ is considered to be searched by a few robots (see Fig. \ref{CV:area}). We  suppose  a multi-robot team consisting some mobile robots which are randomly located in the region $\mathcal{W}$ with random initial values of angles. The goal is to search the whole area or to find some  targets  by the robots using proposed random triangular grid-based search algorithm.

To simulate the algorithm, MobileSime, a simulator of mobile robots developed by  Adept MobileRobots, is used. We also use  Visual C++ for programming and ARIA, a C++ library that provides an interface and framework for controlling the robots. In addition,  Pioneer 3DX is selected as the type of the robots. Robots' parameters in the simulations are given in Table \ref{CV:simParameters}.  Since this simulator simulates the real robots along with all conditions of a real world, the results of the simulations would be obtained in the real world experiments with the real robots indeed. Furthermore, to prevent collisions between robots and to avoid the obstacles and borders,  an obstacle avoidance algorithm is applied using functions provided in ARIA library. Moreover, to avoid hitting and sticking to the borders, we assume a margin near the borders such that the robots do not pass it.

\begin{figure*}[!hbt]
\centering
\mbox{\subfigure[]{\includegraphics[width=7 cm, height=4.5 cm]{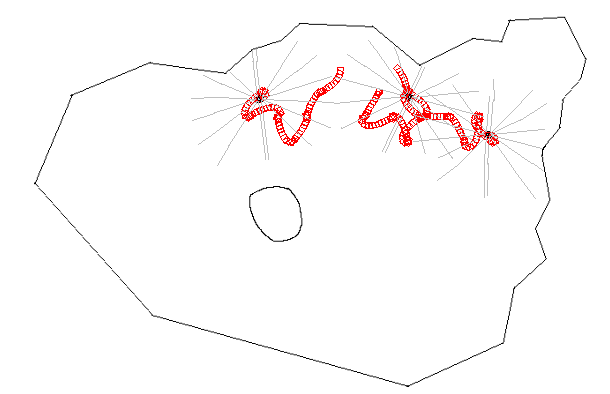}}\quad
\subfigure[]{\includegraphics[width=7 cm, height=4.5 cm]{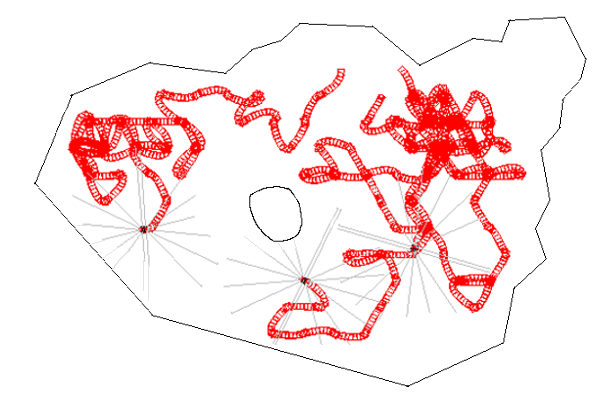}}}
\mbox{\subfigure[]{\includegraphics[width=7 cm, height=4.5 cm]{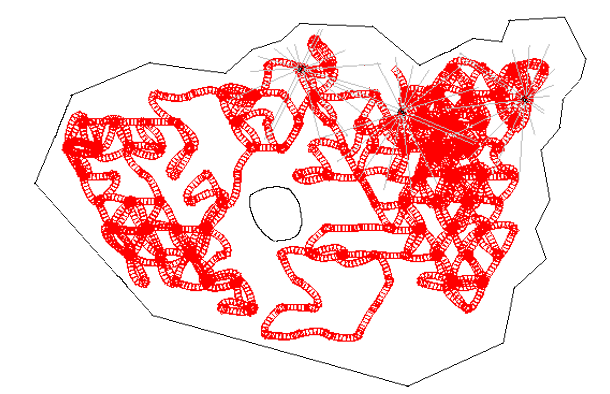}}\quad
\subfigure[]{\includegraphics[width=7 cm, height=4.5 cm]{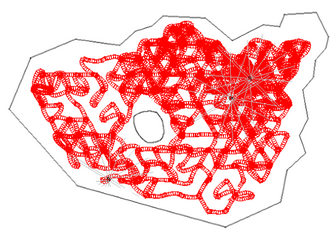}}}
\caption{Robots' trajectories after applying the second stage of the algorithm; (a) After 2m19s, (b) After 7m38s, (c) After 16m50s, (d) After 29m44s}
\label{RS:robotsTrajecs}
\end{figure*}

\subsection{Searching the Whole Area}\label{SearchingWholeArea}
As mentioned  in Chapter \ref{chap:Consensus}, a two-stage algorithm is used to achieve the goal. First, the rule (\ref{CV:consAlg1}),(\ref{CV:consAlg2}) is applied which uses consensus variables in order to drive the robots to the vertices of a common triangular grid. In Fig. \ref{RS:robotsTrajecs}(a), the beginning of the robots' trajectories are the position of the robots after applying the first stage of the search algorithm, i.e., rule (\ref{CV:consAlg1}),(\ref{CV:consAlg2}) (see Fig. \ref{CV:stage1SimResult}).

The second stage of the algorithm, i.e., the algorithm \ref{RS:searchRule}, is applied whenever the first stage is completed. Fig. \ref{RS:robotsTrajecs} demonstrates the result of applying algorithm \ref{RS:searchRule} on a team of three robots. As seen in Fig. \ref{RS:robotsTrajecs}, the robots go through the vertices of the common triangular grid based on the proposed algorithm  until the whole area is explored by the robots. Fig. \ref{RS:robotsTrajecs}(a), Fig. \ref{RS:robotsTrajecs}(b) and Fig. \ref{RS:robotsTrajecs}(c) display the trajectories of the robots at times 2m19s, 7m38s and 16m50s after applying algorithm \ref{RS:searchRule}, respectively. Fig. \ref{RS:robotsTrajecs}(d) shows trajectories of the robots at time 29m44s when the search operation has been  completed. It is obvious that the area $\mathcal{W}$ is completely explored by the robots such that each vertex of the covering triangular grid is occupied at least one time by the robots. In this case study, we assume that the sides of the equilateral triangles are 2 meter (sensing range of the robots is $\frac{2}{\sqrt{3}}$ m) and the communication range between the robots is 10 meter. The area of the region $\mathcal{W}$ is about 528 $m^2$.

\begin{figure} [!bht]
\renewcommand{\topfraction}{.85}
  \centering
  \includegraphics[width=12 cm, height=10 cm]{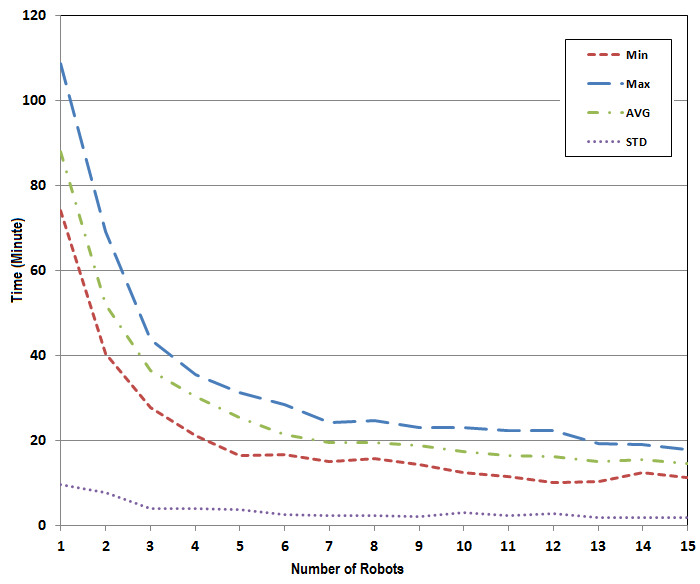}\\
  \caption{Duration of Search Vs. Number of Robots}\label{RS:plotOfTimeOfSearch}
\end{figure}

Although any number of robots can be used to search the whole area,  it is evident  that more robots complete the operation in a shorter time. However, more number of robots certainly increases the cost of the operation. The question is, how many robots should be used in order to optimize both the time and cost. It seems, that is somehow dependent on the shape of the region and the obstacles and also the sides of the triangles. In order to have a better view of a relation between the number of robots and the search duration, twenty  simulations with different number of robots have been done. We consider teams consisting of one to fifteen robots. Then, minimum, maximum, average and standard deviation are calculated for the search duration of each team.  Table \ref{RS:table1} displays the results of these simulations that are also depicted in Fig. \ref{RS:plotOfTimeOfSearch}.
\\
\begin{table}[!bth]
\centering
\caption{Duration of search}\label{RS:table1}
\resizebox{14cm}{!} {
\setlength{\extrarowheight}{20pt}
{\LARGE
\begin{tabular}{|l|l|l|l|l|l|l|l|l|l|l|l|l|l|l|l|}
\hline
\textbf{No of Robots} & \textbf{1} & \textbf{2} & \textbf{3} & \textbf{4} & \textbf{5} & \textbf{6} & \textbf{7} & \textbf{8} & \textbf{9} & \textbf{10}& \textbf{11} & \textbf{12} & \textbf{13} & \textbf{14} & \textbf{15}  \\ \hline
\textbf{Min}   & 74.16  & 40.42 & 27.86 & 21.21 & 16.43 & 16.85 & 15.15 & 15.91 & 14.51 & 12.47 & 11.71 & 10.11 & 10.35 & 12.51 & 11.44 \\ \hline
\textbf{Max}    & 108.64 & 69.13 & 43.68 & 35.59 & 31.43 & 28.48 & 24.27 & 24.80 & 23.08 & 23.17 & 22.52 & 22.50 & 19.40 & 19.15 & 17.93\\ \hline
\textbf{Average}  & 87.88  & 52.09 & 36.43 & 30.51 & 25.56 & 21.35 & 19.54 & 19.69 & 18.94 & 17.52 & 16.44 & 16.23 & 15.14 & 15.52 & 14.64 \\ \hline
\textbf{STD} &  9.67   & 7.83  & 4.17  & 4.18  & 3.74  & 2.73  & 2.51  & 2.52  & 2.22  & 3.17  & 2.36  & 2.89  & 2.06  & 1.95  & 1.94  \\ \hline
\end{tabular}
}
}
\end{table}

\begin{figure*}[!hbtp]
\centering
\mbox{\subfigure[]{\includegraphics[width=6.5 cm, height=4.5 cm]{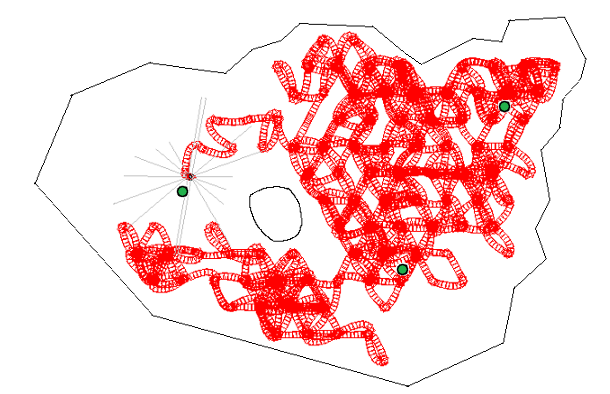}}\quad
\subfigure[]{\includegraphics[width=6.5 cm, height=4.5 cm]{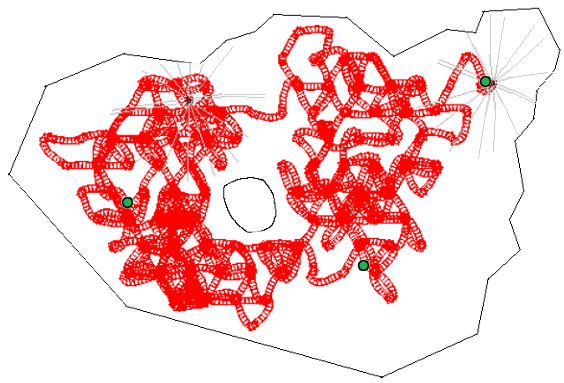}}\quad}
\mbox{\subfigure[]{\includegraphics[width=6.5 cm, height=4.5 cm]{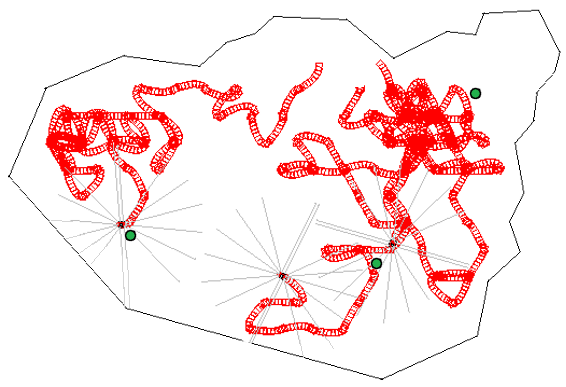}}\quad
\subfigure[]{\includegraphics[width=6.5 cm, height=4.5 cm]{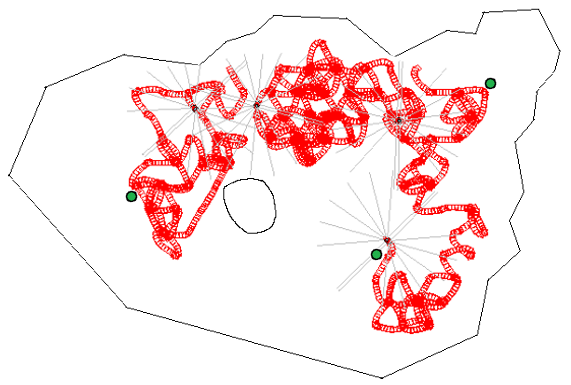}}\quad}
\mbox{\subfigure[]{\includegraphics[width=6.5 cm, height=4.5 cm]{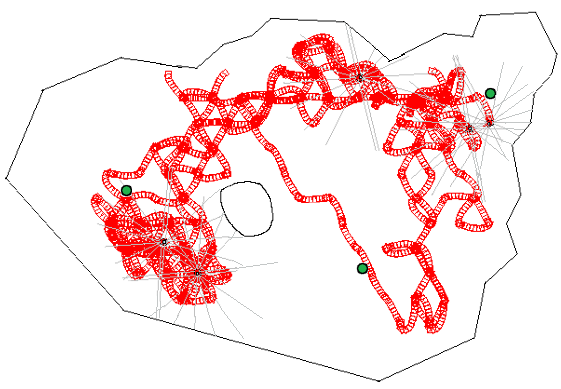}}\quad
\subfigure[]{\includegraphics[width=6.5 cm, height=4.5 cm]{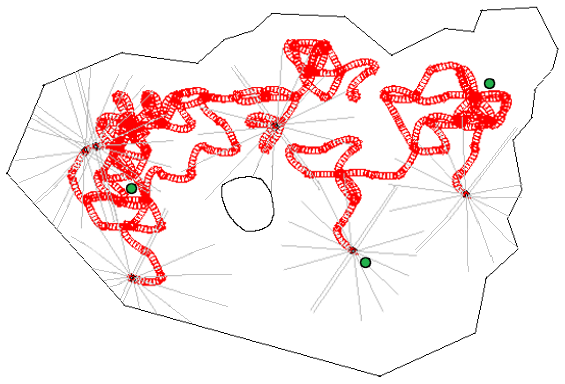}}}
\caption{Robots' trajectories in the case of searching for three targets. (a) One robot detects three targets in 52m25s, (b) Two robots detect the targets in 24m4s, (c) Three robots detect targets in 14m44s, (d) Four robots detect the targets in 14m57s, (e) Five robots detect the targets in 8m27s and (f) Six robots detect the targets in 8m15s. }
 \label{RS:findTargets}
\end{figure*}

 As depicted in this figure, the search time decreases by increasing the number of robots. It is noticeable that  after increasing the  number of robots to a specific number, the search duration almost remains constant. Indeed, increasing the number of robots   increases the probable collision between robots during the operation. Consequently, robots have to turn each other to avoid the collision, and that is  the main reason which increases the search time. Therefore, increasing the number of robots more than a specific value (seven robots in this case) will be ineffectual in terms of time and should be avoided to save cost.  As obvious from Fig. \ref{RS:plotOfTimeOfSearch}, increasing the number of robots more than seven improves the search time less than five minutes.

\subsection{Searching for Targets}

To demonstrate that how the algorithm works in the case of search for a given number of targets, we consider an area the same as in section \ref{SearchingWholeArea} with three targets therein. The targets, shown by green disks in Fig. \ref{RS:findTargets}, are located in different parts of the area. We evaluate the presented algorithm to figure out  how the robots can find the targets using multi-robot teams with different number of robots. As depicted in Fig. \ref{RS:findTargets}, the results of simulation by  the teams consisting of one to six robots have been presented. A target is assumed detected whenever it lies in the sensing range of a robot of the team. If there is just one robot in the team, all the targets should be detected by that robot;  thus, it will take longer time in compare with the cases that there are more robots in the team.

As shown in Fig. \ref{RS:findTargets}(a), a robot searches for three targets using the presented search algorithm and they all have been detected after 52m25s. Fig. \ref{RS:findTargets}(b) shows the same case with  two robots in the team so that a robot has detected one target and the other one has detected two other targets in 24m4s. It should be mentioned that in Fig. \ref{RS:findTargets}, only the paths of the robots after making consensus have been displayed. In  Fig. \ref{RS:findTargets}(c),  three robots search for three targets and each robot finds one of them in 14m44s. As shown in  that figure, when a robot detects a target, it continues the search operation until all targets are detected by the team. Fig. \ref{RS:findTargets}(d)-(f)  show the search operation using 4-6 robots, respectively.

What is very noticeable in this case is that it is expected decreasing the time of detecting the targets by increasing the number of robots while it has not occurred in some instances. For example, when the number of robots has been increased form three to four, the time of detecting the targets has been increased form 14m44s to 14m57s. To explain  why this happens, we should consider the fact that the time needed to detect the targets depends on many parameters not only on the number of robots. The shape of the area and obstacles therein, the initial position of the robots in the area and also the relative distance of the robots and targets are  significant parameters that affect the time of search. The other serious parameter must be considered, is the nature of the search algorithm that is random. That is we might have different paths for the same cases.

\begin{figure}[!hbt]
\centering
\subfigure[Targets are detected in 14m44s]{\includegraphics[width=7 cm, height=5.5 cm]{RandomSearch/RSfindTarget3robots.png}}
\subfigure[Targets are detected in 13m41s ]{\includegraphics[width=7 cm, height=5.5 cm]{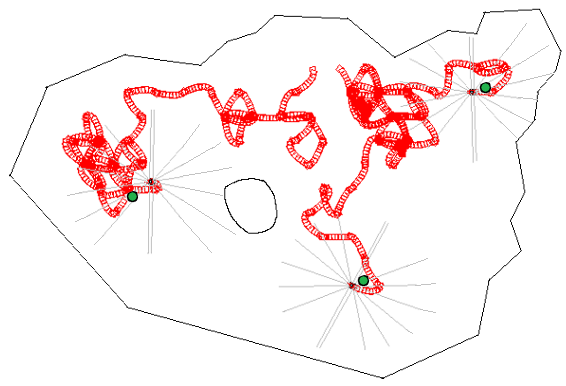}}
\subfigure[ Targets are detected in 19m11s]{\includegraphics[width=7 cm, height=5.5 cm]{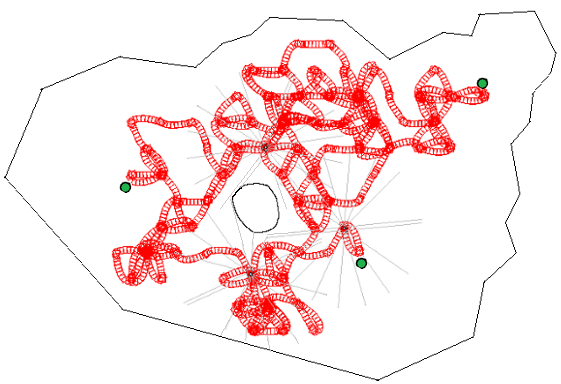}}
\caption{Robots' trajectories in the case of searching for three targets by three robots. } \label{RS:2robotPaths}
\end{figure}

\begin{figure}[!hbt]
\centering
\subfigure[Targets are detected in 8m27s]{\includegraphics[width=7 cm, height=5.5 cm]{RandomSearch/RSfindTarget5robots.png}}
\subfigure[Targets are detected in 11m46s ]{\includegraphics[width=7 cm, height=5.5 cm]{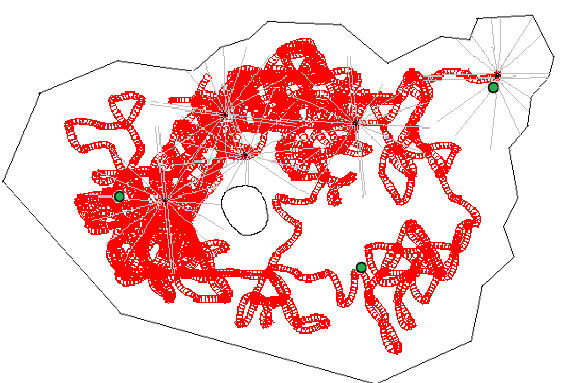}}
\subfigure[ Targets are detected in 14m21s]{\includegraphics[width=7 cm, height=5.5 cm]{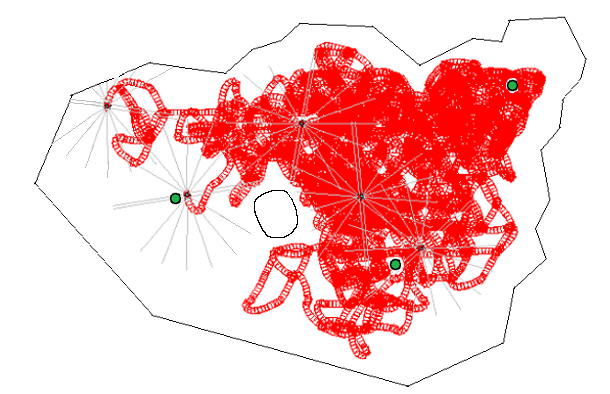}}
\caption{Robots' trajectories in the case of searching for three targets by five robots. } \label{RS:5robotPaths}
\end{figure}

To discover more about how the number of robots affects on the  search duration, we do more simulations for each case. For example, Fig. \ref{RS:2robotPaths} shows the paths of the robots in the case that three robots are looking for three targets similar to the previous instance. It shows that we have different paths thus different  search durations. While in the first simulation, the targets are detected in 14m44s, it takes 13m41 in the second and 19m11s in the third one which is much more than the first one. Fig. \ref{RS:5robotPaths} shows the results of three different simulations using five robots. As depicted in that figure, the period of search for these simulations are 8m27s, 11m46s and 14m21s. Comparing to the case with three robots, it is obvious that the differences between periods of search in this case are less than the case of the team with three robots. That is an expected result because more robots means more coverage of the search area; hence, the chance of detecting  targets increases.

\begin{table}[!bth]
\centering
\caption{Time of detecting targets }\label{RS:table2}
\resizebox{13cm}{!} {
\setlength{\extrarowheight}{6pt}
\begin{tabular}{|l|l|l|l|l|l|l|l|l|l|l|}
\hline
\textbf{No of Robots} & \textbf{1} & \textbf{2} & \textbf{3} & \textbf{4} & \textbf{5} & \textbf{6} & \textbf{7} & \textbf{8} & \textbf{9} & \textbf{10} \\ \hline
\textbf{Min} & 39.13 & 19.14 & 13.74 & 11.21 & 7.97  & 5.62  & 5.38  & 6.23  & 5.41  & 5.04 \\ \hline
\textbf{Max} & 76.38 & 42.67 & 25.87 & 20.49 & 14.35 & 12.87 & 11.97 & 10.95 & 10.67 & 9.83 \\ \hline
\textbf{Average} &55.72 & 29.97 & 19.86 & 15.23 & 11.79 & 9.31  & 8.83  & 8.67  & 8.26  & 7.01 \\ \hline
\textbf{STD} &9.29  & 5.64  & 3.63  & 2.50  & 1.79  & 1.84  & 1.72  & 1.13  & 1.43  & 1.26 \\ \hline
\end{tabular}
}
\end{table}

\begin{figure} [!hbt]
\renewcommand{\topfraction}{.85}
  \centering
  \includegraphics[width=12 cm, height=10 cm]{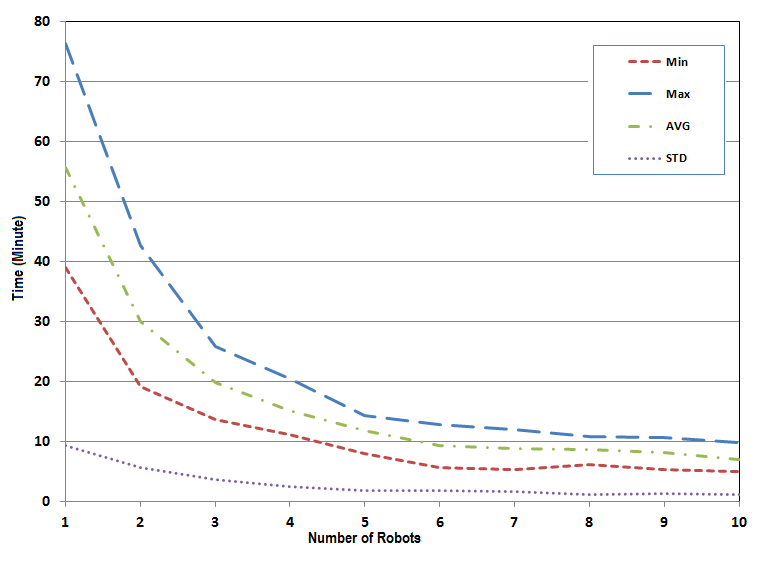}\\
  \caption{Time of target detection Vs. Number of robots}\label{RS:plotOfTimeOfDetect}
\end{figure}

In order to have a better view of the relation between the number of robots and the search duration, twenty  simulations with different number of robots have been done. We consider teams consisting of one to ten robots. Then, minimum, maximum and average of  search duration for simulations with each team as well as their standard deviation are calculated. Table \ref{RS:table2} displays the results of these simulations that are also depicted in Fig. \ref{RS:plotOfTimeOfDetect}. The results show that although the average of search duration to detect the targets decreases by increasing the number of robots, the standard deviation of the team with less number of robots is significantly high. Simply, the number of robots should be proportional to the search area to have an adequate chance to detect targets in an acceptable time.

\section{Summary}
In this chapter, we have developed a distributed control algorithm, namely,  random triangular grid-based search algorithm, to drive a multi-robot team to explore an unknown area. We have used a triangular grid pattern and a two-stage random  search algorithm for the control law so that the robots randomly move through the vertices of the triangular grid during the search operation. Therefore, a complete search of the whole area has been guaranteed. A mathematically rigorous proof of convergence of the presented algorithm has been demonstrated. Furthermore, the computer simulation results using MobileSim, a powerful simulator of real robots and environment, have been presented to show that the algorithm is effective and practicable.

\chapter{Semi-Random Triangular Grid-Based Search Algorithm}\label{chap:SemiRandomSearch}

In Chapter \ref{chap:RandomSearch}, the random triangular grid-based search algorithm was presented. In that method, the robots randomly move between the vertices of a triangular grid so that in each step they only move to the one of the closest neighbouring vertices. In a similar way, we present the second method of search, namely, "semi-random triangular grid-based search algorithm"  in this chapter. By this method, the robots still randomly move between the vertices of a triangular grid so that in each step they move to the one of the closest neighbouring vertices, but  only to those vertices which have not  been visited by the robots yet. If all the neighbouring vertices have already been visited, one of  them will randomly be selected.  A mathematically  rigorous proof of convergence with probability 1 of the algorithm is given. Moreover, our algorithm is implemented and simulated using a simulator of the real robots and environment. The third method of search will be presented  in the next chapter.

\section{Distributed Semi-Random Search Algorithm}

To search an area by a team of mobile robots using the vertices of a grid as the exploring points, we need to locate the searching mobile robots on the vertices of a common grid among the robots. Consensus variables locating algorithm (described in Chapter \ref{chap:Consensus}) can be the first stage of the suggested search algorithm, which  locates all the robots on the vertices of a common triangular grid  $\hat{\mathcal{T}}$ (see Fig. \ref{CV:area}). The next step will be search of the area $\mathcal{W}$ based on moving the robots between the vertices of the covering grid of the area. In this chapter, we propose a semi-random triangular grid-based search algorithm. Suppose a robot located on a vertex of the common triangular grid.  Consequently, it can explore the surrounding  area using its sensors, and that depends on the sensing range of its sensors. After exploring that area,  the robot moves to another point which can be one of the six neighbouring vertices in the triangular grid. As the second method, we suppose that selecting the neighbouring vertex is random but from those neighbouring vertices which have not been visited yet by any of the robots. If all the neighbouring vertices have been visited already, then on of them will be randomly selected.  In this regard, there are a few scenarios  can be considered to search the area. The first scenario is exploring the whole area which  can be applicable when the robots are searching for an undetermined number of targets.  Therefore, to detect  all possible targets, the team of robots must search the whole area. Patrolling of the area is the  other application for this scenario where the robots should move continuously  to detect the possible intruders to the area. In the case of given number of targets which is our second scenario, the search operation should be stopped whenever all the targets are detected without searching  the whole area,.

\subsection{Searching the Whole Area}\label{SR:SearchingtheWholeArea}
To make sure that the whole area is explored by the team of the robots, each vertex in  the triangular covering grid set of the area  $\mathcal{W}$ must be visited at least one time by a member of the team. Consider  $\hat{\mathcal{T}}$ is a triangular covering grid of  $\mathcal{W}$, and also each vertex of $\hat{\mathcal{T}}$ has been  visited at least one time by a robot of the team. This guarantees that the area $\mathcal{W}$ has completely been explored by the multi-robot team. Since the robots do not have any map at the beginning, they need to do map making during the search operation so that their maps will gradually be completed.

\begin{mydef}\label{SR:d8}
Let  $\hat{\mathcal{T}_i}(k)$ be as the set of all the vertices of $\hat{\mathcal{T}}$  have been detected by robot $i$ at time $k$. Then,  $\hat{\mathcal{T}}(k)=\bigcup\hat{\mathcal{T}_i}(k)$ will be the map of the  area $\mathcal{W}$ detected by the team of the robots until time $k$.
\end{mydef}
Note that a detected vertex is different from an explored vertex. These terms are defined in detail in the  following definitions.
\begin{mydef}\label{SR:d9a}
A detected vertex means  that vertex is detected by a robot using map making, and it is in the map of that robot though it might be  visited or not  by the robots.
\end{mydef}
\begin{mydef}\label{SR:d9b}
An explored vertex is a vertex that is visited by, at least, one member of the team.
\end{mydef}
\begin{mydef}\label{SR:d10}
The map of robot $i$ at time $k$, ${\mathcal{M}_i}(k)$, is the set of the vertices in  $\hat{\mathcal{T}}$ detected by robot $i$ itself or received from other robots by which  they are detected until time $k$.
\end{mydef}

\begin{mydef}\label{SR:d11}
Suppose  a Boolean variable $V_{\tau}(k)$ which defines the state of vertex  $ \tau\in \hat{\mathcal{T}}(k)$ at time $k$. $V_{\tau}(k)=1$ if the vertex $\tau$ has already been  visited  by, at least, one of the robots, otherwise $V_{\tau}(k)=0$.
\end{mydef}

\begin{myassump}\label{SR:a4}
The triangular grid set $\hat{\mathcal{T}_i}(k)$; $k=0,1,...$ is connected. That means if $ \tau\in \hat{\mathcal{T}_i}(k)$, then, at least, one of the six nearest neighbours of $\tau$ also belongs to $\hat{\mathcal{T}_i}(k)$.
\end{myassump}

Let $\aleph(p_{i}(k))$ be a set containing all the closest vertices to $p_{i}(k)$ on the triangular grid $\hat{\mathcal{T}}(k)$; also, consider $|\aleph(p_{i}(k))|$  as the number of elements in $\aleph(p_{i}(k))$. It is clear that $1\leq|\aleph(p_{i}(k))|\leq6$. In addition, assume $\nu$ be a randomly opted element of $\aleph(p_{i}(k))$. Moreover, Let $\hat{\aleph}(p_{i}(k))$ be a set containing all the closest vertices to $p_{i}(k)$ on the triangular grid $\hat{\mathcal{T}}(k)$ which have not been visited yet also consider $|\hat{\aleph}(p_{i}(k))|$  as the number of elements in $\hat{\aleph}(p_{i}(k))$. It is clear that $1\leq|\hat{\aleph}(p_{i}(k))|\leq6$. Furthermore, assume $\hat{\nu}$ be a randomly opted element of $\hat{\aleph}(p_{i}(k))$.

Consider at time $k$ robot $i$ is located at point $p_{i}(k)$, and it wants to go to the next vertex. The following rule is proposed as the semi-random triangular grid-based search algorithm:

\begin{equation}\label{SR:searchRule}
p_{i}(k+1)=
\begin{cases} \hat{\nu} & \text{if } |\hat{\mathcal{M}_i}(k)|\neq0 \text{    } \& \text{    } |\hat{\aleph}(p_{i}(k))|\neq0 \text{    } with\text{    } probability\text{    } \frac{1}{|\hat{\aleph}(p_{i}(k))|}
\\
  \nu & \text{if } |\hat{\mathcal{M}_i}(k)|\neq0 \text{    } \& \text{    } |\hat{\aleph}(p_{i}(k))|=0 \text{    } with\text{    } probability\text{    } \frac{1}{|\aleph(p_{i}(k))|}
\\
p_{i}(k) &\text{if } |\hat{\mathcal{M}_i}(k)|=0 \quad
\end{cases}
\end{equation}

where $\hat{\mathcal{M}_i}(k)=\{\mathfrak{m}\in{\mathcal{M}_i}(k);V_{\mathfrak{m}}(k)=0\}$ is the set of all elements of ${\mathcal{M}_i}(k)$ have not been visited before, and $|\hat{\mathcal{M}_i}(k)|$ denotes the number of elements in $\hat{\mathcal{M}_i}(k)$.

Applying the rule (\ref{SR:searchRule}) ensures that the area $\hat{\mathcal{T}}$ is completely explored, and every vertex of it is visited at least one time by a robot of the team.

\begin{mytheorem}\label{SR:th2}
Suppose that all assumptions hold, and the mobile robots move according to the distributed control law (\ref{SR:searchRule}). Then, for any number of robots, with probability 1 there exists a time $k_{0}\geq 0$ such that $V_{\tau}(k_{0})=1 ; \quad \forall  \tau\in \hat{\mathcal{T}}$.
\end{mytheorem}

{\bf Proof of Theorem \ref{SR:th2}}: The algorithm \ref{SR:searchRule} defines an absorbing Markov chain which contains many transient states and some absorbing states  that are impossible to leave. Transient states are all the vertices of the triangular grid $\hat{\mathcal{T}}$ which have been visited by the robots during the search procedure. On the contrary, absorbing states are the vertices  where the robots stop at the end of search. Using the algorithm \ref{SR:searchRule}, a robot goes to the vertices where may have not been visited yet. Therefore, the number of transient states will eventually decrease. That continues until the number of robots is equal to the number of unvisited vertices which will be the absorbing states. It is also clear that these absorbing states can be reached from any initial states, with a non-zero probability. This implies that with probability 1, one of the absorbing states will be reached. This completes the proof of Theorem \ref{SR:th2}. $\square$

\begin{figure}[!htp]
  \centering
  \includegraphics[width=11 cm, height=20 cm]{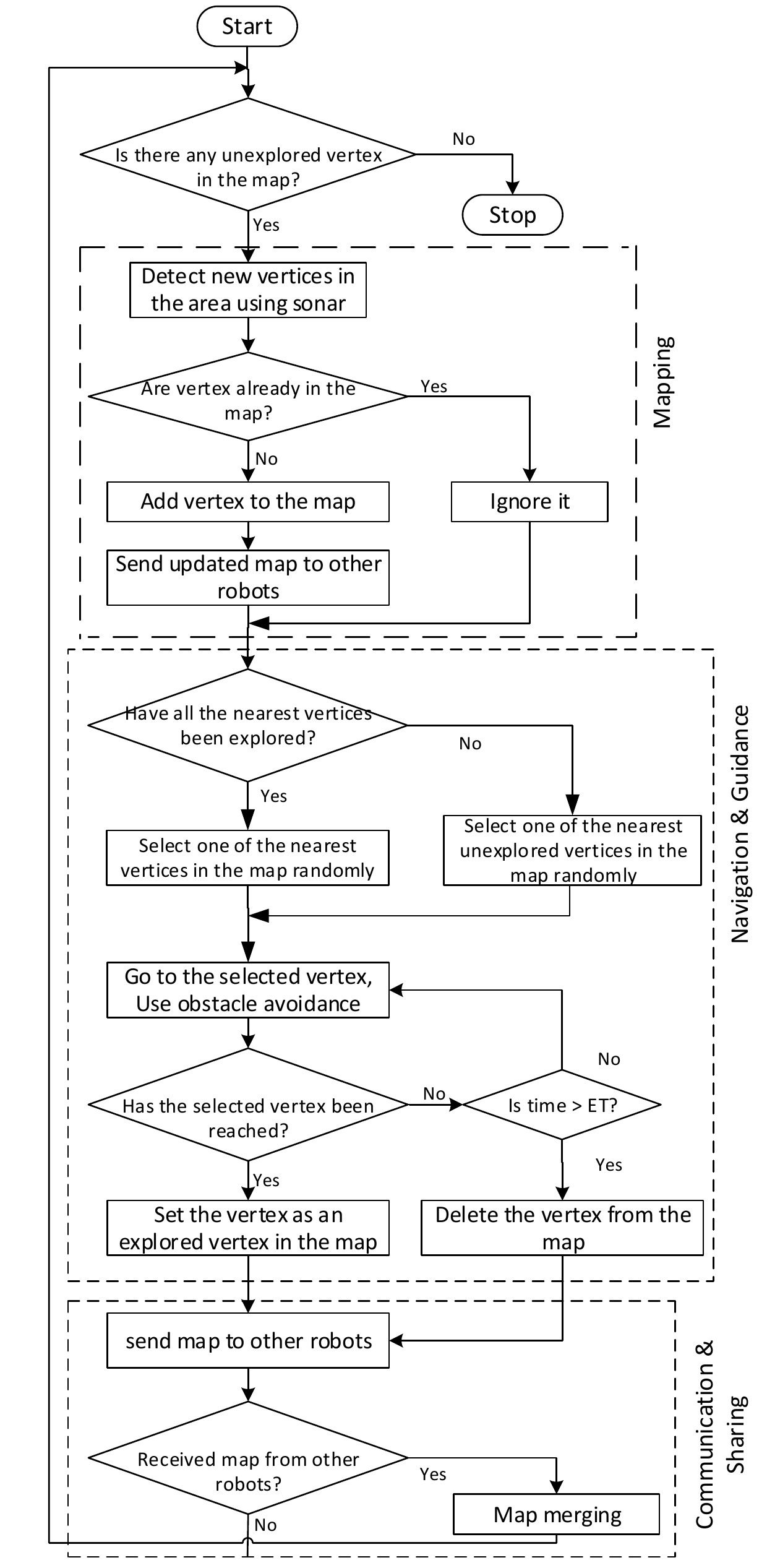}\\
  \caption{The flowchart of the suggested algorithm; searching the whole area}\label{SR:flowchart1}
\end{figure}

In Fig. \ref{SR:flowchart1}, the flowchart of the proposed algorithm is presented that  shows how our decision-making approach is implemented.  At the first step, robots start making their maps using their sonar. Each robot, based on the vertex on which it is located, assumes some probable neighbouring vertices on the common triangular grid. The number of these  probable neighbouring vertices and their distance to the robot depend on the robot's sonar range. Then, the robot uses the sonar to detect its surrounding environment including borders and obstacles. If any of those probable neighbouring vertices is located outside the borders or blocked by an obstacle, it will be ignored. The rest of those probable neighbouring vertices will be added to the map of the robot. This step is repeated every time that the robot occupies a vertex. In order to avoid sticking in borders or obstacles, we consider a margin near the borders and obstacles that depends on the size of the robot. If a vertex  on the map is closer to the borders or obstacles less than the margin, it will be eliminated from the robot's map.

 Whenever a robot makes any changes to its map, it sends the new map to the other robots by transmitting packets. On the other side, whenever a robot receives a packet of data, it extracts  the new vertices from the received map and adds them to its map. Since the communication range of the robots is limited, if two robots are far from each other, they cannot directly communicate but can do it via other robots. In other words, since there is a connected network of robots, each robot has the  role of a hub in the network in order to share the maps among the robots. Therefore, all the robots that are connected and make a network  have a common map.  In the second phase of the algorithm when the robots have a common triangular grid map, a robot can go far from the other robots and be disconnected from the team  for a while. In this case, sharing maps between the disconnected robot and the others is paused until the robot returns back to the communication range of the team again.

 In the next step, each robot randomly chooses one of the nearest unvisited neighbouring vertices in the map and goes there.  If all the neighbouring vertices have been visited already, one of  them is randomly selected. Since we assume that the map of a robot is a connected set, there exists always at least one neighbouring vertex, and at most six vertices. If the robot reaches the target vertex, it marks that vertex as an explored vertex in its map and sends it to the other neighbouring robots as well. However, because of some practical issues, maybe it is impossible to reach the target vertex at a limited time or even maybe the target vertex is fake  that has been wrongly created during mapping. To avoid such problems, we include a factor of time. Since the robot knows its location and the location of the target, it can estimate the time needed to achieve  the goal based on the distance to the target and velocity of the robot. Here, we consider the parameter ET as the expected time to reach the target, that is a factor of the estimated time. This coefficient that has  a value greater than one, actually reflects the effects of a non-straight route because of  the shape of the search area and also  existing obstacles or other robots on the robot's path.  If the travelling time  were more than ET, the robot would ignore that target vertex, delete it from its map, and send the updated map to the other robots.

 Since the area has borders and obstacles, the robots should avoid them while they are searching for  the targets. That is why we have to use an obstacle avoidance in our algorithm. Moreover, because of practical problems like sonar and encoders accuracy and slipping of the robots, there  might be a difference between the actual position of a robot and the coordinates of the vertices stored in its map. Therefore, if a robot is closer to a target vertex than a specified distance, we consider the goal has been achieved.

\subsection{Searching for Targets}
When the robots are looking for some targets in the area, they should continue the search operation till all the targets are detected. If the number of the targets is not specified, they have to search the entire area like what described in Section \ref{SR:SearchingtheWholeArea}. When robots know the number of the  targets, they do not need to explore the entire area but until all the  targets are detected.
Suppose $\mathfrak{T}=\{\mathfrak{T}_1,\mathfrak{T}_2,\dots, \mathfrak{T}_{n_t} \}$ be the set of $n_t$ static targets must be detected by the robots. As it was stated before, we assume the robots equipped with sensors by which the targets can be detected whenever they are close enough to the robots. The distance by which  the robots must be close to the targets to be able to detect them is the sensing range of the robots ($r_s)$.

\begin{mydef}\label{SR:d12}
Suppose  a Boolean variable $V_{\mathfrak{T}_j}(k)$ which defines the state of target  ${\mathfrak{T}_j}$ at time $k$.  $V_{\mathfrak{T}_j}(k)=1$  if the target   ${\mathfrak{T}_j}$ has been detected by at least one of the robots, otherwise  ${\mathfrak{T}_j}(k)=0$.
\end{mydef}

 Therefore, we modify the rule (\ref{SR:searchRule}) as the following rule to ensure that the search operation stops after finding all the targets.

\begin{equation}\label{SR:searchRuleTarget}
p_{i}(k+1)=
\begin{cases} \hat{\nu} & \text{if } |\hat{\mathcal{M}_i}(k)|\neq0 \text{    } \& \text{    } |\hat{\aleph}(p_{i}(k))|\neq0 \text{    }\&\text{    } (\exists {\mathfrak{T}_j} \in \mathfrak{T} ; V_{\mathfrak{T}_j}(k)=0)
\\
  \nu & \text{if } |\hat{\mathcal{M}_i}(k)|\neq0 \text{    } \& \text{    } |\hat{\aleph}(p_{i}(k))|=0 \text{    }\&\text{    } (\exists {\mathfrak{T}_j} \in \mathfrak{T} ; V_{\mathfrak{T}_j}(k)=0)
\\
p_{i}(k) &\text{if } \forall {\mathfrak{T}_j} \in \mathfrak{T} ; V_{\mathfrak{T}_j}(k)=1 \quad
\end{cases}
\end{equation}

\begin{mytheorem}\label{SR:th3}
Suppose that all assumptions hold, and the mobile robots move according to  distributed control law (\ref{SR:searchRuleTarget}). Then, for any number of robots and any number of targets, with probability 1 there exists a time $k_{0}\geq 0$ such that $\forall j$ ; $V_{\mathfrak{T}_j}(k_0)=1$.
\end{mytheorem}

{\bf Proof}: Proof is similar to the  proof of Theorem \ref{SR:th2}.

\begin{figure}[!ht]
  \centering
  \includegraphics[width=8 cm, height=9 cm]{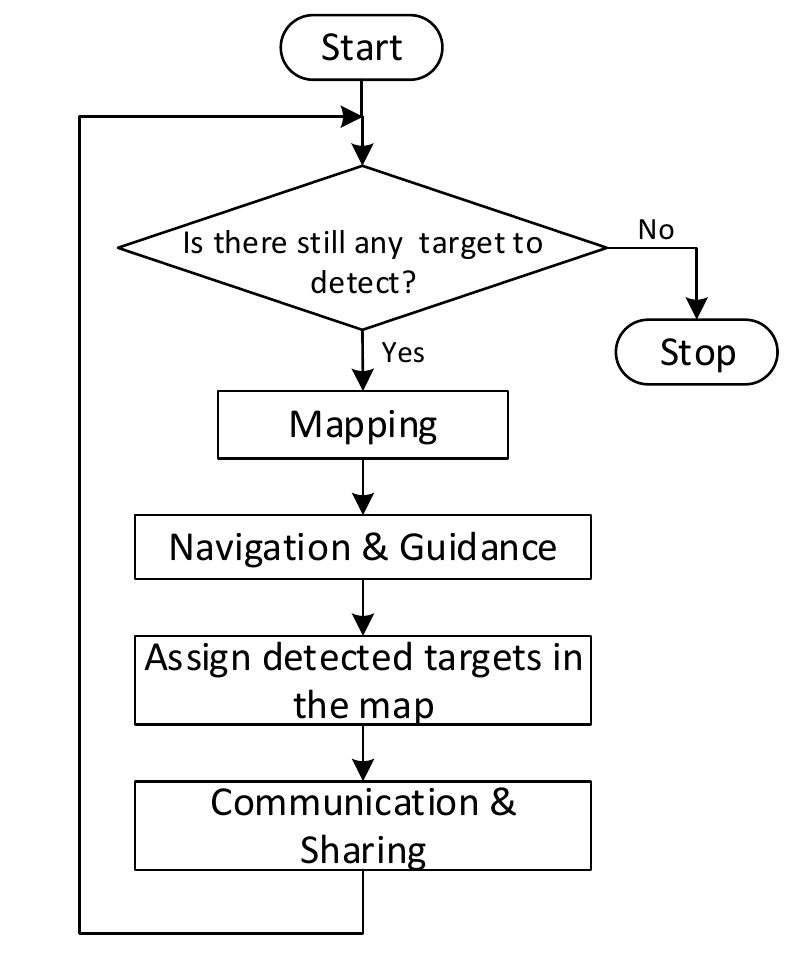}\\
  \caption{The flowchart of the suggested algorithm; searching for targets}\label{SR:flowchart2}
\end{figure}

The flowchart in Fig. \ref{SR:flowchart1} can also be used to describe this operation. Fig. \ref{SR:flowchart2} depicts the procedure we use to implement this algorithm. Most procedures are the same as the previous one; therefore, we ignore their description. We only need to change the condition that stops the operation in the algorithm. The operation will be stopped if all the targets are detected. Also, the robots send  the information about the detected targets  to the other members of the team.

\subsection{Patrolling}

  The above algorithms are appropriate for the cases when the robots should break the search operation; for instance, when the aim is finding some predetermined objects or labeling  some vertices of a  grid. In cases where a permanent search is needed; for example, in applications such as continuously patrolling or surveying a region, the algorithm should be modified such that the search procedure maintains. That is, we modify control law \ref{SR:searchRule} as follows:

\begin{equation}\label{SR:patrolRule}
p_{i}(k+1)=
\begin{cases} \hat{\nu} & \text{if }  \text{    }  |\hat{\aleph}(p_{i}(k))|\neq0 \text{    } with\text{    } probability\text{    } \frac{1}{|\hat{\aleph}(p_{i}(k))|}
\\
  \nu & \text{if }  \text{    } |\hat{\aleph}(p_{i}(k))|=0 \text{    } with\text{    } probability\text{    } \frac{1}{|\aleph(p_{i}(k))|}
\end{cases}
\end{equation}

 hence, the robots do not stop and continuously move between the vertices of the grid.

\subsection{Robots' Motion}
To have a more real estimation of time that the robots spend to search an area to detect the targets, we apply motion dynamics in our simulations. In addition, since the environment is unknown to the team, it is essential to use an obstacle avoidance in the low level control of the robots. As a result, we use a method of reactive potential field control in order to avoid obstacles \cite{khatib1986real}.

\section{Simulation Results}

To verify the suggested algorithm, computer simulations are employed. The region $\mathcal{W}$ is considered to be searched by a few robots (see Fig. \ref{CV:area}). We  suppose  a multi-robot team with some autonomous mobile robots which are randomly located in the region $\mathcal{W}$ with random initial values of angles. The goal is to search the whole area  by the robots using proposed semi-random triangular grid-based search algorithm.

To simulate the algorithm, MobileSime, a simulator of mobile robots developed by Adept MobileRobots, is used. We also use  Visual C++ for programming and ARIA, a C++ library that provides an interface and framework for controlling the robots. In addition,  Pioneer 3DX is selected as the type of the robots.  Robots’ parameters in the simulations are given in Table \ref{CV:simParameters}. Since this simulator simulates the real robots along with all conditions of a real world, the results of the simulations would be obtained in the real world experiments with the real robots indeed. Furthermore, to prevent collisions between robots and to avoid the obstacles and borders,  an obstacle avoidance algorithm is applied using functions provided in ARIA library. Moreover, to avoid hitting and sticking to the borders, we assume a margin near the borders such that the robots do not pass it.

\subsection{Searching the Whole Area}\label{SR:SearchingtheWholeareaSimulation}
\begin{figure*}[!ht]
\centering
\mbox{\subfigure[]{\includegraphics[width=7 cm, height=5 cm]{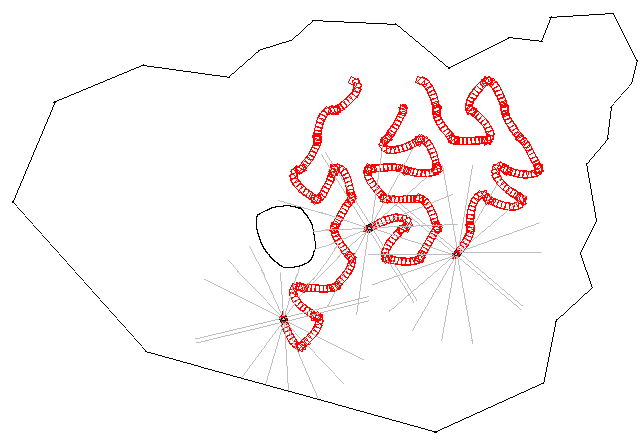}}\quad
\subfigure[]{\includegraphics[width=7 cm, height=5 cm]{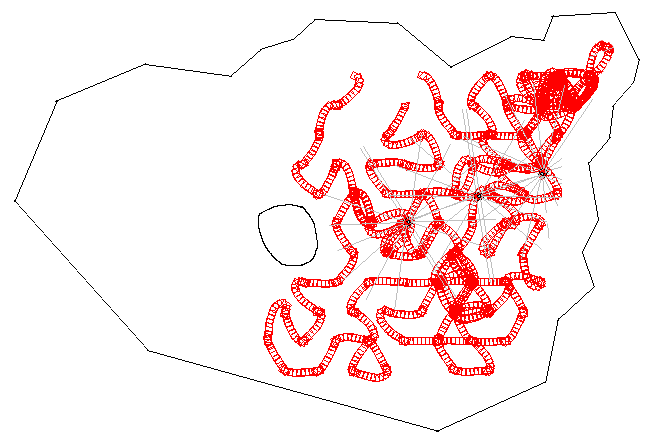}}}
\mbox{\subfigure[]{\includegraphics[width=7 cm, height=5 cm]{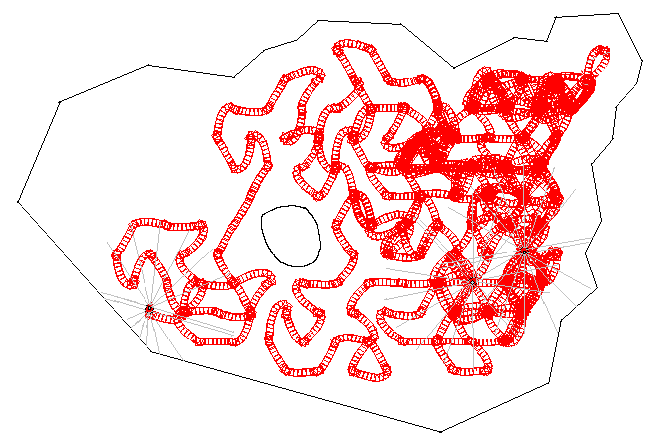}}\quad
\subfigure[]{\includegraphics[width=7 cm, height=5 cm]{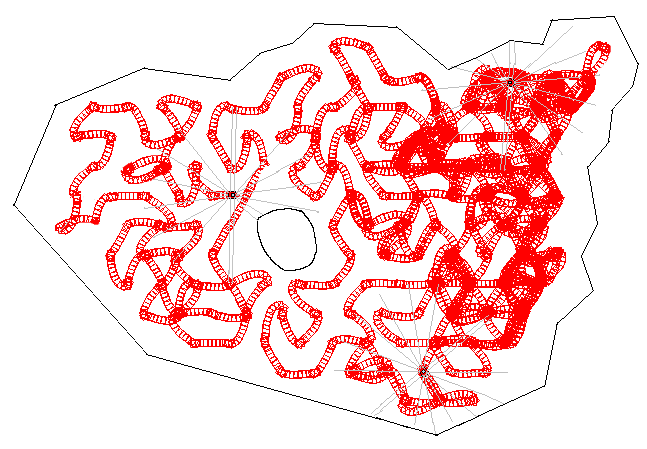}}}
\caption{Robots' trajectories after applying the second stage of the algorithm; (a) After 3m48s, (b) After 8m5s, (c) After 15m47s, (d) After 18m3s}
\label{SR:robotsTrajecs}
\end{figure*}

As mentioned  in Chapter \ref{chap:Consensus}, a two-stage algorithm is used to achieve the goal. First, the algorithm (\ref{CV:consAlg1}),(\ref{CV:consAlg2}) is applied which uses consensus variables in order to drive the robots to the vertices of a common triangular grid. In Fig. \ref{SR:robotsTrajecs}(a), the beginning of the robots' trajectories are the position of the robots after applying the first stage of the search algorithm, i.e., rule (\ref{CV:consAlg1}),(\ref{CV:consAlg2}) (see Fig. \ref{CV:stage1SimResult}).

The second stage of the algorithm, i.e., the algorithm \ref{SR:searchRule}, is applied whenever the first stage is completed. Fig. \ref{SR:robotsTrajecs} demonstrates the result of applying algorithm \ref{SR:searchRule} on a team of three robots. As seen in Fig. \ref{SR:robotsTrajecs}, the robots go through the vertices of the common triangular grid based on the proposed algorithm  until the whole area is explored by the robots. Fig. \ref{SR:robotsTrajecs}(a), Fig. \ref{SR:robotsTrajecs}(b) and Fig. \ref{SR:robotsTrajecs}(c) display the trajectories of the robots at times 3m48s, 8m5s and  15m47s after applying algorithm \ref{SR:searchRule}, respectively. Fig. \ref{SR:robotsTrajecs}(d) shows trajectories of the robots at time 18m3s when the search operation has been  completed. It is obvious that the area $\mathcal{W}$ is completely searched by the robots such that each vertex of the covering triangular grid is occupied at least one time by the robots. In this case study, we assume that the sides of the equilateral triangles are 2 meter (sensing range of the robots is $\frac{2}{\sqrt{3}}$ m) and the communication range between the robots is 10 meter. The area of the region $\mathcal{W}$ is about 528 $m^2$.

\begin{figure} [!ht]
\renewcommand{\topfraction}{1}
  \centering
  \includegraphics[width=13 cm, height=11 cm]{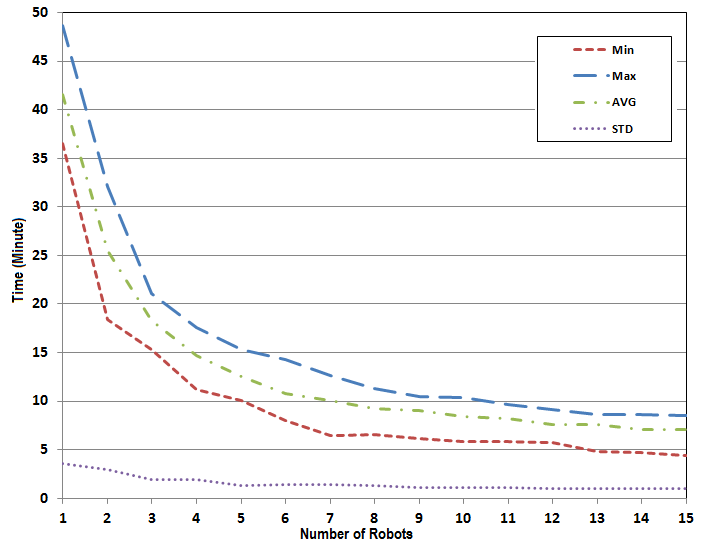}\\
  \caption{Duration of Search Vs. Number of Robots}\label{SR:plotOfTimeOfSearch}
\end{figure}

Although any number of robots can be used to search the whole area, it is obvious that more robots complete the operation in a shorter time. However, more number of robots certainly increases the cost of the operation. The question is, how many  robots should be used in order to optimize both the time and cost. It seems, that is somehow dependent on the shape of the region and the obstacles and also the sides of the triangles. In order to have a better view of the relation between the number of robots and the search duration, twenty  simulations with different number of robots have been done. We consider teams consisting of one to fifteen robots. Then, minimum, maximum, average and standard deviation are calculated for the search duration of each team.  Table \ref{SR:table1} displays the results of these simulations that are also depicted in Fig. \ref{SR:plotOfTimeOfSearch}.
\\

\begin{table}[!ht]\footnotesize
\centering
\caption{Duration of search}\label{SR:table1}
\resizebox{14cm}{!} {
\setlength{\extrarowheight}{18pt}
{\huge
\begin{tabular}{|l|l|l|l|l|l|l|l|l|l|l|l|l|l|l|l|}
\hline
\textbf{No of Robots} & \textbf{1} & \textbf{2} & \textbf{3} & \textbf{4} & \textbf{5} & \textbf{6} & \textbf{7} & \textbf{8} & \textbf{9} & \textbf{10}& \textbf{11} & \textbf{12} & \textbf{13} & \textbf{14} & \textbf{15}  \\ \hline
\textbf{Min}  & 36.51 & 18.40 & 15.28 & 11.21 & 10.05 & 8.03  & 6.47  & 6.56  & 6.13  & 5.82  & 5.81 & 5.74 & 4.80 & 4.69 & 4.37\\ \hline
\textbf{Max} & 48.62 & 32.05 & 21.06 & 17.56 & 15.36 & 14.25 & 12.61 & 11.33 & 10.52 & 10.37 & 9.65 & 9.11 & 8.59 & 8.62 & 8.51\\ \hline
\textbf{Average}   & 41.57 & 25.52 & 18.29 & 14.69 & 12.56 & 10.81 & 10.02 & 9.24  & 9.05  & 8.43  & 8.21 & 7.55 & 7.56 & 7.10 & 7.03 \\ \hline
\textbf{STD} & 3.53  & 2.95  & 1.93  & 1.92  & 1.29  & 1.46  & 1.46  & 1.34  & 1.10  & 1.09  & 1.06 & 1.01 & 1.02 & 0.97 & 0.96\\ \hline
\end{tabular}
}
}
\end{table}

 As depicted in this figure, the search time decreases by increasing the number of robots. It is noticeable that  after increasing the  number of robots to a specific number, the search duration almost remains constant. Indeed, increasing the number of robots   increases the probable collision between robots during the operation. Consequently, the robots have to turn each other to avoid the collision, and that is  the main reason which increases the search time. Therefore, increasing the number of robots more than a particular value (seven robots in this case) will be ineffectual in terms of time and should be avoided to save cost.  As apparent from Fig. \ref{SR:plotOfTimeOfSearch}, increasing the number of robots more than seven improves the search time only about three minutes.

\subsection{Searching for Targets}

\begin{figure*}[!htp]
\centering
\mbox{\subfigure[]{\includegraphics[width=6.5 cm, height=5 cm]{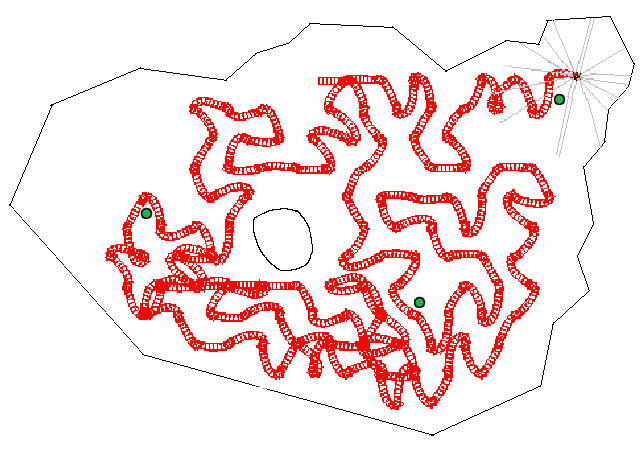}}\quad
\subfigure[]{\includegraphics[width=6.5 cm, height=5 cm]{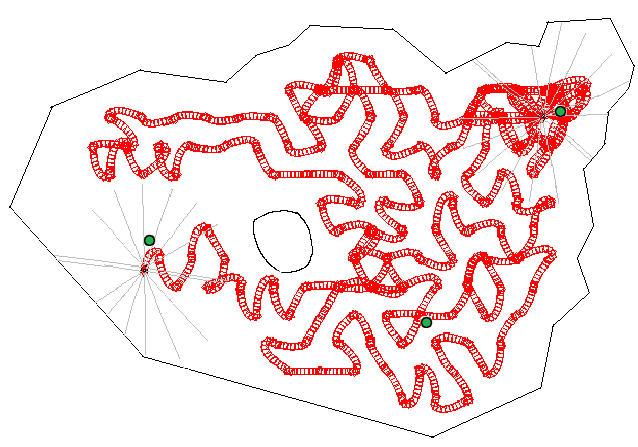}}\quad}
\mbox{\subfigure[]{\includegraphics[width=6.5 cm, height=5 cm]{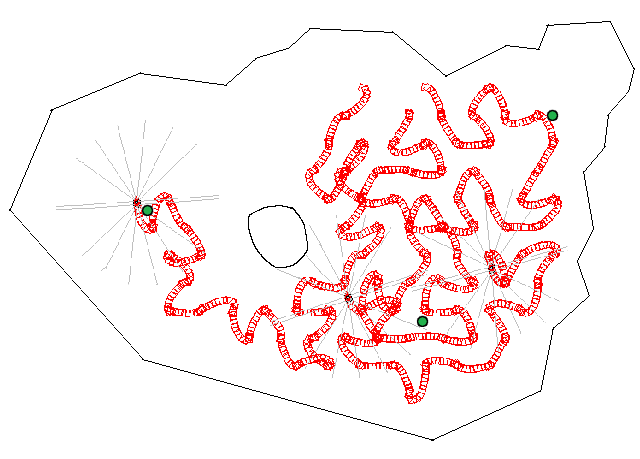}}\quad
\subfigure[]{\includegraphics[width=6.5 cm, height=5 cm]{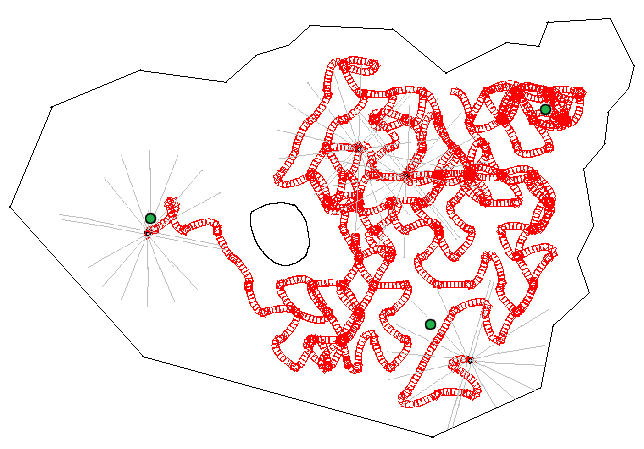}}\quad}
\mbox{\subfigure[]{\includegraphics[width=6.5 cm, height=5 cm]{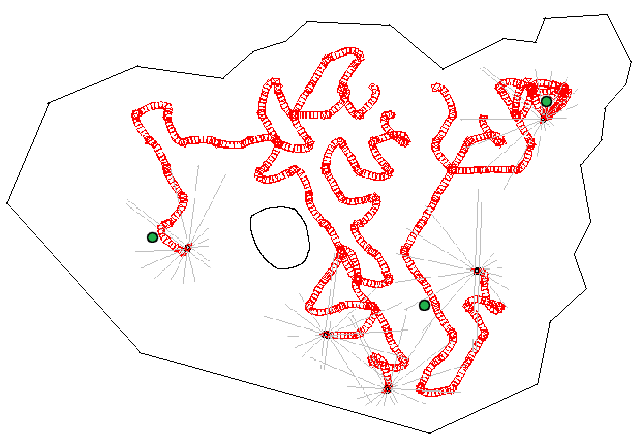}}\quad
\subfigure[]{\includegraphics[width=6.5 cm, height=5 cm]{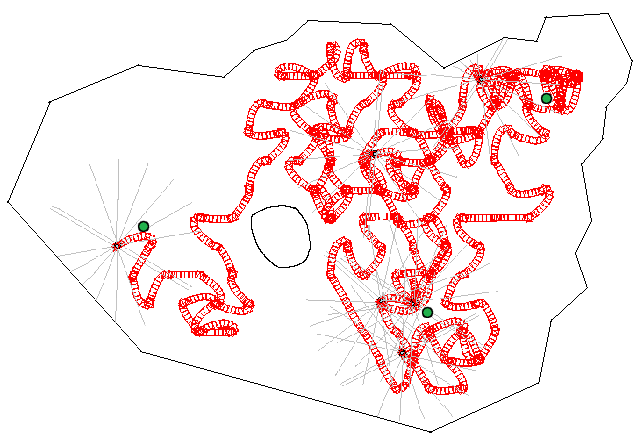}}}
\caption{Robots' trajectories in the case of searching for three targets. (a) One robot detects three targets in 17m13s, (b) Two robots detect the targets in 12m24s, (c) Three robots detect targets in 5m49s, (d) Four robots detect the targets in 7m33s, (e) Five robots detect the targets in 4m8s and (f) Six robots detect the targets in 5m19s. }
 \label{SR:findTargets}
\end{figure*}

To demonstrate that how the algorithm works in the case of search for a given number of targets, we consider an area the same as in section \ref{SR:SearchingtheWholeareaSimulation} with three targets therein. The targets, shown by green disks in Fig. \ref{SR:findTargets}, are located in different parts of the area. We evaluate the presented algorithm to figure out  how the robots can find the targets using multi-robot teams with different number of robots. As depicted in Fig. \ref{SR:findTargets}, the results of simulation by  the teams consisting of one to six robots have been presented. A target is assumed detected whenever it lies in the sensing range of a robot of the team. If there is just one robot in the team, all the targets should be detected by that robot; thus, it will take longer time in compare with the cases that there are more robots in the team.

As shown in Fig. \ref{SR:findTargets}(a), a robot searches for three targets using the presented search algorithm and they all have been detected after 17m13s. Fig. \ref{SR:findTargets}(b) shows the same case with  two robots in the team so that a robot has detected one target and the other one has detected two other targets in 12m24s. It should be mentioned that in Fig. \ref{SR:findTargets}, only the paths of the robots after making consensus have been displayed. In  Fig. \ref{SR:findTargets}(c)  three robots search for three targets and each robot finds one of them in 5m49s. As shown in  that figure, when a robot detects a target, it continues the search operation until all targets are detected by the team. Fig. \ref{SR:findTargets}(d)-(f)  show the search operation using 4-6 robots, respectively.

What is very noticeable in this case is that, it is expected decreasing the time of detecting the targets by increasing the number of robots while it has not occurred in some instances. For example, when the number of robots has been increased from three to four, the time of detecting the targets has been increased from 5m49s to 7m33s. To explain  why this happens, we should consider the fact that the time needed to detect the targets depends on many parameters not only on the number of robots. The shape of the area and obstacles therein, the initial position of the robots in the area and also the relative distance of the robots and targets are  significant parameters that affect the time of search. The other serious parameter must be considered is the nature of the search algorithm that is semi-random. That is we might have different paths for the same cases.

\begin{figure}[!htp]
\centering
\subfigure[Targets are detected in 5m49s]{\includegraphics[width=8 cm, height=5.5 cm]{SemiRandomSearch/SRfindTarget3robots.png}}
\subfigure[Targets are detected in 4m33s ]{\includegraphics[width=8 cm, height=5.5 cm]{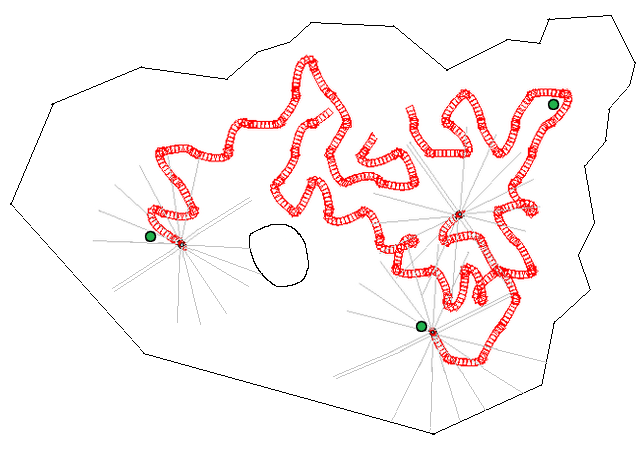}}
\subfigure[ Targets are detected in 14m21s]{\includegraphics[width=8 cm, height=5.5 cm]{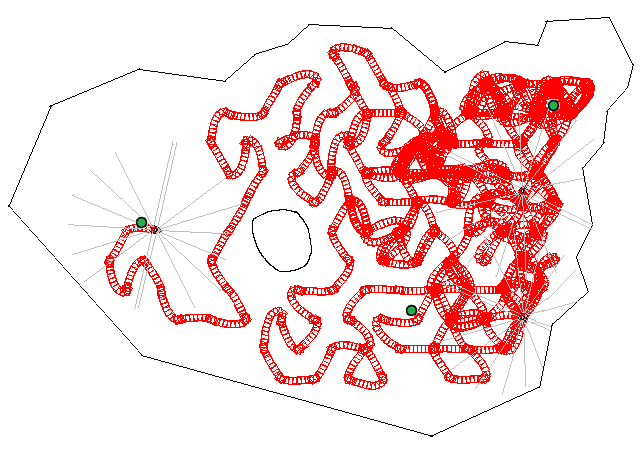}}
\caption{Robots' trajectories in the case of searching for three targets by three robots. } \label{SR:2robotPaths}
\end{figure}

\begin{figure}[!htp]
\centering
\subfigure[Targets are detected in 4m8s]{\includegraphics[width=8 cm, height=5.5 cm]{SemiRandomSearch/SRfindTarget5robots.png}}
\subfigure[Targets are detected in 4m37s ]{\includegraphics[width=8 cm, height=5.5 cm]{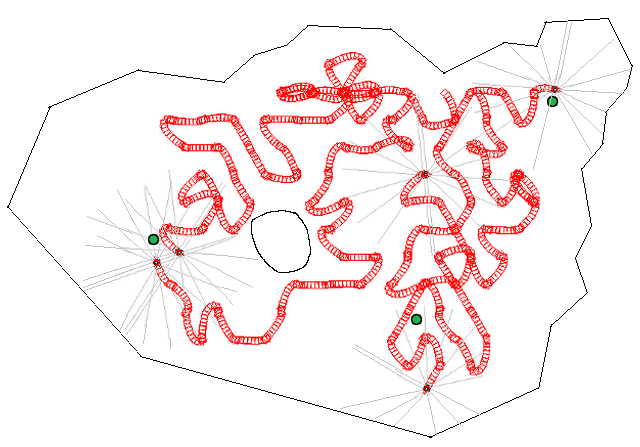}}
\subfigure[ Targets are detected in 7m33s]{\includegraphics[width=8 cm, height=5.5 cm]{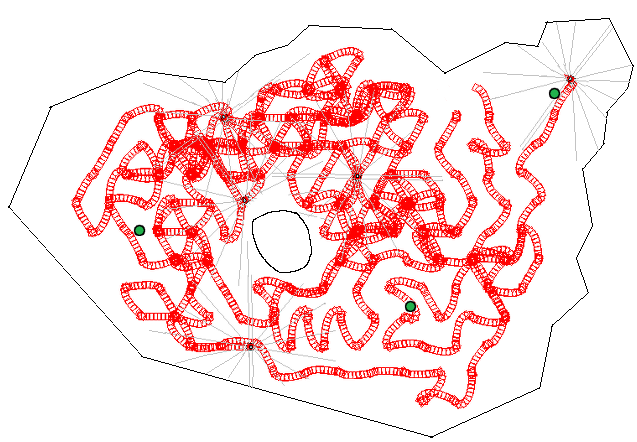}}
\caption{Robots' trajectories in the case of searching for three targets by five robots. } \label{SR:5robotPaths}
\end{figure}

To discover more about how the number of robots affects on the  search duration, we do more simulations for each case. For example, Fig. \ref{SR:2robotPaths} shows the paths of the robots in the case that three robots are looking for three targets similar to the previous instance. It shows that we have different paths thus different  search durations. While in the first simulation, the targets are detected in 5m49s, it takes 4m33s in the second and 14m21s in the last one, which is much more than the second one. Fig. \ref{SR:5robotPaths} shows the results of three different simulations using five robots. As depicted in that figure, the period of search for these simulations are 4m8s, 4m37s and 7m33s. Comparing to the case with three robots, it is obvious that the differences between the periods of search in this instance are less than the case of the team with three robots. That is an expected result because more robots means more coverage of the search area; hence, the chance of detecting  targets increases.

\begin{table}[!ht]
\centering
\caption{Time of detecting targets }\label{SR:table2}
\resizebox{13cm}{!} {
\setlength{\extrarowheight}{6pt}
\begin{tabular}{|l|l|l|l|l|l|l|l|l|l|l|}
\hline
\textbf{No of Robots} & \textbf{1} & \textbf{2} & \textbf{3} & \textbf{4} & \textbf{5} & \textbf{6} & \textbf{7} & \textbf{8} & \textbf{9} & \textbf{10} \\ \hline
\textbf{Min}  & 13.45 & 7.26  & 4.13  & 3.59  & 2.72 & 2.49 & 2.07 & 2.01 & 1.95 & 1.72 \\ \hline
\textbf{Max} & 29.50 & 19.95 & 14.35 & 10.70 & 7.55 & 6.08 & 5.23 & 5.15 & 4.97 & 4.95 \\ \hline
\textbf{Average}  & 21.99 & 11.36 & 8.41  & 6.95  & 4.44 & 3.96 & 3.46 & 3.30 & 3.02 & 3.13 \\ \hline
\textbf{STD} & 5.79  & 3.11  & 2.34  & 1.96  & 1.12 & 1.02 & 0.85 & 0.94 & 0.77 & 0.82 \\ \hline
\end{tabular}
}
\end{table}

\begin{figure} [!ht]
\renewcommand{\topfraction}{.85}
  \centering
  \includegraphics[width=13 cm, height=11 cm]{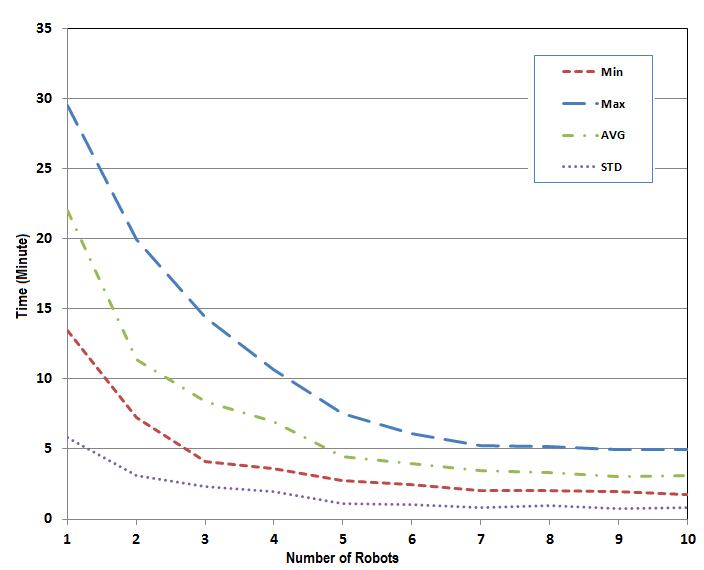}\\
  \caption{Time of target detection Vs. Number of robots}\label{SR:plotOfTimeOfDetect}
\end{figure}

In order to have a better view of the relation between the number of robots and the search duration, twenty  simulations with different number of robots have been done. We consider teams consisting of one to ten robots. Then, minimum, maximum and average of  search duration for simulations with each team as well as their standard deviation are calculated. Table \ref{SR:table2} displays the results of these simulations that are also depicted in Fig. \ref{SR:plotOfTimeOfDetect}. The results show that although the average of search duration to detect the targets decreases by increasing the number of robots, the standard deviation of the team with less number of robots is significantly high. Simply, the number of robots should be proportional to the search area to have an adequate chance to detect targets in a desirable time.
\\

\section{Summary}
In this chapter, we have developed a distributed control algorithm, namely,  semi-random triangular grid-based search algorithm, to drive a multi-robot team to explore an unknown area. We have used a triangular grid pattern and a two-stage semi-random  search algorithm for the control law so that the robots randomly move through the vertices of the triangular grid during the search operation. Therefore, a complete search of the whole area has been guaranteed. A mathematically rigorous proof of convergence of the presented algorithm has been demonstrated. Furthermore, the computer simulation results using MobileSim, a powerful simulator of real robots and environment, have been presented to demonstrate that the algorithm is effective and practicable.

\chapter{Modified Triangular-Grid-Based Search Algorithm}\label{chap:ModSearch}

In Chapters \ref{chap:RandomSearch} and \ref{chap:SemiRandomSearch}, two methods for exploring an unknown environment using a team of mobile robots were presented. In those methods, the robots randomly move between the vertices of a common triangular grid so that in each step they only move to the one of the neighbouring vertices. In this  chapter, we present the third method of search, namely, "modified triangular grid-based search algorithm". By this method, the robots are not confined to move to the closest neighbouring vertices. Instead, they move to the nearest unvisited vertex. A mathematically rigorous proof of convergence with probability 1 of the algorithm is given. Moreover, our algorithm is implemented and simulated using a simulator of the real robots and environment and also tested via experiments with Adept Pioneer 3DX wheeled mobile robots. Finally, a comparison between our three proposed algorithms and three  algorithms from other researchers is given.

\section{Distributed Triangular Grid-Based Search Algorithm}

To search an area by a team of mobile robots using the vertices of a grid as the exploring points, we need to locate the searching mobile robots on the vertices of a common grid among the robots. Consensus variables locating algorithm (described in Chapter \ref{chap:Consensus}) can be the first stage of the suggested search algorithm, which  locates` all the robots on the vertices of a common triangular grid  $\hat{\mathcal{T}}$ (see Fig. \ref{CV:area}). The next step will be  search of the area $\mathcal{W}$ based on moving the robots between the vertices of the covering grid of the area. In this chapter, we propose a triangular grid-based search algorithm. Suppose a robot located on a vertex of the common triangular grid.  Consequently, it can explore the surrounding  area using its sensors, and that depends on the sensing range of its sensors. After exploring that area,  the robot moves to another point which can be one of the unexplored vertices in the triangular grid. As a modified version of the previous methods presented in Chapters \ref{chap:RandomSearch} and \ref{chap:SemiRandomSearch}, we suppose that a robots selects the nearest unvisited vertex as the next destination in every step.  If there is more than one vertex, any of them can be chosen randomly. In this regard, there are various scenarios  can be considered to search the area. The first scenario is exploring the whole area which  can be applicable when the robots are searching for an undetermined number of targets.  Therefore, to detect  all possible targets, the team of the robots must search the whole area. Patrolling of the area is the  other application for this scenario where the robots should move continuously  to detect the possible intruders to the area. In the case of given number of targets which is our second scenario, the search operation should be stopped without searching  the whole area; whenever all the targets are detected.

\subsection{Searching the Whole Area}\label{MS:SearchingtheWholeArea}
To make sure that the whole area is explored by the team of the robots, each vertex in  the triangular covering grid set of the area  $\mathcal{W}$ must be visited at least one time by a member of the team. Consider  $\hat{\mathcal{T}}$ is a triangular covering grid of  $\mathcal{W}$, and also each vertex of $\hat{\mathcal{T}}$ has been  visited at least one time by a robot of the team. This guarantees that the area $\mathcal{W}$ has completely been explored by the multi-robot team. Since the robots do not have any map at the beginning, they need to do map making during the search operation so that their maps will gradually be completed.

\begin{mydef}\label{MS:d8}
Let  $\hat{\mathcal{T}_i}(k)$ be the set of all the vertices of $\hat{\mathcal{T}}$  have been detected by robot $i$ at time $k$. Then,  $\hat{\mathcal{T}}(k)=\bigcup\hat{\mathcal{T}_i}(k)$ will be the map of the  area $\mathcal{W}$ detected by the team of the robots until time $k$.
\end{mydef}
Note that a detected vertex is different from an explored vertex. These terms are defined in detail in the  following definitions.
\begin{mydef}\label{MS:d9a}
A detected vertex means  that vertex is detected by a robot using map making, and it is in the map of that robot though it might be  visited or not  by the robots.
\end{mydef}
\begin{mydef}\label{MS:d9b}
An explored vertex is a vertex that is visited by at least one member of the team.
\end{mydef}
\begin{mydef}\label{MS:d10}
The map of robot $i$ at time $k$, ${\mathcal{M}_i}(k)$, is the set of those vertices in  $\hat{\mathcal{T}}$ detected by robot $i$ itself or received from other robots by which  they are detected until time $k$.
\end{mydef}

\begin{mydef}\label{MS:d11}
Suppose  a Boolean variable $V_{\tau}(k)$ which defines the state of vertex  $ \tau\in \hat{\mathcal{T}}(k)$ at time $k$. $V_{\tau}(k)=1$ if the vertex $\tau$ has already been  visited  by at least one of the robots, otherwise $V_{\tau}(k)=0$.
\end{mydef}

\begin{myassump}\label{MS:a4}
The triangular grid set $\hat{\mathcal{T}_i}(k)$; $k=0,1,...$ is connected. That means if $ \tau\in \hat{\mathcal{T}_i}(k)$, then at least one of the six nearest neighbours of $\tau$ also belongs to $\hat{\mathcal{T}_i}(k)$.
\end{myassump}

Consider at time $k$ robot $i$ is located at point $p_{i}(k)$, and it wants to go to the next vertex. The following rule is proposed as the modified triangular grid-based search algorithm:

\begin{equation}\label{MS:eq4}
p_{i}(k+1)=
\begin{cases} \hat{\mathfrak{m}_i}(k) & \text{if } |\hat{\mathcal{M}_i}(k)|\neq0 \quad
\\
p_{i}(k) &\text{if } |\hat{\mathcal{M}_i}(k)|=0 \quad
\end{cases}
\end{equation}
\\
where $\hat{\mathcal{M}_i}(k)=\{\mathfrak{m}\in{\mathcal{M}_i}(k);V_{\mathfrak{m}}(k)=0\}$ is the set of all elements of ${\mathcal{M}_i}(k)$ have not been visited before, $|\hat{\mathcal{M}_i}(k)|$ denotes the number of elements in $\hat{\mathcal{M}_i}(k)$ and $\hat{\mathfrak{m}_i}(k)$ is the vertex in $\hat{\mathcal{M}_i}(k)$ nearest to robot $i$ at time $k$.

Applying the rule (\ref{MS:eq4}) ensures that the area $\hat{\mathcal{T}}$ is completely explored, and every vertex of it is visited at least one time by a robot of the team.

\begin{mytheorem}\label{MS:th2}
Suppose that all assumptions hold, and the mobile robots move according to the distributed control law (\ref{MS:eq4}). Then, for any number of robots, with probability 1 there exists a time $k_{0}\geq 0$ such that $V_{\tau}(k_{0})=1 ; \quad \forall  \tau\in \hat{\mathcal{T}}$.
\end{mytheorem}

{\bf Proof of Theorem \ref{MS:th2}}: The algorithm \ref{MS:eq4} defines an absorbing Markov chain which contains many transient states and a number of absorbing states  that are impossible to leave. Transient states are all the vertices of the triangular grid $\hat{\mathcal{T}}$ which have been occupied already by the robots during the search procedure. On the other hand, absorbing states are the vertices  where the robots stop at the end of the search operation. Using the algorithm \ref{MS:eq4}, a robot goes to the vertices where may have not been visited yet. Therefore, the number of transient states will eventually decrease. That continues until the number of robots is equal to the number of unvisited vertices which will be the absorbing states. It is also explicit  that these absorbing states can be reached from any initial states, with a non-zero probability. This implies that with probability 1, one of the absorbing states will be reached. This completes the proof of Theorem \ref{MS:th2}. $\square$

\begin{figure}[!hbt]
  \centering
  \includegraphics[width=11 cm, height=20 cm]{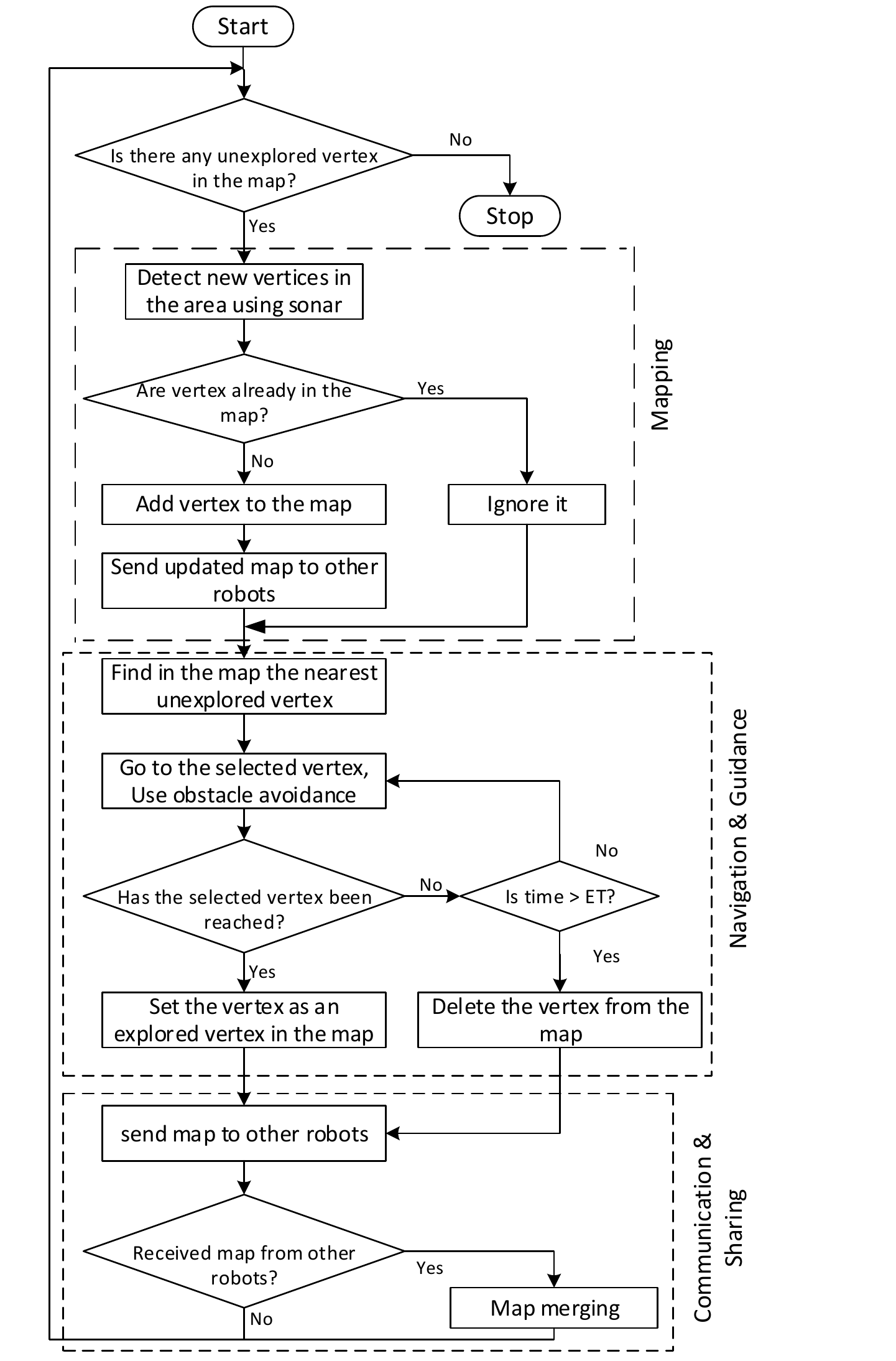}\\
  \caption{The flowchart of the suggested algorithm; searching the whole area}\label{MS:flowchart1}
\end{figure}

In Fig. \ref{MS:flowchart1}, the flowchart of the proposed algorithm is presented that  shows how our decision-making approach is implemented.  At the first step, robots start making their maps using their sonar. Each robot, based on the vertex on which it is located, assumes some probable neighbouring vertices on the common triangular grid. The number of these  probable neighbouring vertices and their distance to the robot depend on the robot's sonar range. Then, the robot uses the sonar to detect its surrounding environment including borders and obstacles. If any of those probable neighbouring vertices is located outside the borders or blocked by an obstacle, it will be ignored. The rest of those probable neighbouring vertices will be added to the map of the robot. This step is repeated every time that the robot occupies a vertex. In order to avoid sticking in borders or obstacles, we consider a margin near the borders and obstacles that depends on the size of the robot. If a vertex  on the map is closer to the borders or obstacles less than the margin, it will be eliminated from the robot's map.

 Whenever a robot makes any changes to its map, it sends the new map to the other robots by transmitting packets. On the other side, whenever a robot receives a packet of data, it extracts  the new vertices from the received map and adds them to its map. Since the communication range of the robots is limited, if two robots are far from each other, they cannot directly communicate but can do it via other robots. In other words, since there is a connected network of robots, each robot has the  role of a hub in the network in order to share the maps among the robots. Therefore, all the robots that are connected and make a network  have a common map.  In the second phase of the algorithm when the robots have a common triangular grid map, a robot can go far from the other robots and be disconnected from the team  for a while. In this case, sharing maps between the disconnected robot and the others is paused until the robot returns back to the communication range of the team again.

 In the next step, each robot chooses the nearest unexplored vertex in the map and goes there. Whenever a robot visits a vertex, it marks that as an explored vertex in its map. Therefore, all the robots know which vertices have not been explored yet. To find the nearest vertex that has not been  explored yet, each robot searches the map stored in its memory. This vertex has probably been  detected by the robot itself or added to robot's map via sharing map among robots. After finding the nearest unexplored vertex, the robot moves to reach  there. If the robot reaches the target vertex, it marks that vertex as an explored vertex in its map and sends it to the other neighbouring robots as well. However, because of some practical issues, maybe it is impossible to reach the target vertex at a limited time or even maybe the target vertex is fake  that has been wrongly created during mapping. To avoid such problems, we include a factor of time. Since the robot knows its location and the location of the target, it can estimate the time needed to achieve  the goal based on the distance to the target and velocity of the robot. Here, we consider the parameter ET as the expected time to reach the target, that is a factor of the estimated time. This coefficient with a value greater than one, actually reflects the effects of a non-straight route because of  the shape of the search area and also  existing obstacles or other robots on the robot's path.  If the traveling time  were more than ET, the robot would ignore that target vertex, delete it from its map, and send the updated map to the other robots.

 Since the area has borders and obstacles, the robots should avoid them while they are searching for  the targets. That is why we have to use an obstacle avoidance in our algorithm. In addition, because of practical problems like sonar and encoders accuracy and slipping of the robots, there  might be a difference between the actual position of a robot and the coordinates of the vertices stored in its map. Therefore, if a robot is closer to a target vertex than a specified distance, we consider the goal has been achieved.

\subsection{Searching for Targets}
When the robots are looking for some targets in the area, they should continue the search operation till all the targets are detected. If the number of the targets is not given, they have to search the entire area like what described in Section \ref{MS:SearchingtheWholeArea}. When robots know the number of the  targets, they do not need to explore the entire area but until all the  targets are detected.
Suppose $\mathfrak{T}=\{\mathfrak{T}_1,\mathfrak{T}_2,\dots, \mathfrak{T}_{n_t} \}$ be the set of $n_t$ static targets should be detected by the robots. As it was stated before, we assume the robots equipped with sensors by which the targets can be detected whenever they are close enough to the robots. The distance by which  the robots must be close to the targets to be able to detect them is the sensing range of the robots ($r_s)$.

\begin{mydef}\label{MS:d12}
Suppose  a Boolean variable $V_{\mathfrak{T}_j}(k)$ which defines the state of target  ${\mathfrak{T}_j}$ at time $k$.  $V_{\mathfrak{T}_j}(k)=1$  if the target   ${\mathfrak{T}_j}$ has  been detected by at least one of the robots, otherwise  ${\mathfrak{T}_j}(k)=0$.
\end{mydef}

 Therefore, we modify the rule (\ref{MS:eq4}) as the following rule to ensure that the search operation stops after finding all the targets.

\begin{equation}\label{MS:eq5}
p_{i}(k+1)=
\begin{cases} \hat{\mathfrak{m}_i}(k) & \text{if } \exists {\mathfrak{T}_j} \in \mathfrak{T} ; V_{\mathfrak{T}_j}(k)=0 \quad
\\
p_{i}(k) &\text{if } \forall {\mathfrak{T}_j} \in \mathfrak{T} ; V_{\mathfrak{T}_j}(k)=1 \quad
\end{cases}
\end{equation}

\begin{mytheorem}\label{MS:th3}
Suppose that all assumptions hold, and the mobile robots move according to  distributed control law (\ref{MS:eq5}). Then, for any number of robots and any number of targets, with probability 1 there exists a time $k_{0}\geq 0$ such that $\forall j$ ; $V_{\mathfrak{T}_j}(k_0)=1$.
\end{mytheorem}

{\bf Proof}: Proof is similar to the  proof of Theorem \ref{MS:th2}.

\begin{figure}[!hbt] 
  \centering
  \includegraphics[width=8 cm, height=9 cm]{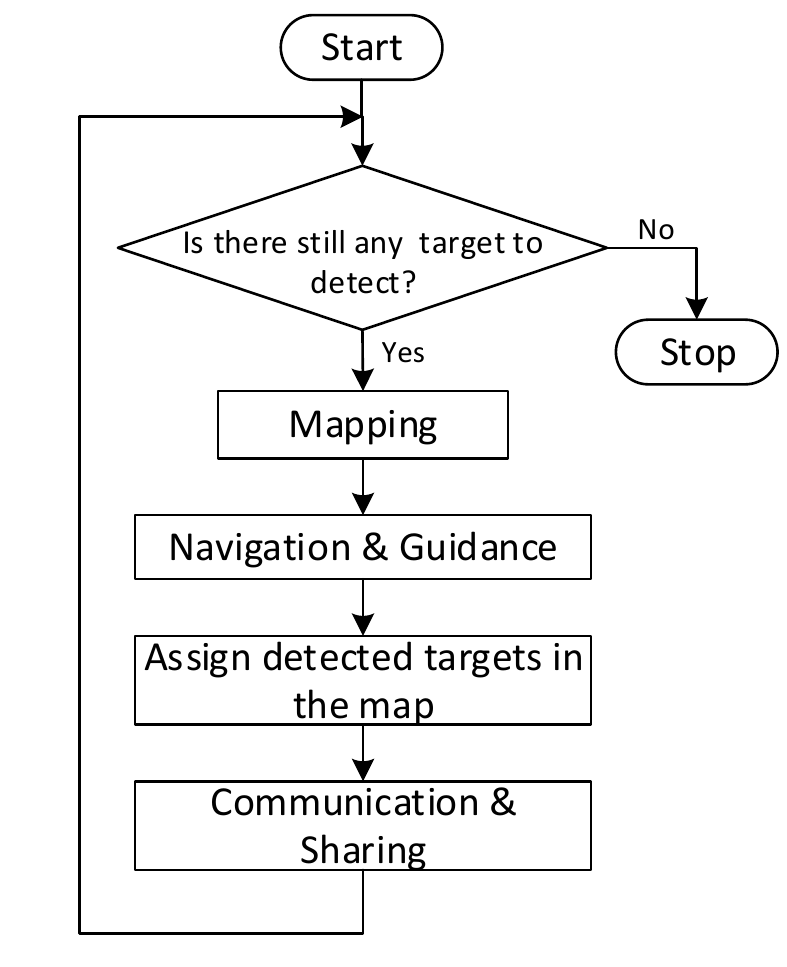}\\
  \caption{The flowchart of the suggested algorithm; searching for targets}\label{MS:flowchart2}
\end{figure}

The flowchart in Fig. \ref{MS:flowchart1} can also be used to describe this operation. Fig. \ref{MS:flowchart2} depicts the procedure we use to implement this algorithm. Most procedures are the same as the previous one; therefore, we ignore their description. We only need to change the condition that stops the operation in the algorithm. The operation will be stopped when all the targets are detected. Also, the robots send  the information about the detected targets  to the other members of the team.

\subsection{Patrolling}

 The above algorithms are appropriate for the cases when the robots should break the search operation; for instance, when the aim is finding some predetermined objects or labeling  some vertices of a  grid. In cases where a permanent search is needed; for example, in applications such as continuously patrolling or surveying a region, the algorithm should be modified such that the search procedure maintains. That can be done by periodically changing the states of the vertices in the maps of the robots; hence, the robots consider those vertices as the targets again. To archive that, the state of the previously visited vertices should be changed  to unvisited in the map of the robots. Based on  some conditions such as the size of the area and the number of the robots, the period of changing the state of vertices can be chosen.

\subsection{Robots' Motion}
To have a more real estimation of time that the robots spend to search an area to detect the targets, we apply motion dynamics in our simulations. In addition, since the environment is unknown to the team, it is essential to use an obstacle avoidance in the low-level control of the robots. As a result, we use a method of reactive potential field control in order to avoid obstacles \cite{khatib1986real}.

\section{Simulation Results}

To verify the suggested algorithm, computer simulations are employed. The region $\mathcal{W}$ is considered to be searched by a few robots (see Fig. \ref{CV:area}). We  suppose  a multi-robot team of some autonomous mobile robots which are randomly located in the region $\mathcal{W}$ with random initial values of angles. The goal is to search the whole area  by the robots using proposed grid-based search algorithm.

To simulate the algorithm, MobileSime, a simulator of mobile robots developed by  Adept MobileRobots, is used. We also use  Visual C++ for programming and ARIA, a C++ library that provides an interface and framework for controlling the robots. In addition,  Pioneer 3DX is selected as type of the robots. Robots’ parameters in the simulations  are given in Table \ref{CV:simParameters}. Since, this simulator simulates the real robots along with all conditions of a real world, the results of the simulations would be obtained in the real world experiments with the real robots indeed. Furthermore, to prevent collisions between robots and to avoid the obstacles and borders,  an obstacle avoidance algorithm is applied using functions provided in ARIA library. Moreover, to avoid hitting and sticking to the borders, we assume a margin near the borders such that the robots do not pass it.

\begin{figure*}[!hbt]
\centering
\mbox{\subfigure[]{\includegraphics[width=7 cm, height=5 cm]{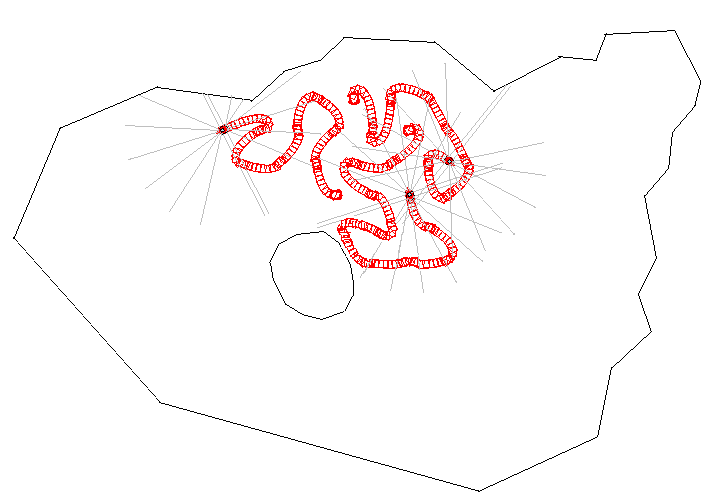}}\quad
\subfigure[]{\includegraphics[width=7 cm, height=5 cm]{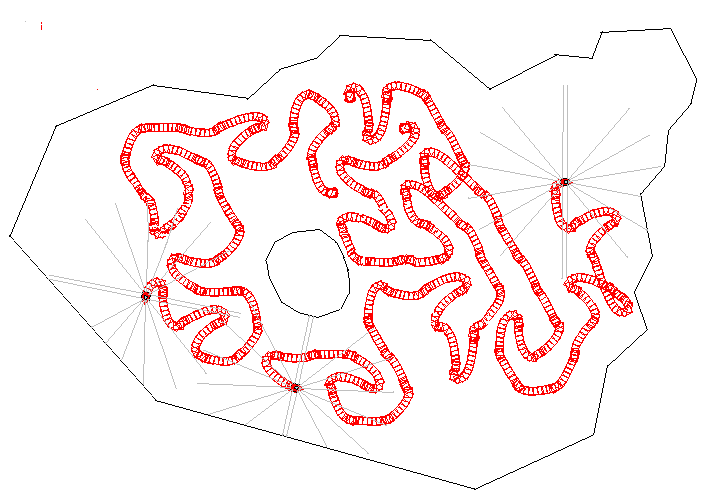}}}
\mbox{\subfigure[]{\includegraphics[width=7 cm, height=5 cm]{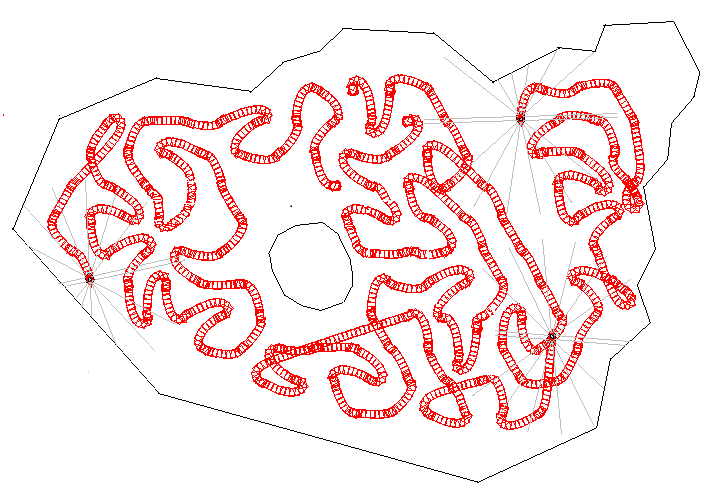}}\quad
\subfigure[]{\includegraphics[width=7 cm, height=5 cm]{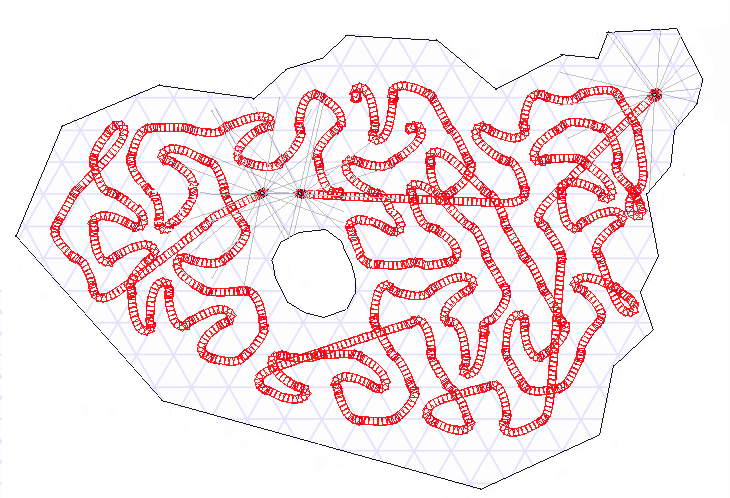}}}
\caption{Robots' trajectories after applying the second stage of the algorithm; (a) After 3m21s, (b) After 5m45s, (c) After 7m26s, (d) After 8m23s}
\label{MS:robotsTrajecs}
\end{figure*}

\subsection{Searching the Whole Area}\label{MS:SearchingtheWholeAreaSimulation}
As mentioned  in Chapter \ref{chap:Consensus}, a two-stage algorithm is used to achieve the goal. First, the algorithm (\ref{CV:consAlg1}),(\ref{CV:consAlg2}) is applied which uses consensus variables in order to drive the robots to the vertices of a common triangular grid. In Fig. \ref{MS:robotsTrajecs}(a), the beginning of the robots' trajectories are the position of the robots after applying the first stage of the search algorithm, i.e., rule (\ref{CV:consAlg1}),(\ref{CV:consAlg2}) (see Fig. \ref{CV:stage1SimResult}).

The second stage of the algorithm, i.e., the algorithm \ref{MS:eq4}, is applied whenever the first stage is completed. Fig. \ref{MS:robotsTrajecs} demonstrates the result of applying algorithm \ref{MS:eq4} on a team of three robots. As seen in Fig. \ref{MS:robotsTrajecs}, the robots go through the vertices of the common triangular grid based on the proposed algorithm  until the whole area is explored by the robots. Fig. \ref{MS:robotsTrajecs}(a), Fig. \ref{MS:robotsTrajecs}(b) and Fig. \ref{MS:robotsTrajecs}(c) display the trajectories of the robots at times 3m21s, 5m45s and 7m26s after applying algorithm \ref{MS:eq4}, respectively. Fig. \ref{MS:robotsTrajecs}(d) shows trajectories of the robots at time 8m23s when the search operation has been  completed. It is obvious that the area $\mathcal{W}$ is completely explored by the robots such that each vertex of the covering triangular grid is occupied at least one time by the robots. In this case study, we assume that the sides of the equilateral triangles are 2 meter (sensing range of the robots is $\frac{2}{\sqrt{3}}$ m) and the communication range between the robots is 10 meter. The area of the region $\mathcal{W}$ is about 528 $m^2$.

\begin{figure} [!hbt]
  \centering
  \includegraphics[width=9 cm, height=6.5 cm]{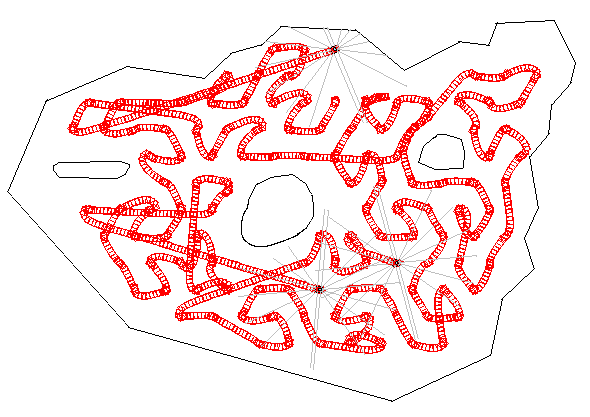}\\
  \caption{Searching an area with three obstacles}\label{MS:3robots3obstacles}
\end{figure}

Fig. \ref{MS:3robots3obstacles} shows another simulation result but for an area with three obstacles with different shapes. It shows that the algorithm indeed works with any number of obstacles.

\begin{figure} [!hbt]
\renewcommand{\topfraction}{.85}
  \centering
  \includegraphics[width=13 cm, height=11 cm]{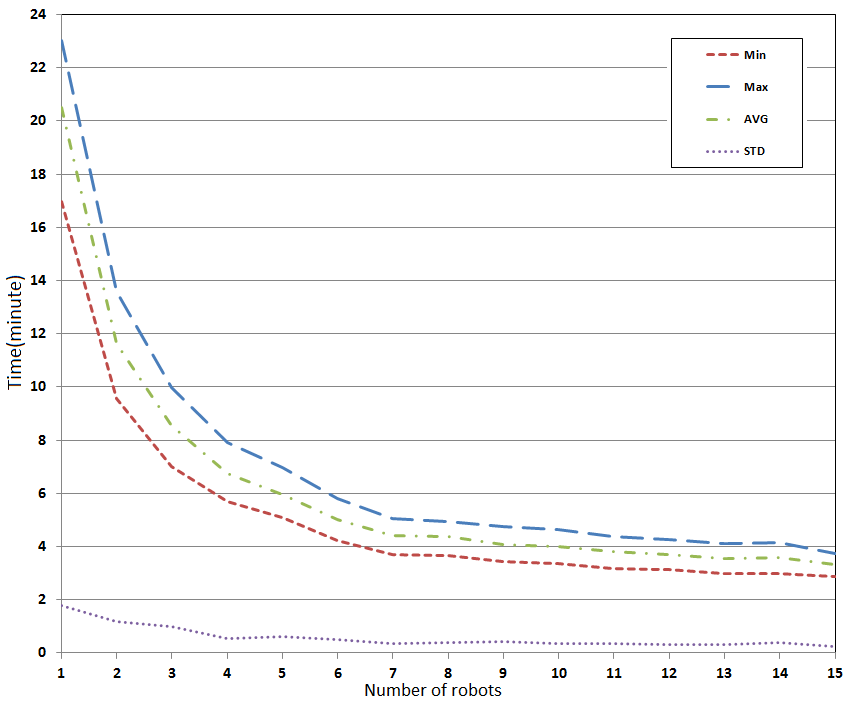}\\
  \caption{Duration of Search Vs. Number of Robots}\label{MS:plotOfTimeOfSearch}
\end{figure}

Although any number of robots can be used for the search operation, it is obvious that more robots complete the operation in a shorter time. However, more robots certainly increases the cost of the operation. The question is, how many number of robots should be used in order to optimize both the time and cost. It seems, that may be somehow dependent on the shape of the region and the obstacles and also the sides of the triangles. In order to have a better view of the relation between the number of robots and the search duration, twenty  simulations with different number of robots have been done. We consider teams consisting of one to fifteen robots. Then, minimum, maximum, average and standard deviation are calculated for the search duration of each team.  Table \ref{MS:table1} displays the results of these simulations that are also depicted in Fig. \ref{MS:plotOfTimeOfSearch}.

\begin{table}[!hbt]\footnotesize
\centering
\caption{Duration of search}\label{MS:table1}
\resizebox{14cm}{!} {
\setlength{\extrarowheight}{18pt}
{\huge
\begin{tabular}{|l|l|l|l|l|l|l|l|l|l|l|l|l|l|l|l|}
\hline
\textbf{No of Robots} & \textbf{1} & \textbf{2} & \textbf{3} & \textbf{4} & \textbf{5} & \textbf{6} & \textbf{7} & \textbf{8} & \textbf{9} & \textbf{10}& \textbf{11} & \textbf{12} & \textbf{13} & \textbf{14} & \textbf{15}  \\ \hline
\textbf{Min} & 16.98 & 9.55  & 7.03 & 5.68 & 5.09 & 4.23 & 3.72 & 3.66 & 3.44 & 3.38 & 3.18 & 3.12 & 2.98 & 3.01 & 2.88 \\ \hline
\textbf{Max}  & 23.02 & 13.61 & 9.98 & 7.90 & 6.98 & 5.82 & 5.05 & 4.96 & 4.77 & 4.65 & 4.38 & 4.26 & 4.12 & 4.17 & 3.76 \\ \hline
\textbf{Average} & 20.51 & 11.62 & 8.56 & 6.73 & 5.97 & 5.00 & 4.42 & 4.37 & 4.09 & 4.00 & 3.83 & 3.71 & 3.56 & 3.57 & 3.31 \\ \hline
\textbf{STD} & 1.77  & 1.20  & 0.98 & 0.54 & 0.62 & 0.52 & 0.37 & 0.40 & 0.42 & 0.36 & 0.35 & 0.33 & 0.34 & 0.41 & 0.25 \\ \hline
\end{tabular}
}
}
\end{table}

\begin{figure*}[!hbtp]
\centering
\mbox{\subfigure[]{\includegraphics[width=6.5 cm, height=5 cm]{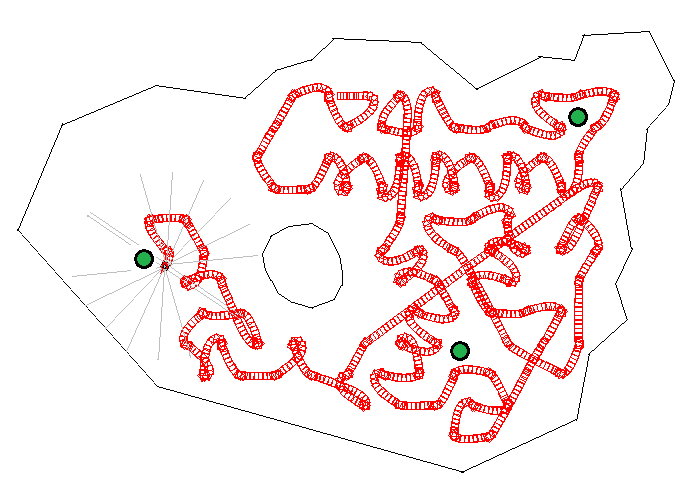}}\quad
\subfigure[]{\includegraphics[width=6.5 cm, height=5 cm]{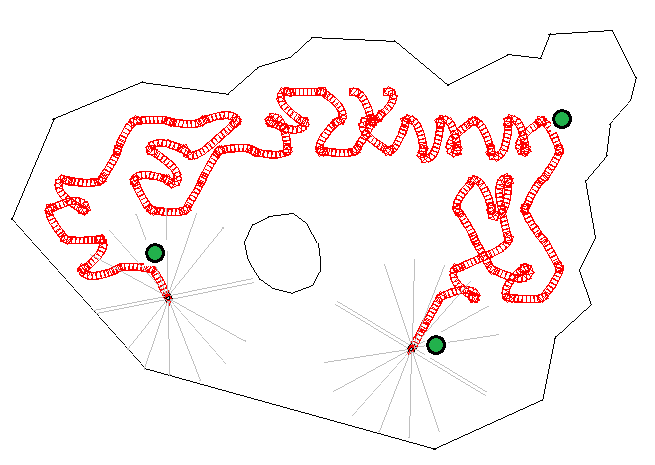}}\quad}
\mbox{\subfigure[]{\includegraphics[width=6.5 cm, height=5 cm]{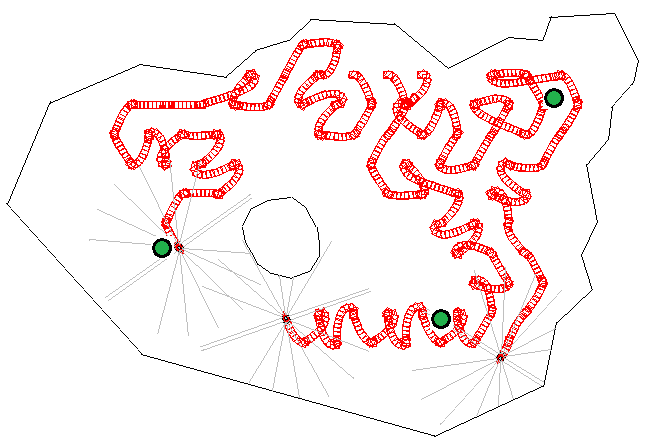}}\quad
\subfigure[]{\includegraphics[width=6.5 cm, height=5 cm]{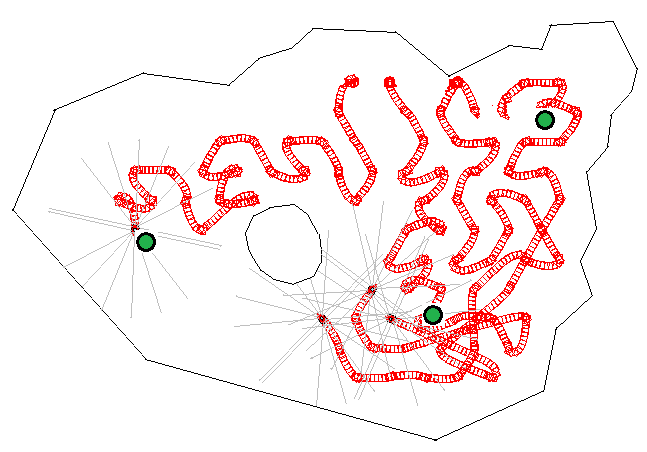}}\quad}
\mbox{\subfigure[]{\includegraphics[width=6.5 cm, height=5 cm]{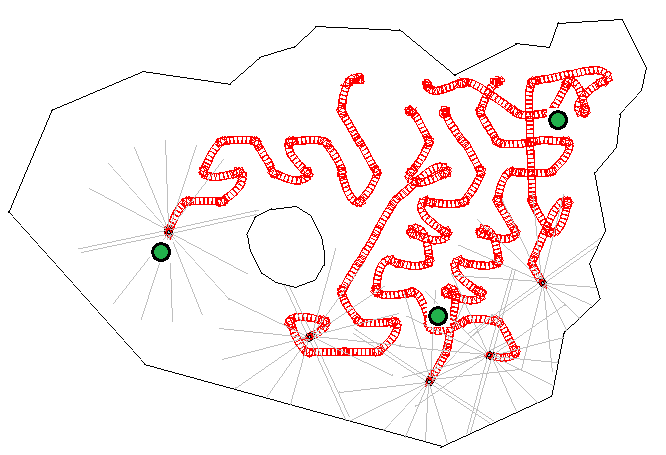}}\quad
\subfigure[]{\includegraphics[width=6.5 cm, height=5 cm]{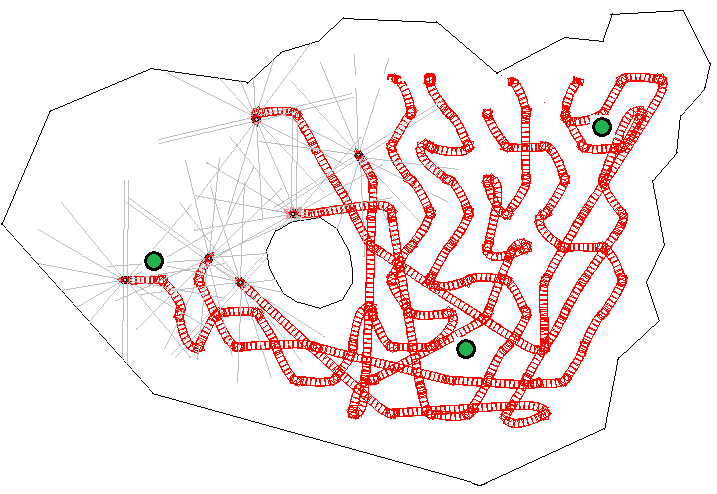}}}
\caption{Robots' trajectories in the case of searching for three targets. (a) One robot detects three targets in 12m49s, (b) Two robots detect the targets in 4m3s, (c) Three robots detect targets in 3m32s, (d) Four robots detect the targets in 3m50s, (e) Five robots detect the targets in 1m51s and (f) Six robots detect the targets in 2m15s. }
 \label{MS:findTargets}
\end{figure*}

 As depicted in this figure, the search time decreases by increasing the number of robots. It is noticeable that  after increasing the  number of robots to a specific number, the search duration almost remains constant. Indeed, increasing the number of robots   increases the probable collision between robots during the operation. Consequently, robots have to turn each other to avoid the collision, and that is  the main reason which increases the search time. Therefore, increasing the number of robots more than a specific value (six robots in this case) will be ineffectual in terms of time and should be avoided to save cost.  As obvious from Fig. \ref{MS:plotOfTimeOfSearch}, increasing the number of robots more than six, improves the search time less than one minute.

\subsection{Searching for Targets}

To demonstrate that how the algorithm works in the case of search for a given number of targets, we consider an area the same as in section \ref{MS:SearchingtheWholeAreaSimulation} with three targets therein. The targets, shown by green disks in Fig. \ref{MS:findTargets}, are located in different parts of the area. We evaluate the presented algorithm to figure out  how the robots can find the targets using multi-robot teams with different number of robots. As depicted in Fig. \ref{MS:findTargets}, the results of simulation by  the teams consisting of one to six robots have been presented. A target is assumed detected whenever it lies in the sensing range of a robot of the team. If there is just one robot in the team, all the targets should be detected by that robot;  thus, it will take longer time in compare with the cases that there are more robots in the team.

As shown in Fig. \ref{MS:findTargets}(a), a robot searches for three targets using the presented search algorithm and they all have been detected after 12m49s. Fig. \ref{MS:findTargets}(b) shows the same case with  two robots in the team so that a robot has detected one target and the other one has detected two other targets in 4m3s. It should be mentioned that in Fig. \ref{MS:findTargets}, only the paths of the robots after making consensus have been displayed. In  Fig. \ref{MS:findTargets}(c),  three robots search for three targets and each robot finds one of them in 3m32s. As shown in  that figure, when a robot detects a target, it continues the search operation until all the targets are detected by the team. Fig. \ref{MS:findTargets}(d)-(f)  show the search operation using 4-6 robots, respectively.

What is very noticeable in this case is that it is expected decreasing the time of detecting the targets by increasing the number of robots while it has not occurred in some instances. For example, when the number of robots has been increased from three to four, the time of detecting the targets has been increased from 3m32s to 3m50s. To explain  why this happens, we should consider the fact that the time needed to detect the targets depends on many parameters not only on the number of robots. The shape of the area and obstacles therein, the initial position of the robots in the area and also the relative distance of the robots and targets are  significant parameters that affect the time of search. The other serious parameter must be considered is the nature of the search algorithm that is semi-random. Indeed, a robot always chooses the nearest unexplored vertex as its destination vertex, but sometimes there are a few vertices which can be chosen as the nearest, and the robot randomly selects one of them. That is we might have different paths for the same cases.

\begin{figure}[!hbtp]
\centering
\subfigure[Targets are detected in 4m3s]{\includegraphics[width=7 cm, height=5.5 cm]{ModifiedSearch/findTarget2robots.png}}
\subfigure[Targets are detected in 6m44s ]{\includegraphics[width=7 cm, height=5.5 cm]{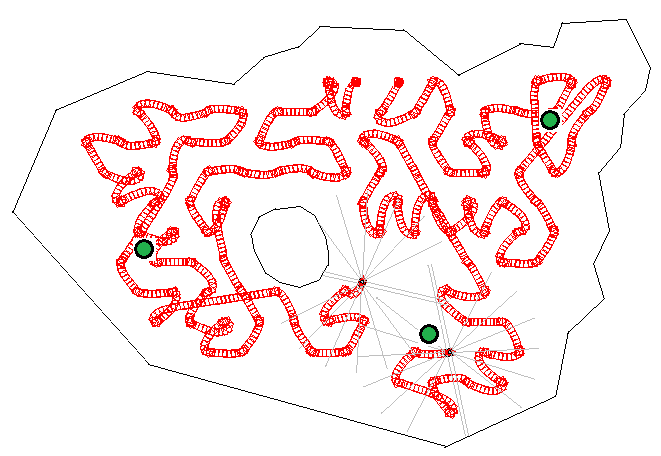}}
\subfigure[ Targets are detected in 7m41s]{\includegraphics[width=7 cm, height=5.5 cm]{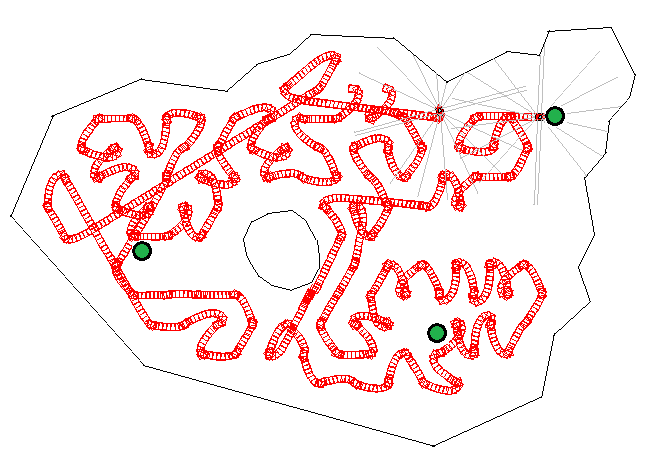}}
\caption{Robots' trajectories in the case of searching for three targets by two robots. } \label{MS:2robotPaths}
\end{figure}

\begin{figure}[!hbtp]
\centering
\subfigure[Targets are detected in 1m51s]{\includegraphics[width=7 cm, height=5.5 cm]{ModifiedSearch/findTarget5robots.png}}
\subfigure[Targets are detected in 1m58s ]{\includegraphics[width=7 cm, height=5.5 cm]{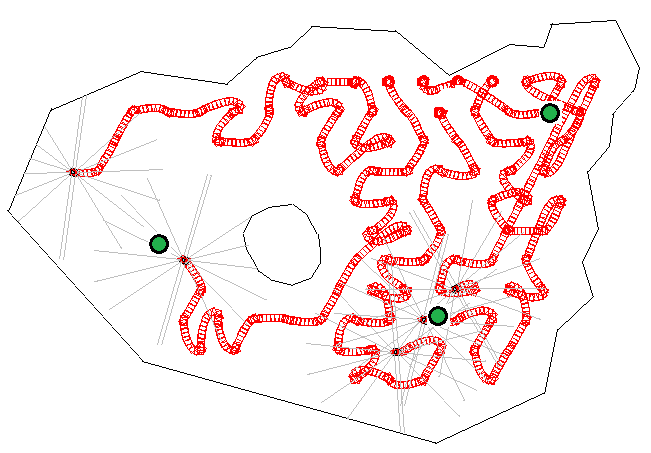}}
\subfigure[ Targets are detected in 2m33s]{\includegraphics[width=7 cm, height=5.5 cm]{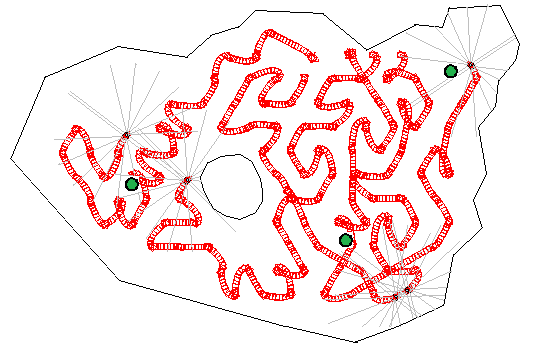}}
\caption{Robots' trajectories in the case of searching for three targets by five robots. } \label{MS:5robotPaths}
\end{figure}

To discover more about how the number of robots affects on the  search duration, we do more simulations for each case. For example, Fig. \ref{MS:2robotPaths} shows the paths of the robots in the case that two robots are looking for three targets similar to the previous instance. It shows that we have different paths thus different  search durations. While in the first simulation, the targets are detected in 4m3s, in the last one it occurs in 7m41s which is much more than the first one. Fig. \ref{MS:5robotPaths} shows the results of three different simulations using five robots. As depicted in that figure, the period of search for these simulations are 1m51s, 1m58s and 2m33s. Comparing to the case with two robots, it is obvious that the differences between periods of search in this case are less than the case of the team with two robots. That is an expected result because more robots means more coverage of the search area; hence, the chance of detecting  targets increases.

\begin{figure} [!hbt]
\renewcommand{\topfraction}{.85}
  \centering
  \includegraphics[width=13 cm, height=11 cm]{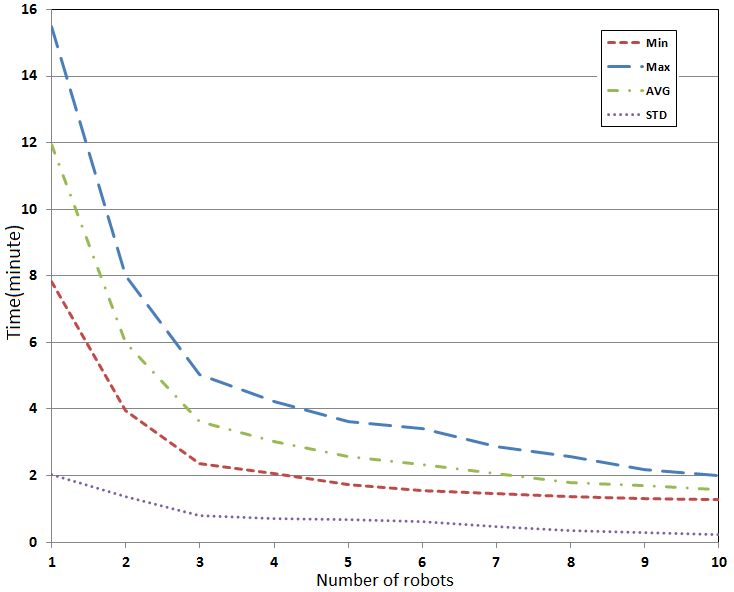}\\
  \caption{Time of target detection Vs. Number of robots}\label{MS:plotOfTimeOfDetect}
\end{figure}

To have a better view of the relation between the number of robots and the search duration, twenty  simulations with  different number of robots have been done. We consider teams consisting of one to ten robots. Then minimum, maximum and average of  search duration for simulations with each team as well as their standard deviation are calculated. Table \ref{MS:table2} displays the results of these simulations that are also depicted in Fig. \ref{MS:plotOfTimeOfDetect}. The results show that although the average of search duration to detect the targets decreases by increasing the number of robots, the standard deviation of the team with less number of robots is significantly high. Simply, the number of robots should be proportional to the search area to have an adequate chance to detect the targets in an acceptable time.
\begin{table}[!hbt]
\centering
\caption{Time of detecting targets }\label{MS:table2}
\resizebox{13cm}{!} {
\setlength{\extrarowheight}{6pt}
\begin{tabular}{|l|l|l|l|l|l|l|l|l|l|l|}
\hline
\textbf{No of Robots} & \textbf{1} & \textbf{2} & \textbf{3} & \textbf{4} & \textbf{5} & \textbf{6} & \textbf{7} & \textbf{8} & \textbf{9} & \textbf{10} \\ \hline
\textbf{Min} & 7.82 & 3.95 & 2.35 & 2.05 & 1.73 & 1.54 & 1.45 & 1.38 & 1.32 & 1.27 \\ \hline
\textbf{Max} & 15.49 & 8.02 & 5.05 & 4.21 & 3.63 & 3.42 & 2.88 & 2.56 & 2.19 & 2.01 \\ \hline
\textbf{Average} & 11.95 & 5.99 & 3.61 & 3.02 & 2.58 & 2.34 & 2.07 & 1.80 & 1.70 & 1.60 \\ \hline
\textbf{STD} & 2.03 & 1.37 & 0.79 & 0.71 & 0.69 & 0.61 & 0.47 & 0.36 & 0.30 & 0.24 \\ \hline
\end{tabular}
}
\end{table}

\section{Experiments with Real Robots}

\begin{figure}[!hbt]
  \centering
  \includegraphics[width=8 cm, height=7 cm]{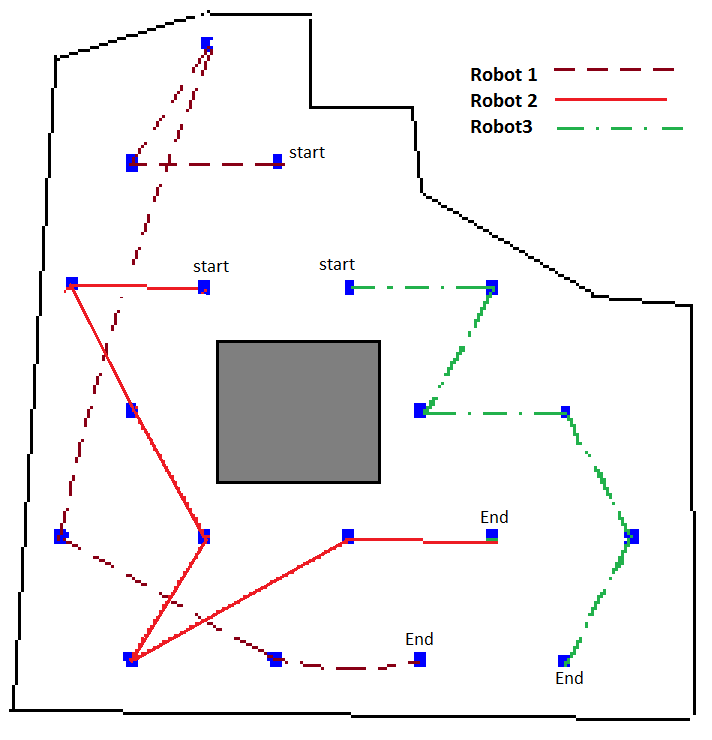}\\
  \caption{Robots' trajectories in the experiment }\label{MS:trajectExp}
\end{figure}

In this section, the experiment results with real robots are presented. We use Pioneer 3DX mobile robots to implement the proposed algorithm. Pioneer 3DX is  one of the  world's most popular research mobile robots. These robots are equipped  with encoders and sonar by which the localization and mapping can be done during the experiments. We do experiment with three robots in a surrounded area. The workspace is an  irregularly shaped fenced area of about $16.5 \ m^2$ with a rectangular obstacle of about $0.8 \ m^2$ (see Fig. \ref{MS:trajectExp}). The goal is searching the whole area by three robots using the proposed algorithm.

 We have used sonar to detect area boundaries, obstacles and other robots. A method of reactive potential field control is  applied to avoid  collisions with obstacles. Moreover, other robots are viewed as  obstacles by a robot. We have used ArNetworking, a library provided by Adept MobileRobots for communication between robots. It works with ARIA, the library we have applied to control the robots. The robots exchange the map of the environment that they make during the search operation and the vertices which they visit.

\begin{figure}[!hbt]
  \centering
  \includegraphics[width=5 cm, height=6 cm]{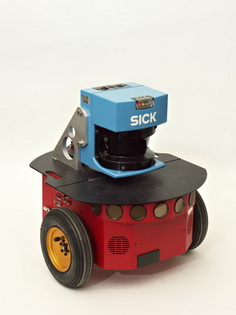}\\
  \caption{Pioneer 3DX robot used in the experiments }\label{MS:pioneer3dx}
\end{figure}

\begin{figure}[!hbtp]
\centering
\subfigure[]{\includegraphics[width=8 cm, height=4.5 cm]{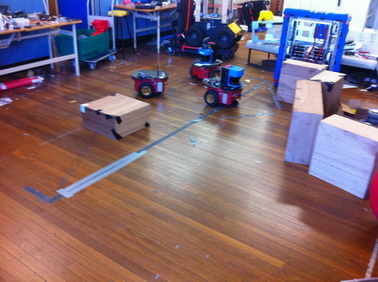}}\quad
\subfigure[]{\includegraphics[width=8 cm, height=4.5 cm]{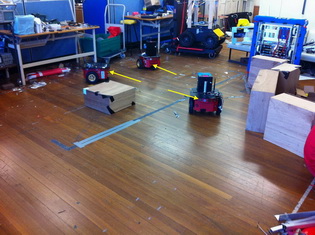}}\quad

\subfigure[]{\includegraphics[width=8 cm, height=4.5 cm]{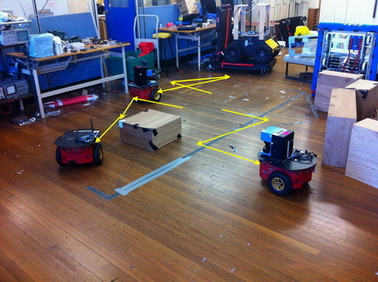}}\quad
\subfigure[]{\includegraphics[width=8 cm, height=4.5 cm]{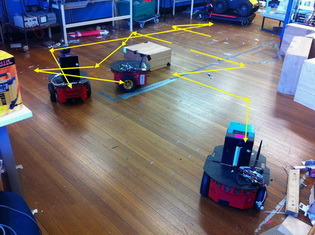}}\quad

\caption{Three robots exploring a whole area}
 \label{MS:expPics}
\end{figure}

 The snapshots of the experiment results are demonstrated in Fig. \ref{MS:expPics}. First, the robots are randomly placed in the field with an obstacle . After applying the algorithm, the robots  start to do the first stage of the algorithm, i.e.,  locating themselves on the vertices of  a common triangular grid using consensus variables locating algorithm. Fig. \ref{MS:expPics}(a) shows the result of this stage. Then, the  robots begin to search the area based on the second stage of the algorithm that is modified triangular grid-based search algorithm presented in this chapter. Fig. \ref{MS:expPics}(b)- Fig. \ref{MS:expPics}(d) are snapshots during this stage. The overall paths taken by the robots in the experiment are depicted in Fig. \ref{MS:trajectExp}. The results show that the area has been completely explored based on the proposed algorithm. Note that the robots do not necessarily reach the end points at the same time.

\begin{table}[!hbt]
\centering
\caption{search duration in the experiments (seconds) }\label{MS:timeTableExper}
\begin{tabular}{|l|l|l|l|l|l|l|l|l|l|l|}
\hline
\textbf{No of Robots} & \textbf{1} & \textbf{2} & \textbf{3}  \\ \hline
\textbf{Min} & 57 & 35 & 18   \\ \hline
\textbf{Max} & 66 & 41 & 22  \\ \hline
\textbf{Average} & 61.6 & 37.8 & 19.4  \\ \hline
\textbf{STD} & 4.04 & 2.39 & 1.67  \\ \hline
\end{tabular}
\end{table}

Also, to have a better view of a relation between the number of robots and the search duration in the experiments, five runs with one, two and three robots have been done. Table \ref{MS:timeTableExper} displays the calculated values for the minimum, maximum, average and  standard deviation of the search duration in the experiments.

Note that since the area and the triangles sides in the simulations and the experiments are different, and we had only three robots in the experiments, an exact comparison between results is not possible. But as the Tables \ref{MS:table1} and \ref{MS:timeTableExper} show, the results from the simulations and experiments give similar outcomes.  For example, in the simulations, three robots explore a 528 $m^2$  area in about 546 seconds (with exploring speed of 0.97 $m^2/s$), and in the experiments three robots explore a 16.5 $m^2$ area in about 19.4 seconds (with exploring speed of 0.85 $m^2/s$).  We have used 1 m for the sides of triangles in the experiments, while in the simulations, they have been  2 m; therefore, the robots have had more stops in the experiments thus the lower speed of  exploring.

\section{Comparison Between Presented Search Algorithms and Other Methods}
To evaluate the proposed triangular grid-based search algorithms  in  Chapters \ref{chap:RandomSearch}, \ref{chap:SemiRandomSearch} and this chapter  against other algorithms, we compare them with each other  also  with three more  algorithms: fixed-length random, Levy random-walk and Levy random-walk with potential field algorithms  \cite{sutantyo2010multi}. In \cite{sutantyo2010multi}, the simulation results for these three algorithms have been given in a plot of search time versus the number of robots. They consider a $400 \ m^{2}$ regular square  area to be explored by a team of mobile robots with a speed of 60 cm/s. We have considered an irregular shape with an area of about 528 $m^{2}$, and the robots with a maximum speed of  40 cm/s.

First, we compare our presented algorithm in this chapter with algorithms presented in Chapters \ref{chap:RandomSearch} and \ref{chap:SemiRandomSearch}. All algorithms are grid-based such that the robots certainly go through the vertices of a triangular grid. In the algorithm of random triangular grid-based search, presented in Chapter \ref{chap:RandomSearch}, robots go to the one of the six nearest vertices in each step regardless that the vertex has been explored already or not; therefore, we have a pure random grid-based search algorithm. On the other side, in the algorithm of semi-random triangular grid-based search presented in Chapter \ref{chap:SemiRandomSearch}, a robot goes to the one of the unexplored neighbouring vertices (each vertex has at most six neighbouring vertices). If all the nearest neighbouring vertices have been visited,  one of them is randomly selected. It is clear that a semi-random algorithm is faster than a pure random algorithm but requires more resources and processing. Our third proposed algorithm, i.e., modified triangular grid-based search algorithm presented in this chapter, drives the robots to move to the nearest unexplored vertex in the area in each step. In this algorithm, the robots are not constrained to only move to the one of the six nearest neighbouring vertices in the grid; but, they can move to the nearest  unexplored vertex anywhere in the area.

\begin{figure}[!hbt]
  \centering
  \includegraphics[width=13 cm, height=10 cm]{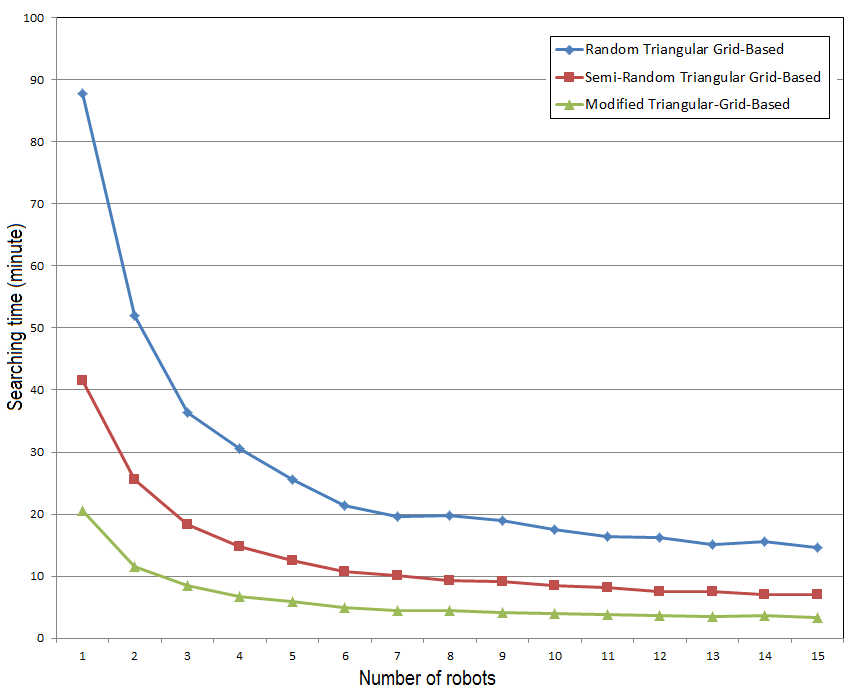}\\
  \caption{Comparison between presented search algorithms }\label{MS:comparison3Algorithms}
\end{figure}

Fig. \ref{MS:comparison3Algorithms} displays a comparison between our presented search algorithms using data from Tables \ref{RS:table1}, \ref{SR:table1} and \ref{MS:table1}. As shown in Fig. \ref{MS:comparison3Algorithms}, semi-random triangular grid-based search algorithm is  faster than  random triangular grid-based search algorithm, and modified triangular grid-based search algorithm is much faster than both. Numerically, the speed of search by  semi-random triangular grid-based search algorithm is about fifty percent more than by random triangular grid-based search algorithm, and the speed of search by  modified triangular grid-based search algorithm is more than fifty percent greater than by random triangular grid-based search algorithm.
Also, a comparison between the standard deviation of these algorithms shows that standard deviation in modified triangular grid-based search algorithm is much less than the other methods meaning that in this algorithm, repeating the simulation for a team of robots does not have a considerable effect on the time of search.

\begin{figure}[!hbt]
  \centering
  \includegraphics[width=13 cm, height=10 cm]{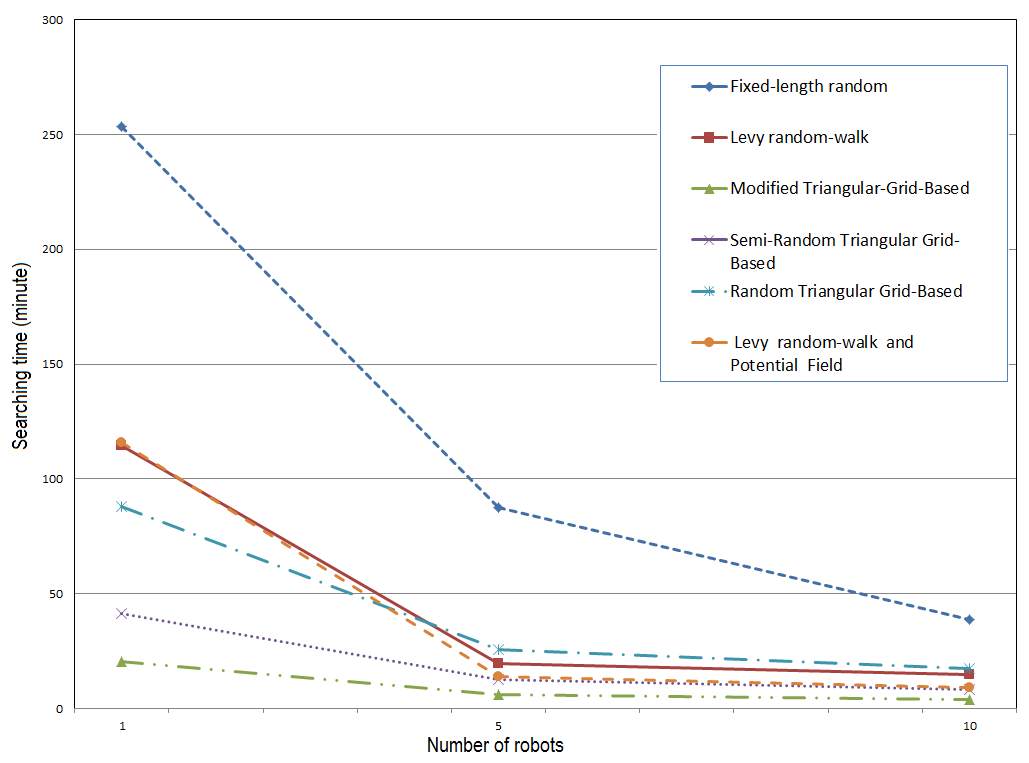}\\
  \caption{Comparison between presented search algorithms and other methods}\label{MS:comparisonPlotTime}
\end{figure}

\begin{figure}[!hbt]
  \centering
  \includegraphics[width=13 cm, height=10 cm]{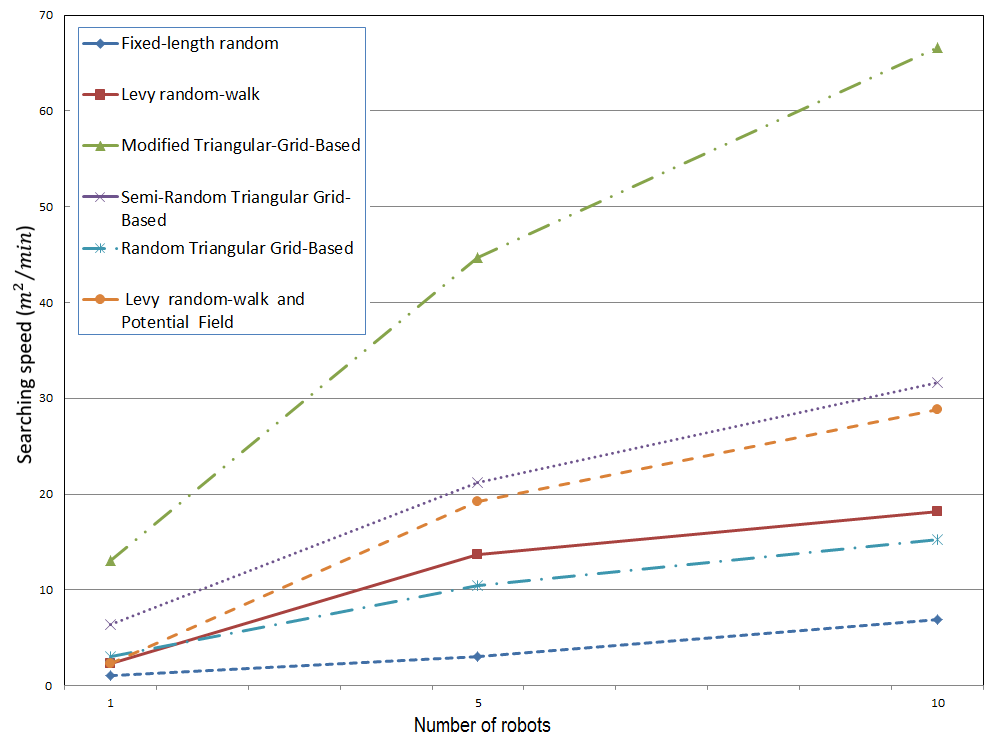}\\
  \caption{Comparison between presented search algorithms and other methods}\label{MS:comparisonPlotSpeed}
\end{figure}

Second, we compare our three algorithms with three other algorithms: fixed-length random, Levy random-walk and Levy random-walk with potential field algorithms.  Table \ref{MS:comparisonTimeTable} shows the search time for these algorithms and our proposed algorithms that also depicted in Fig. \ref{MS:comparisonPlotTime}. Note that the search times given in Table \ref{MS:comparisonTimeTable} are approximated values extracted from the graph given in \cite{sutantyo2010multi}, and  also average values from Tables  \ref{RS:table1}, \ref{SR:table1} and \ref{MS:table1}. Table \ref{MS:comparisonSpeedTable} shows the speed of search versus the number of robots for the algorithms that also depicted in Fig. \ref{MS:comparisonPlotSpeed}. Since  the size of  the areas  and speed of the robots in our simulations and the simulations in  \cite{sutantyo2010multi} are different, they have been normalized; then, we can compare them.    As shown in Fig. \ref{MS:comparisonPlotSpeed}, the fastest  algorithm is our algorithm presented in this chapter, i.e., modified triangular grid-based search algorithm; more than two times faster than the fastest algorithm presented in \cite{sutantyo2010multi}. Fig. \ref{MS:comparisonPlotSpeed} also shows that semi-random triangular grid-based search algorithm presented in Chapter \ref{chap:SemiRandomSearch}, is almost as fast as Levy random-walk and potential field algorithm, the fastest algorithm in \cite{sutantyo2010multi};  but, still slightly faster. Finally, Fig. \ref{MS:comparisonPlotSpeed} shows that  random triangular grid-based search algorithm presented in Chapter \ref{chap:RandomSearch}, is slightly slower than Levy random-walk algorithm but noticeably faster than the fixed-length random algorithm; about two times faster.

\begin{table}[!hbt]
\centering
\caption{Comparison Between Presented Search Algorithms and Other Methods  (Search time, minute)}
\label{MS:comparisonTimeTable}
\resizebox{\textwidth}{!}{%
\setlength{\extrarowheight}{14pt}
\begin{tabular}{|l|p{2cm}|p{1.8cm}|p{2cm}|p{1.9cm}|p{1.8cm}|p{2.2cm}|}
\hline
\diaghead(1, -1){Number of robots Search Algorithm}{\large Number of\\ \large robots}{\large Search \\ \large Algorithm} & \textbf{Fixed-length random} & \textbf{Levy random-walk} & \textbf{Levy  random-walk  and  Potential Field} & \textbf{Random Triangular Grid-Based} & \textbf{Semi-Random Triangular Grid-Based} & \textbf{Modified Triangular-Grid-Based} \\ \hline
1  & 253.3 & 114.7 & 115.7 & 87.9 & 41.6 & 20.5 \\ \hline
5  & 87.5  & 19.5  & 13.9  & 25.6 & 12.6 & 6.0  \\ \hline
10 & 38.8  & 14.7  & 9.3   & 17.5 & 8.4  & 4.0  \\ \hline
\end{tabular}
}
\end{table}

\begin{table}[!hbt]
\centering
\caption{Comparison Between Presented Search Algorithms and Other Methods  (Searching speed, $m^{2}/min$)}
\label{MS:comparisonSpeedTable}
\resizebox{\textwidth}{!}{%
\setlength{\extrarowheight}{16pt}
\begin{tabular}{|l|p{2cm}|p{1.8cm}|p{2cm}|p{1.9cm}|p{1.8cm}|p{2.2cm}|}
\hline
\diaghead(1, -1){Number of robots Search Algorithm}{\large Number of\\ \large robots}{\large Search \\ \large Algorithm}& \textbf{Fixed-length random} & \textbf{Levy random-walk} & \textbf{Levy  random-walk  and  Potential Field} & \textbf{Random Triangular Grid-Based} & \textbf{Semi-Random Triangular Grid-Based} & \textbf{Modified Triangular-Grid-Based} \\ \hline
1  & 1.1 & 2.3  & 2.3  & 3.0  & 6.4  & 13.0                    \\ \hline
5  & 3.0 & 13.7 & 19.2 & 10.4 & 21.2 & 44.7           \\ \hline
10 & 6.9 & 18.2 & 28.8 & 15.2 & 31.6 & 66.7 \\ \hline
\end{tabular}
}
\end{table}

\section{Summary}
In this chapter, we have developed a distributed control algorithm to drive a multi-robot team to explore an unknown area. We have used a triangular grid pattern and a two-stage algorithm for the control law so that the robots move through the vertices of the grid during the search procedure. Therefore, a complete search of the whole area has been guaranteed. A mathematically rigorous proof of convergence of the presented algorithm has been demonstrated. Furthermore, the computer simulation results using MobileSim, a powerful simulator of real robots and environments, have been presented to confirm that the algorithm is effective and practicable. Also, the experiments with Pioneer 3DX wheeled mobile robots have been done to confirm the performance of our suggested algorithm. The presented results of  the experiments with real robots  show  that the algorithm is quite practical. Finally, a comparison between  the proposed triangular grid-based search algorithms  in  Chapters \ref{chap:RandomSearch},  \ref{chap:SemiRandomSearch}  and this chapter against three other algorithms have been given.

\chapter{Formatiom Building with Obstacle Avoidance}\label{chap:Formation}

In this chapter, we propose a distributed motion coordination control algorithm for a team of mobile robots so that the robots collectively move in a desired geometric pattern from any initial position while  avoiding the obstacles on their routes. In the proposed method, the robots have no information on the shape and position of the obstacles and only use range sensors to obtain the information. We use  standard kinematic equations for the robots with hard constraints on the linear and angular velocities.  There is no leader in the team and the robots apply a  distributed control algorithm  based on the local information they obtain from their nearest neighbours. We take the advantage of using the  consensus variables approach that is a known rule in multi-agent systems. Also, we propose a randomized algorithm for the anonymous robots, which achieves the convergence to the desired configuration with probability 1. Furthermore,  we propose a novel obstacle avoidance technique based on the information from the range sensors. Mathematically rigorous proofs of the proposed control algorithms are given, and  the effectiveness of the algorithms are illustrated via computer simulations.

\section{Multi-Robot System}

We consider a system consisting of n autonomous mobile robots labeled 1 through $n$ moving in a plane. The kinematic equations of  motion for the robots are given by

\begin{equation}\label{FB:kinematicEq}
\begin{split}
&\dot{x}_{i}(t)=\ v_{i}(t)\cos(\theta_{i}(t))\\
&\dot{y}_{i}(t)=\ v_{i}(t)sin(\theta_{i}(t))\\
&\dot{\theta}_{i}(t)=\ \omega_{i}(t)
\end{split}
\end{equation}

for all $i=1,2,\ldots,n$, where $(x_{i}(t),\ y_{i}(t))$ are the Cartesian coordinates of robot $i$ at time $t$, and $\theta_{i}(t)$ is its orientation with respect to the $x$-axis  measured in the counter-clockwise direction. Also, $v_{i}(t)$, the speed of the robot, and $\omega_{i}(t)$, its angular velocity, are the control inputs. Note that model (\ref{FB:kinematicEq}) is a very common model, and many mobile agents (UGVs, UAVs, missiles, etc.) can be described by this model \cite{manchester2006circular, fierro1998control, desai2001modeling, yang1999sliding, barfoot2004motion, tzafestas2013introduction}. Furthermore, we need the following practical constraints:

\begin{equation}\label{FB:constrain1}
-\omega^{\max}\leq\omega_{i}(t)\leq\omega^{\max}\qquad\forall t\geq 0
\end{equation}
\begin{equation}\label{FB:constrain2}
V^{m}\leq v_{i}(t)\leq V^{M}\qquad\forall t\geq 0
\end{equation}
for all $i=1,2,\ldots,n$. Here, $\omega^{\max}>0$ and $0<V^{m}<V^{M}$ are given constants.

Moreover, let  $z_{i}(t)$ be the vector of the robots' coordinates and $V_{i}(t)$ as the robots' velocity vector defined by
\begin{equation}\label{FB:vectors}
z_{i}(t):=\left(\begin{array}{l}
x_{i}(t)\\
y_{i}(t)
\end{array}\right),\ V_{i}(t):= \left(\begin{array}{l}
v_{i}(t)\mathrm{c}\mathrm{o}\mathrm{s}(\theta_{i}(t))\\
v_{i}(t)\mathrm{s}\mathrm{i}\mathrm{n}(\theta_{i}(t))
\end{array}\right)
\end{equation}
for all $i=1,2,\ldots,n$.

We assume that the robots share their information via a wireless communication at discrete time instants $k=0,1,2,\ldots$. Due to limited communication range of the robots, we assume  $r_{c}$ as the communication range  for all   the mobile robots, meaning that a robot can only receive information from the robots which are located not farther than $r_{c}$.
\begin{mydef}\label{FB:neighbour}
Robot $j$ is the neighbour of robot $i$ at time $k$ if and only if it is located on the disk  of radius $r_{c}$ with the center of robot $i$'s position. Also, let $\mathcal{N}$$_{i}(k)$ be  the set of all neighbours of the robot $i$ at time $k$, and $|\mathcal{N}$$_{i}(k)|$ be the number of elements in $\mathcal{N}$$_{i}(k)$.
\end{mydef}

The relationship among the  robots can be defined by an undirected graph $\mathcal{G}(k)$. We assume that any robot of the multi-robot team is a node of the graph $\mathcal{G}(k)$ at time $k$, i.e., $i$ in $V_{\mathcal{G}}=\{1,2,\ldots,n\}$, the node set of $\mathcal{G}(k)$, is related to robot $i$.  In addition, robot $i$ is the neighbour of  robot $j$  at time $k$ if and only if  there is an edge between the nodes $i$ and $j$ of the graph $\mathcal{G}(k)$ where $i\neq j$. Therefore, the problem of communication among the robots equals the problem of the connectivity of the related graph. Note that robot $i$ does not need  to be the neighbour of robot $j$  to get the information from. The information is transferred through the other robots which connect these robots in the related graph. We will also need the following assumption.

\begin{myassump}\label{FB:timeIntervals}
There exists an infinite sequence of contiguous, non-empty, bounded time-intervals $[k_{j},  k_{j+1})$, $j= 0,1,2,\ldots$, starting at $k_{0}=0$, such that across each $[k_{j},\ k_{j+1})$, the union of the collection $\{\mathcal{G}(k):k\in[k_{j},\ k_{j+1})\}$ is a connected graph.
\end{myassump}

To achieve the common heading and speed of formation, we use the consensus variables $\tilde{\theta}_{i}(k)$ and $\tilde{v}_{i}(k)$, respectively. Also, we need a common origin of coordinates of the formation for the multi-robot system; therefore, $\tilde{x}_{i}(k)$ and $\tilde{y}_{i}(k)$ are   used as the consensus variables for the coordinates of the robots. In other words, the robots start with different initial values of consensus variables $\tilde{x}_{i}(0), \tilde{y}_{i}(0), \tilde{\theta}_{i}(0)$ and $\tilde{v}_{i}(0)$, and each robot calculates these consensus variables at any time $k$ such that eventually the consensus variables converge to some consensus values which define a common speed and orientation in a common coordinate system.

\begin{myassump}\label{FB:initialValues}
The initial values of the consensus variables $\tilde{\theta}_{i}$ satisfy $\tilde{\theta}_{i}(\mathrm{0})\in[0,\ \pi)$ for all $i=1,2,\ldots,n$.
\end{myassump}

\begin{myassump}\label{FB:information}
The information on the other robots that is available to robot $i$ at time $k$ is the coordinates $(x_{j}(k),\ y_{j}(k))$, and the consensus variables $\tilde{\theta}_{j}(k),\tilde{x}_{j}(k),\tilde{y}_{j}(k)$ and $\tilde{v}_{j}(k)$ for all $j\in \mathcal{N}_{i}(k)$.
\end{myassump}

In practice, the coordinates of neighbouring robots can be obtained using Kalman state estimation via limited capacity communication channels \cite{malyavej2005problem}.

\section{Formation Building}\label{FB:SecFormation}
We propose the following rules for updating the consensus variables $\tilde{\theta}_{i}(k),\tilde{x}_{i}(k),\tilde{y}_{i}(k)$ and $\tilde{v}_{i}(k)$ :

\begin{equation}\label{FB:consensusVar}
\begin{aligned}
\tilde{\theta}_{i}(k+1)=&\frac{\tilde{\theta}_{i}(k)+\sum\limits_{j\in \mathcal{N}_{i}(k)}\tilde{\theta}_{j}(k)}{1+|\mathcal{N}_{i}(k)|}\\
  \tilde{x}_{i}(k+1)=&\frac{x_{i}(k)+\tilde{x}_{i}(k)+\sum\limits_{j\in \mathcal{N}_{i}(k)}(x_{j}(k)+\tilde{x}_{j}(k))}{1+|\mathcal{N}_{i}(k)|}\displaystyle-x_{i}(k+1)\\
 \tilde{y}_{i}(k+1)=&\frac{y_{i}(k)+\tilde{y}_{i}(k)+\sum\limits_{j\in \mathcal{N}_{i}(k)}(y_{j}(k)+\tilde{y}_{j}(k))}{1+|\mathcal{N}_{i}(k)|}\displaystyle-y_{i}(k+1)\\
 \tilde{v}_{i}(k+1)=&\frac{\tilde{v}_{i}(k)+\sum\limits_{j\in \mathcal{N}_{i}(k)}\tilde{v}_{j}(k)}{1+|\mathcal{N}_{i}(k)|}
\end{aligned}
\end{equation}

Based on  rule (\ref{FB:consensusVar}), the mobile robots use the consensus variables to achieve a consensus on the heading, speed and origin of the coordinate system of the formation.

\begin{mylemma}\label{FB:lem1}
 Suppose that Assumptions \ref{FB:timeIntervals} and \ref{FB:initialValues} hold, and the consensus variables are updated according to the decentralized control rule (\ref{FB:consensusVar}). Then, there exist constants $\tilde{\theta}_{0},\tilde{X}_{0},\tilde{Y}_{0}$ and $\tilde{v}_{0}$ such that

\begin{flalign}\label{FB:constants}
\begin{aligned}
    &\lim_{k\rightarrow\infty}\tilde{\theta}_{i}(k)=\tilde{\theta}_{0}\\
    &\lim_{k\rightarrow\infty}\tilde{v}_{i}(k)=\tilde{v}_{0}\\
    &\lim_{k\rightarrow\infty}(x_{i}(k)+\tilde{x}_{i}(k))=\tilde{X}_{0}\\
    &\lim_{k\rightarrow\infty}(y_{i}(k)+\tilde{y}_{i}(k))=\tilde{Y}_{0}
    \end{aligned}
\end{flalign}

for all $i=1,2,\ldots,n$. Furthermore, the convergence in (\ref{FB:constants}) is exponentially fast.
\end{mylemma}
The statement of Lemma \ref{FB:lem1} immediately follows from the main result of \cite{jadbabaie2003coordination}. Note that  the constants $\tilde{\theta}_{0}, \tilde{X}_{0}, \tilde{Y}_{0}$ and $\tilde{v}_{0}$ are the same for all the robots.

\begin{mydef}\label{FB:globalStabilizing}
A navigation law is said to be globally stabilizing with initial conditions $(x_{i}(0),\ y_{i}(0),\ \theta_{i}(0))$, $i=1,2,\ldots,n$ and the given values of configuration $\mathcal{C} = \{X_{1},\ X_{2},\ \ldots,\ X_{n},\ Y_{1},\ Y_{2},\ \ldots,\ Y_{n}\}$, if there exists a Cartesian coordinate system and $\tilde{v}_{0}$ such that the solution of the closed-loop system (\ref{FB:kinematicEq}) with these initial conditions and the proposed navigation law in this Cartesian coordinate system satisfies:

\begin{align}\label{FB:limit1}
\begin{split}
\lim_{t\rightarrow\infty}(x_{i}(t)-x_{j}(t))=X_{i}-X_{j}\\
\lim_{t\rightarrow\infty}(y_{i}(t)-y_{j}(t))=Y_{i}-Y_{j}
\end{split}
\end{align}
and
\begin{align}\label{FB:limit2}
\begin{split}
\lim_{t\rightarrow\infty}\theta_{i}(t)=0\\
\lim_{t\rightarrow\infty}v_{i}(t)=\tilde{v}_{0}
\end{split}
\end{align}
for all $1\leq i\neq j\leq n$.
where $X_{1},X_{2},\ldots,X_{n},Y_{1},Y_{2},\ldots,$ $Y_{n}$ are given constants.
\end{mydef}

 Rule (\ref{FB:limit2}) means that as $t\rightarrow\infty$ all the robots will finally move in the same direction along the $x$-axis with the same speed. Furthermore, rule (\ref{FB:limit1}) indicates that a geometric configuration of the robots  given by $\mathcal{C}$ will be obtained. For instance, if we have four robots and $\mathcal{C}=\{0,0,2,2,0,1,0,1\}$, then  the geometric formation of the robots will be a rectangle of sides 1 and 2.

Since we use the discrete time consensus variables $\tilde{\theta}_{i}(k),\tilde{x}_{i}(k),\tilde{y}_{i}(k)$ and $\tilde{v}_{i}(k)$ updated according to (\ref{FB:consensusVar}), we need to define the corresponding piecewise constant continuous time variables as

\begin{align}\label{FB:continuousTimeVar}
\begin{split}
\tilde{\theta}_{i}(t)\ :=\tilde{\theta}_{i}(k)\ \forall t\in(k,\ k+1)\\
\tilde{x}_{i}(t)\ :=\tilde{x}_{i}(k)\ \forall t\in(k,\ k+1)\\
\tilde{y}_{i}(t)\ :=\tilde{y}_{i}(k)\ \forall t\in(k,\ k+1)\\
\tilde{v}_{i}(t)\ :=\tilde{v}_{i}(k)\ \forall t\in(k,\ k+1).
\end{split}
\end{align}

For any time $t$ and any robot $i$, we consider a Cartesian coordinate system with the $x$-axis in the direction $\tilde{\theta}_{i}(t)$ (according to the definition (\ref{FB:continuousTimeVar}), $\tilde{\theta}_{i}(t)$ is piecewise constant). In other words, in this coordinate system $\tilde{\theta}_{i}(t)=0$ and $\ x_{i}(t)$, $y_{i}(t)$ are now coordinates of robot $i$ in this system. Notice that we now formulate our decentralized control law for each robot in its own coordinate system. Since according to Lemma \ref{FB:lem1}, $\tilde{\theta}_{i}(k)$ converges to the same value for all $i$, all these robots' coordinate systems converge to the same coordinate system in which (\ref{FB:limit1}) holds.

\begin{myassump}\label{FB:c and C}
Let $c>0$ be any constant such that
\begin{equation}\label{FB:cConstant}
c>\frac{2V^{M}}{\omega^{\max}}.
\end{equation}
We assume that the constant $c$ and also the configuration $\mathcal{C}$ are known to all the robots.
\end{myassump}

Introduce the functions $h_{i}(t)$ as

\begin{equation}\label{FB:hFunction}
h(t)\ :=(x_{i}(t)+\tilde{x}_{i}(t))+X_{i}+t\tilde{v}_{i}(t)
\end{equation}

for all $i =$ 1,2,$\ldots,n$. Also, introduce two-dimensional vector $g_{i}(t)$ as

\begin{equation}\label{FB:gVector}
g_{i}(t)\ :\ =\binom{g_{i}^{x}(t)}{g_{i}^{y}(t)}
\end{equation}

where

\begin{align}\label{FB:gxy}
\begin{split}
&g_{i}^{x}(t) : =\left\{\begin{array}{l}
h_{i}(t)+c\ \mathrm{i}\mathrm{f}\ x_{i}(t)\leq h_{i}(t)\\
x_{i}(t)+c\ \mathrm{i}\mathrm{f}\ x_{i}(t)>h_{i}(t)
\end{array}\right.\\
&g_{i}^{y}(t):=(y_{i}(t)+\tilde{y}_{i}(t))+Y_{i}
\end{split}
\end{align}

and two-dimensional vector $d_{i}(t)$ as

\begin{equation}\label{FB:dVector}
d_{i}(t):=g_{i}(t)-z_{i}(t)
\end{equation}

for all $i =$ 1,2,$\ldots,n$, where $z_{i}(t)$ is defined by (\ref{FB:vectors}).

\begin{figure} [bt]
  \centering
  \includegraphics[width=10 cm, height=9 cm]{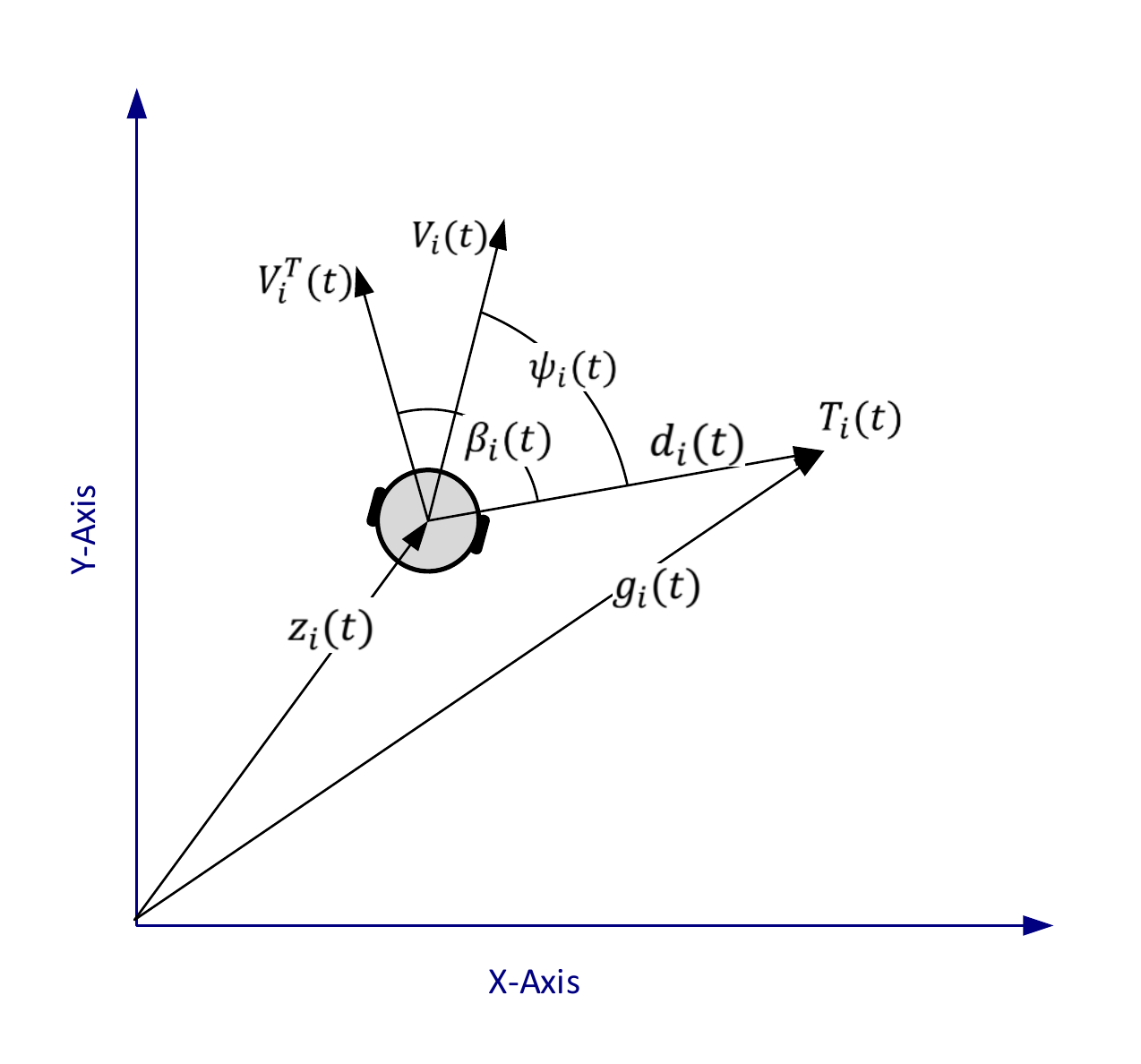}\\
  \caption{Vectors geometry}\label{FB:controlFig1}
\end{figure}

Now, we introduce the following decentralized control law:

\begin{align}\label{FB:controlLaw}
\begin{split}
&v_{i}(t)=\left\{\begin{array}{ll}
V^{M} & \mathrm{i}\mathrm{f}\ x_{i}(t)\leq h_{i}(t)\\
V^{m} & \mathrm{i}\mathrm{f}\ x_{i}(t)>h_{i}(t)
\end{array}\right.\\
&\omega_{i}(t)=\omega^{max}sign(\psi_{i}(t))
\end{split}
\end{align}

for all $i=1,2,\ldots,n$, where  $\psi_{i}(t)$ is the angle between $V_{i}(t)$ and  $d_{i}(t)$ measured from $V_{i}(t)$ in the counter-clockwise direction, i.e.,
\begin{equation}\label{FB:sign}
\psi_{i}(t) = \angle (V_{i}(t) , d_{i}(t))
\end{equation}
(see Fig. \ref{FB:controlFig1}), and $sign(\cdot)$ is defined by

\begin{equation}\label{FB:sign}
sign(\alpha) :=\left\{\begin{array}{ll}
-1 & \mathrm{i}\mathrm{f}\ \alpha<0\\
0 & \mathrm{i}\mathrm{f}\ \alpha=0\\
1 & \mathrm{i}\mathrm{f}\ \alpha>0
\end{array}\right.
\end{equation}

We also need the following assumption.

\begin{myassump}\label{FB:initialRobotsSpeeds}
The initial robots' speeds satisfy
\begin{equation*}\label{FB:}
V^{m}<v_{i}(0)<V^{M}
\end{equation*}
\end{myassump}
for all $i=1,2,\ldots,n$.

Notice that Assumption \ref{FB:initialRobotsSpeeds} is just slightly stronger than  the requirement (\ref{FB:constrain2}) for $t=0$ where non-strict inequalities are required.

The proposed algorithm is based on robots' headings and coordinates which, of course, depend on initial conditions. Therefore, the proposed law depends on initial conditions on robots' headings and coordinates. The connectivity of the multi-robot formation is maintained due to Assumption \ref{FB:timeIntervals} which is a standard assumption in numerous papers on multi- agent systems; see, e.g., \cite{jadbabaie2003coordination, savkin2010decentralized} and the references therein.

Now, we are in a position to present the main result of this section.

\begin{mytheorem}\label{FB:theoremForm}
Consider the autonomous mobile robots described by the equations (\ref{FB:kinematicEq}) and the constraints (\ref{FB:constrain1}), (\ref{FB:constrain2}). Let $\mathcal{C}=\{X_{1},\ X_{2},\ \ldots,\ X_{n},\ Y_{1},\ Y_{2},\ \ldots,\ Y_{n}\}$ be a given configuration. Suppose that Assumptions \ref{FB:timeIntervals}, \ref{FB:initialValues}, \ref{FB:c and C} and \ref{FB:initialRobotsSpeeds} hold. Then, the decentralized control law (\ref{FB:consensusVar}), (\ref{FB:controlLaw}) is globally stabilizing with any initial conditions and the configuration $\mathcal{C}$.
\end{mytheorem}

{\bf Proof of Theorem }\ref{FB:theoremForm}: Let $1\leq i\leq n$. We consider a fictitious target $T_{i}$ moving on the plane with coordinates $g_{i}(t)$ defined by (\ref{FB:gxy}). Furthermore, introduce another fictitious target $\tilde{T}_{i}$ moving on the plane with coordinates $\tilde{g}_{i}(t)$ defined by

\begin{equation}\label{FB:gTildeVector}
\tilde{g}_{i}(t)\ :\ =\binom{\tilde{g}_{i}^{x}(t)}{\tilde{g}_{i}^{y}(t)}
\end{equation}

where

\begin{align}\label{FB:gxyTilda}
\begin{split}
&\tilde{g}_{i}^{x}(t) : =\left\{\begin{array}{l}
X_{0}+X_{i}+t\tilde{v}_{0}+c\ \  \mathrm{i}\mathrm{f}\ x_{i}(t)\leq\tilde{X}_{0}+X_{i}+t\tilde{v}_{0}\\
x_{i}(t)+c\ \ \  \mathrm{i}\mathrm{f}\ \ \ x_{i}(t)>\tilde{X}_{0}+X_{i}+t\tilde{v}_{0}
\end{array}\right.\\
&\tilde{g}_{i}^{y}(t):=\tilde{Y}_{0}+Y_{i}
\end{split}
\end{align}

It immediately follows from Lemma \ref{FB:lem1} that
\begin{equation}\label{FB:lim gi g}
\lim_{t\rightarrow\infty}(\tilde{g}_{i}(t)-g_{i}(t))=0.
\end{equation}
Moreover, this convergence is exponentially fast. Let $\psi_{i}(t)$ be the angle between the velocity vector $V_{i}(t)$ of robot $i$ and the line-of-sight between the robot and $T_{i}$; and $\beta_{i}(t)$ be the angle between the velocity vector $V_{i}^{T}(t)$ of $T_{i}$ and the line-of-sight from robot $i$ to $T_{i}$ (see Fig.\ref{FB:controlFig1}).

\begin{figure} [bt]
  \centering
  \includegraphics[trim=2cm 1.2cm 2cm 2cm ,width=8 cm, height=7 cm]{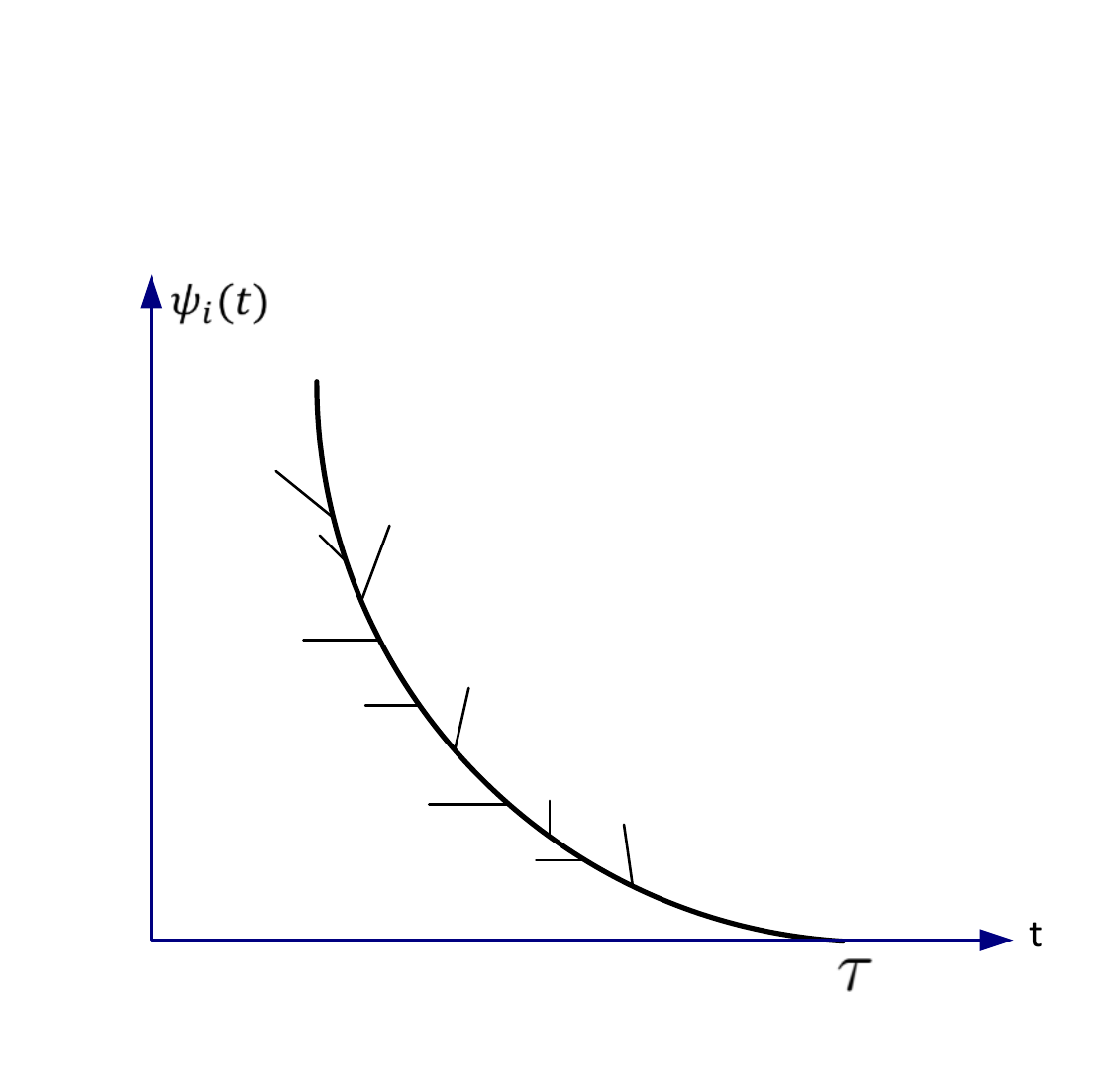}\\
  \caption{Sliding mode solution}\label{FB:sliding}
\end{figure}

It is well-known (see, e.g., \cite{teimoori2010equiangular}) that the following equation holds:\

\begin{equation}\label{FB:psiDotEq}
\dot{\psi}_{i}(t)=\frac{\Vert V_{i}(t)\Vert\sin\psi_{i}(t)}{\Vert\tilde{d}_{i}(t)\Vert}-\omega_{i}(t)-\frac{\Vert V_{i}^{T}(t)\Vert\sin\beta_{i}(t)}{\Vert\tilde{d}_{i}(t)\Vert}
\end{equation}

where $\tilde{d}_{i}(t)$ is defined as
\begin{equation}\label{FB:dTilda}
\tilde{d}_{i}(t)\ :=\tilde{g}_{i}(t)-z_{i}(t),
\end{equation}

 $z_{i}(t)$ is defined by (\ref{FB:vectors}), and $\Vert\cdot\Vert$ denotes the standard Euclidean vector norm. It obviously follows from (\ref{FB:gxyTilda}),(\ref{FB:dTilda})  that

\begin{equation}\label{FB:diTilda c}
\Vert\tilde{d}_{i}(t)\Vert\geq c\ \ \ \forall t\geq 0.
\end{equation}
Furthermore, (\ref{FB:gxyTilda}) implies that

\begin{equation*}\label{FB:VT Vector}
V_{i}^{T}(t):=\binom{(V_{i}^{Tx}(t)}{V_{i}^{Ty}(t)};
\end{equation*}

\begin{align}\label{FB:VTxy}
\begin{split}
&V_{i}^{Tx}(t) =\left\{\begin{array}{l}
\tilde{v}_{0}\  \mathrm{i}\mathrm{f}\ \ x_{i}(t)\leq X_{0}+X_{i}+t\tilde{v}_{0}\\
v_{i}(t)\   \mathrm{i}\mathrm{f}\ \ x_{i}(t)>X_{0}+X_{i}+t\tilde{v}_{0}
\end{array}\right.\\
&V_{i}^{Ty}(t) = 0.
\end{split}
\end{align}

It follows from (\ref{FB:VTxy}) and (\ref{FB:constrain2}) that
\begin{equation}\label{FB:ViTilda}
\Vert V_{i}^{T}(t)\Vert\leq V^{M}.
\end{equation}
Now, we consider the control law (\ref{FB:controlLaw}) with $d_{i}$ replaced by $\tilde{d}_{i}$. The inequality (\ref{FB:ViTilda}) together with (\ref{FB:cConstant}), (\ref{FB:psiDotEq}) and (\ref{FB:diTilda c}) implies that under this control law, there exists a constant $\epsilon>0$ such that

\begin{equation}\label{FB:psiEpsilon}
\begin{split}
\dot{\psi}_{i}(t)<-\epsilon\ \ if\ \ \psi_{i}(t)>0\\
\dot{\psi}_{i}(t)>\epsilon\ \  if\ \ \psi_{i}(t)<0.
\end{split}
\end{equation}
Therefore, there exists a time $\tau>0$ such that
\begin{equation}\label{FB:psiZero}
\psi_{i}(t)=0\ \ \ \forall t\geq \tau.
\end{equation}
Notice that the closed-loop system (\ref{FB:kinematicEq}), (\ref{FB:controlLaw}) is a system of differential equations with discontinuous right-hand sides; the equation $\psi_{i}=0$ defines a switching surface of this system, and a solution satisfying (\ref{FB:psiZero}) is a sliding mode; see, e.g.,  \cite{utkin2013sliding}. Also,   (\ref{FB:consensusVar}), (\ref{FB:controlLaw}) that belongs to the class of switched control laws and the system (\ref{FB:kinematicEq}) with such controller is a hybrid dynamical system \cite{savkin1996hybrid, matveev2000qualitative, savkin2002hybrid}. The inequalities (\ref{FB:psiEpsilon}) guarantee that this sliding mode solution of the closed-loop system looks as it is shown in Fig. \ref{FB:sliding} and satisfies
\begin{equation}\label{FB:psiDotZero}
\dot{\psi}_{i}(t)=0\ \ \  \ \forall i\ \ \forall t\geq \tau.
\end{equation}
From this and (\ref{FB:psiDotEq}), we obtain that
\begin{equation}\label{FB:omegaEq}
\omega_{i}(t)=-\frac{\Vert V_{i}^{T}(t)\Vert\sin\beta_{i}(t)}{||\tilde{d}_{i}(t)\Vert}
\end{equation}
for all sliding mode solutions. Therefore, for any initial condition, the sliding mode solution is unique and well-defined. Furthermore, (\ref{FB:omegaEq}), (\ref{FB:ViTilda}), (\ref{FB:cConstant}) and (\ref{FB:diTilda c}) imply that the constraint (\ref{FB:constrain1}) holds for any sliding mode solution satisfying (\ref{FB:psiZero}).

Furthermore, the condition (\ref{FB:psiZero}) means that the velocity vector $V_{i}(t)$ is parallel to the vector $\tilde{d}_{i}(t)$ for all $t\geq \tau$. Hence, for all $t\geq \tau$, we have that the robot's velocity vector is always pointed at $\tilde{g}_{i}(t)$ . Since $\tilde{g}_{i}^{y}(t)=\tilde{Y}_{0}+Y_{i}$, we obtain that $y_{i}(t)\rightarrow\tilde{Y}_{0}+Y_{i}$. The second of the conditions (\ref{FB:limit1}) immediately follows from this. Furthermore, Assumption \ref{FB:initialRobotsSpeeds} implies that $V^{m}\leq\tilde{v}_{0}\leq V^{M}$. The fact that the velocity vector $V_{i}(t)$ is parallel to the vector $\tilde{d}_{i}(t)$ for all $t\geq \tau$, and the control law (\ref{FB:controlLaw}) with $d_{i}$ replaced by $\tilde{d}_{i}$ imply that
$$
\tilde{d}_{i}(t)=\left(\begin{array}{l}
c\\
\mathrm{0}
\end{array}\right)
$$
for all $i$ and all large enough $t$. The first of the conditions (\ref{FB:limit1}) immediately follows from this. We proved the statement of the theorem for the control law (\ref{FB:controlLaw}) with $d_{i}$ replaced by $\tilde{d}_{i}$. This and the exponential convergence (\ref{FB:lim gi g}) together with the inequality (\ref{FB:diTilda c}) imply that the same statement holds for the original control law (\ref{FB:controlLaw}). This completes the proof of Theorem \ref{FB:theoremForm}. $\square $

\begin{myremark}
It is evident  from the proof of Theorem \ref{FB:theoremForm} that the main idea of the control law (\ref{FB:controlLaw}) can be explained as follows. Each robot $i$ is guided towards a fictitious target $T_{i}$ that is always located ahead of the desired robot's position relative to its neighbours. The reason we guide the robot towards a fictitious target but not the desired relative robot's position itself is clear from (\ref{FB:omegaEq}). If we conducted  the robot towards the desired relative position, we would have $\Vert d_{i}(t)\Vert\rightarrow 0$; therefore, $\omega_{i}(t)\rightarrow\infty$ and the constraint (\ref{FB:constrain1}) would be violated. Notice that our method for guidance towards a fictitious target $T_{i}$ is a pure pursuit type guidance law (see, e.g., \cite{savkin2010bearings}).
\end{myremark}

\section{Formation Building with Anonymous Robots}\label{FB:SecAnonyFormation}
In the area of robotics, it is common to use the multi-robot task allocation approach to similar problems. However, most work on multi-robot task allocation has been ad hoc and empirical especially in the case of an arbitrarily large number of robots; see, e.g., \cite{gerkey2004formal, lerman2006analysis}. In this section, we propose a randomized algorithm to handle this problem which leads to a mathematically rigorous theoretical analysis for any number of robots.
In Section \ref{FB:SecFormation}, an algorithm of formation building for a team of mobile robots was proposed in which positions of all robots are pre-assigned, i.e., each robot knows a priori its final position in the desired geometric configuration. In this section, we present the algorithm of formation building with anonymous robots meaning that the robots do not know their final position in the desired geometric configuration at the beginning but using a randomized algorithm, they eventually reach a consensus on their positions. In other words, each robot does not know a priori its position in the configuration $\mathcal{C}=\{X_{1},\ X_{2},\ \ldots,\ X_{n},\ Y_{1},\ Y_{2},\ \ldots,\ Y_{n}\}$.

\begin{mydef}\label{FB:navLawGlobalStabilizig}
A navigation law is said to be globally stabilizing with anonymous robots and the configuration $\mathcal{C}= \{X_{1},\ X_{2},\ \ldots,\ X_{n},\ Y_{1},\ Y_{2},\ \ldots,\ Y_{n}\}$ if for any initial conditions $(x_{i}(0),\ y_{i}(0),\ \theta_{i}(0))$, there exists a permutation $r(i)$ of the index set $\{$1, 2, $\ldots,\ n\}$ such that for any $i=1,2,\ \ldots,\ n$, there exist a Cartesian coordinate system and $\tilde{v}_{0}$ such that the solution of the closed-loop system (\ref{FB:kinematicEq}) with the proposed navigation law in this Cartesian coordinate system satisfies (\ref{FB:limit2}) and

\begin{equation}\label{FB:psiDotZero}
\begin{split}
\lim_{t\rightarrow\infty}(x_{i}(t)-x_{j}(t))=&X_{r(i)}-X_{r(j)}\\
\lim_{t\rightarrow\infty}(y_{i}(t)-y_{j}(t))=&Y_{r(i)}-Y_{r(j)}
\end{split}
\end{equation}

for all $1\leq i\neq j\leq n$.
\end{mydef}

Let $R>0$ be a given constant. We assume that each robot $i$ has the capacity to detect all other robots inside the circle of radius $R$ centred at the current position of robot $i$. Furthermore, let $0<\displaystyle \epsilon<\frac{R}{2}$ be a given constant. For any configuration $\mathcal{C}=\{X_{1},\ X_{2},\ \ldots,\ X_{n},\ Y_{1},\ Y_{2},\ \ldots,\ Y_{n}\}$ introduce a undirected graph $\mathcal{P}$ consisting of $n$ vertices. Vertices $i$ and $j$ of the graph $\mathcal{P}$ are connected by an edge if and only if $\sqrt{(X_{i}-X_{j})^{2}+(Y_{i}-Y_{j})^{2})}\leq R-2\epsilon$. We will need the following assumption.

\begin{myassump}\label{FB:graphPisConnected}
 The graph $\mathcal{P}$ is connected.
 \end{myassump}

We present a randomized algorithm to build an index permutation function $r(i)$ . Let $N\geq 1$ be a given integer. Let $r(0,\ i)\in\{1,2,\ \ldots,\ n\}$ be any initial index values where $i=1,2,\ \ldots,\ n$.

As in the navigation law (\ref{FB:consensusVar}), (\ref{FB:controlLaw}), for any time $t$ and any robot $i$, we consider a Cartesian coordinate system with the $x$-axis in the direction $\tilde{\theta}_{i}(t)$ (according to the definition (\ref{FB:continuousTimeVar}), $\tilde{\theta}_{i}(t)$ is piecewise constant). In other words, in this coordinate system $\tilde{\theta}_{i}(t)=0,\ x_{i}(t)$, $y_{i}(t)$ are now coordinates of robot $i$ in this system. Furthermore, we say that a vertex $j$ of the graph $\mathcal{P}$ is vacant at time $kN$ for robot $i$ if there is no any robot inside the circle of radius $\epsilon$ centred at the point\\

$\left(\begin{array}{l}
(x_{i}(kN)+\tilde{x}_{i}(kN))+X_{j}+kN\tilde{v}_{i}(kN)\\
(y_{i}(kN)+\tilde{y}_{i}(kN))+Y_{j}
\end{array}\right)$\\

Let $S(kN,\ i)$ denote the set of vertices of $\mathcal{P}$ consisting of $r(kN,\ i)$ and those of vertices of $\mathcal{P}$ that are connected to $r(kN,\ i)$ and vacant at time $kN$ for robot $i$. Let $|S(kN,\ i)|$ be the number of elements in $S(kN,\ i)$. It is clear that 1 $\leq |S(kN,\ i)|$ because $r(kN,\ i) \in S(kN,\ i)$. Moreover, introduce the Boolean variable $b_{i}(kN)$ such that $b_{i}(kN) :=1$ if there exists another robot $j\neq i$ that is inside of the circle of radius $\epsilon$ centred at\\

$\left(\begin{array}{ll}
(x_{i}(kN)+\tilde{x}_{i}(kN))+X_{i}+ & kN\tilde{v}_{i}(kN)\\
(y_{i}(kN)+\tilde{y}_{i}(kN))+Y_{i} &
\end{array}\right)$\\

at time $kN$, and $b_{i}(kN) :=0$ otherwise. We propose the following random algorithm:

\begin{equation}\label{FB:randAlg}
r((k+1)N,\ i)=\left\{\begin{array}{ll}
r(kN,\ i) & \text{if  $(b_{i}(kN)=0$ or\ $(b_{i}(kN)=1)$}\\
  & \text{ \ and\  $|S(kN,\ i)|=1$}\\
j & \text{if\ $b_{i}(kN)=1$}\\
  & \text{ and\ $|S(kN,\ i)|>1$}
\end{array}\right.
\end{equation}

Now, we are in a position to present the main result of this section.

\begin{mytheorem}\label{FB:theoremAnonyForm}
 Consider the autonomous robots described by the equations (\ref{FB:kinematicEq}) and the constraints (\ref{FB:constrain1}), (\ref{FB:constrain2}). Let $\mathcal{C}=\{X_{1},\ X_{2},\ \ldots,\ X_{n},\ Y_{1},\ Y_{2},\ \ldots,\ Y_{n}\}$ be a given configuration. Suppose that Assumptions \ref{FB:timeIntervals}, \ref{FB:initialValues}, \ref{FB:initialRobotsSpeeds} and \ref{FB:graphPisConnected} hold, and $c$ is a constant satisfying (\ref{FB:cConstant}). Then, for initial conditions $(x_{i}(0),\ y_{i}(0),\ \theta_{i}(0))$, $i=1,2,\ \ldots,\ n$, there exists an integer $N_{0} > 0$ such that for any $N \geq N_{0}$, the decentralized control law (\ref{FB:consensusVar}), (\ref{FB:controlLaw}), (\ref{FB:randAlg}) with probability 1 is globally stabilizing with these initial conditions and the configuration $\mathcal{C}$.
\end{mytheorem}

\textbf{Proof of Theorem \ref{FB:theoremAnonyForm}:} The algorithm (\ref{FB:randAlg}) defines an absorbing Markov chain which contains a number of absorbing states that are impossible to leave; in this case, the states when different robots correspond to different vertices of the desired configuration. It is also obvious that these absorbing states can be reached from any initial state with a non-zero probability. It is a well-known theorem of the Markov chain theory that a Markov chain with finite number of states has, at least, one absorbing state which can be reached from any other states with non-zero probability. Then, with probability 1, one of the absorbing states will be reached. This completes the proof of Theorem \ref{FB:theoremAnonyForm}. $\square $

\section{Obstacle Avoidance}\label{FB:secObstAvoid}
We consider a more challenging problem of navigation of a group of mobile robots for formation in the existence of obstacles. The map of the environment, information about the obstacles including their shapes, positions and geometric distribution are not known to the robots a priori. To detect an obstacle, the robots must be equipped with a range sensor like sonar or laser. The robots can detect an obstacle when it lies within their range. Then, they obtain range and angle to  the obstacle. The algorithm of obstacle avoidance employs this  information and calculates an appropriate route to avoid collision with  the obstacle.

\begin{figure} [bt]
  \centering
  \includegraphics[trim=1cm 1.2cm 1cm .5cm ,width=7 cm, height=6 cm]{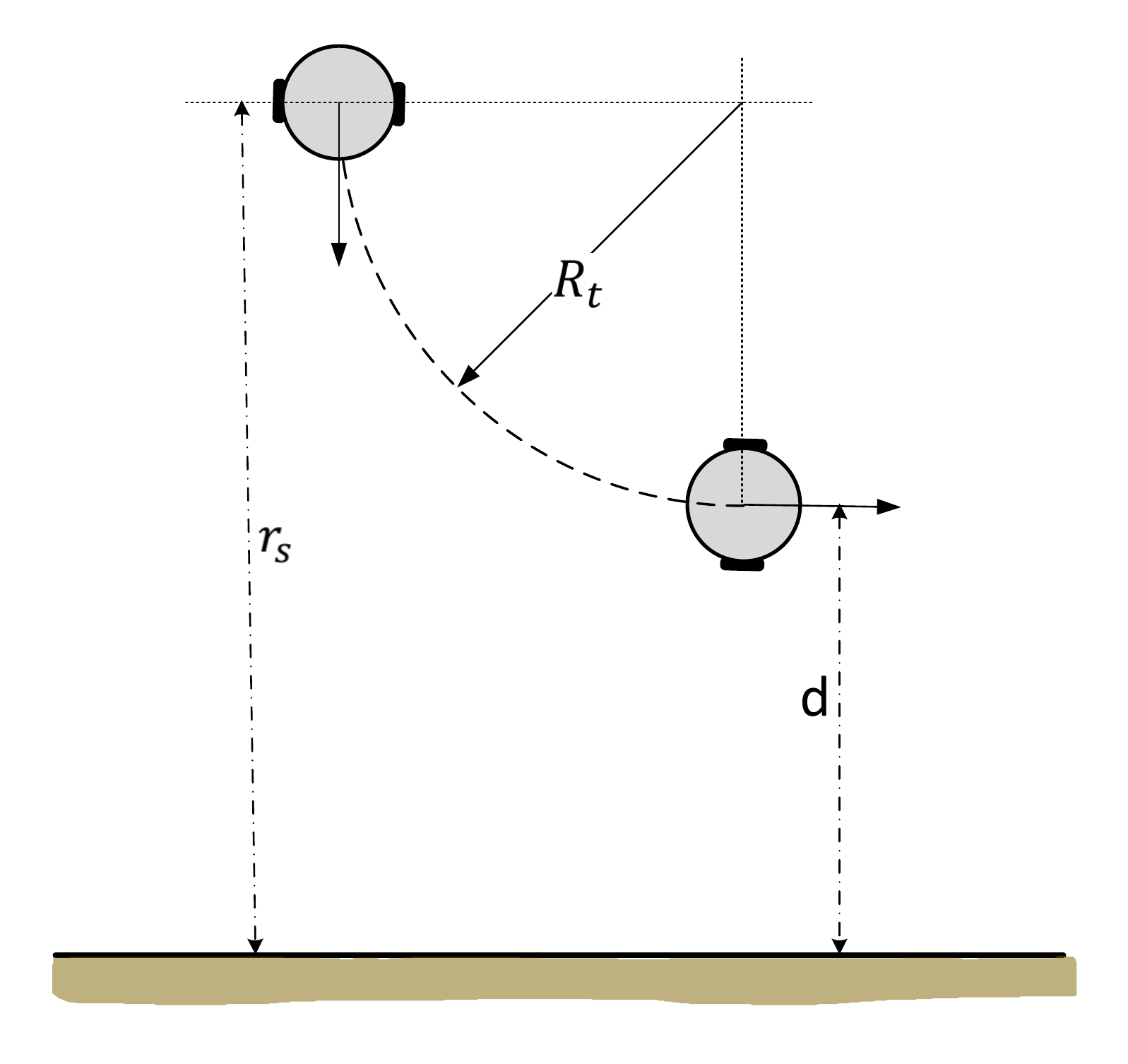}\\
  \caption{Detecting an obstacle}\label{FB:detectingObstacle}
\end{figure}

We apply an algorithm for obstacle avoidance  that uses  angles and distances  provided by range sensors. We assume that the range sensors are located on the robot's  perimeter, in the forepart with $180^\circ$ field of view; $\pm90^\circ$ with respect to robot's heading. Also, we assume that the maximum range of robots' range sensors is $r_s$.
 As shown in Fig. \ref{FB:detectingObstacle}, a robot moving toward an obstacle detects the obstacle as soon as  the obstacle is placed in the sensing range of the robot. Then, the robot  changes its route to turn the obstacle preserving a distance  to  it. Suppose $R_t$ be the turning radius of the robot and $d$ be the distance to the obstacle when the robot's heading is parallel to the obstacle surface. Since

 \begin{equation*}\label{FB:}
R_t^{max}=\frac{V^M}{\omega^{min}}
 \end{equation*}
and
\begin{equation*}\label{FB:}
d_{min}=r_s-R_t^{max}
 \end{equation*}

 thus,  we need following assumption.

  \begin{myassump}\label{FB:minDist}
$d_{min}>d_0$ where $d_{0}$ is a given constant.
\end{myassump}

\begin{figure} [bt]
  \centering
  \includegraphics[trim=3cm 13cm 3cm 0cm ,width=8 cm, height=7 cm]{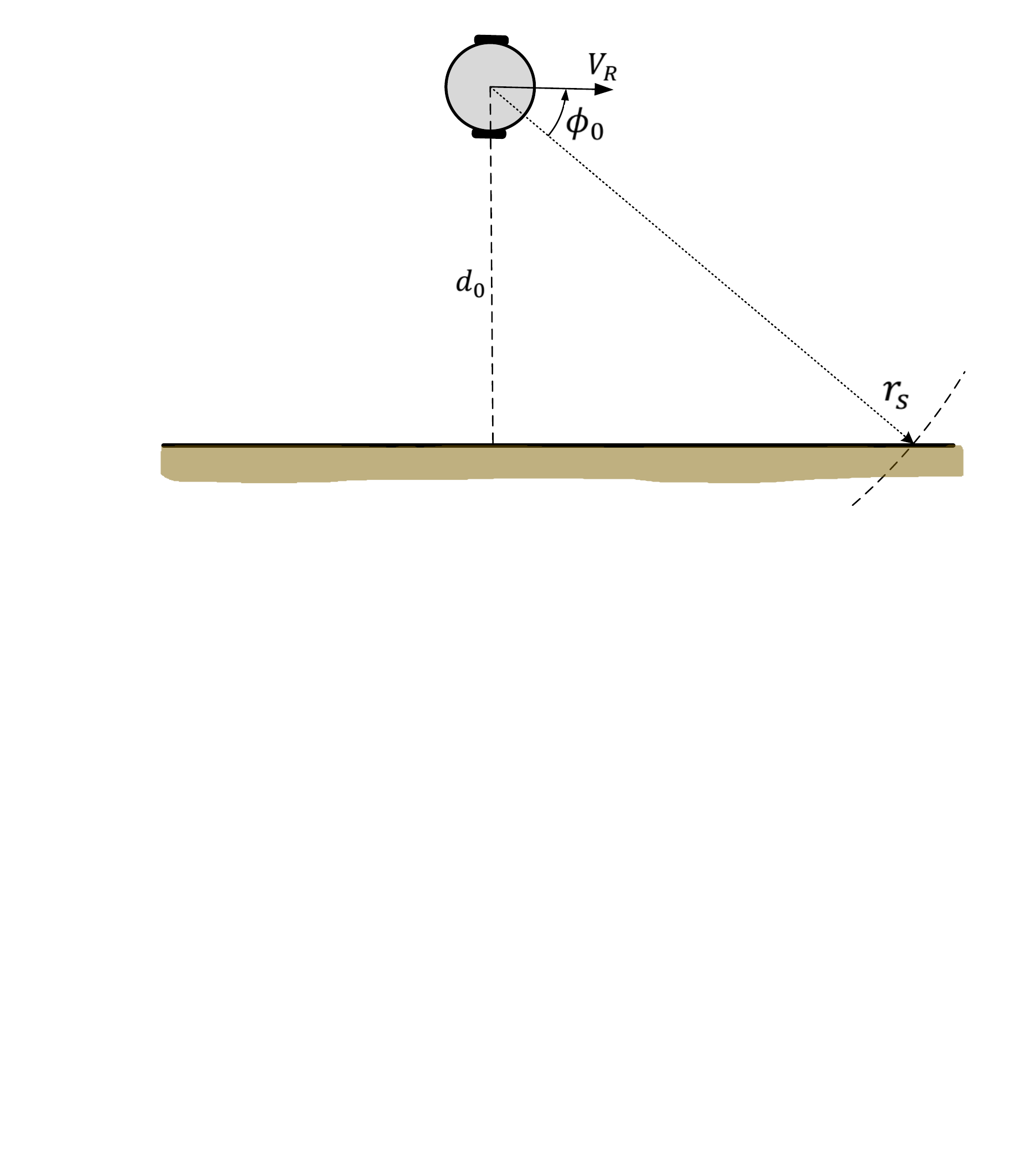}\\
  \caption{Moving with a constant distance to the obstacle }\label{FB:anglePhi0}
\end{figure}

Assume a robot is moving along the circumference of an obstacle (see Fig. \ref{FB:anglePhi0}). Also, suppose that the curvature radius of the obstacle is big enough such that the surface of the obstacle is assumed flat. As shown in Fig. \ref{FB:anglePhi0}, if the robot picks the farthest detectable point on the obstacle surface using its range sensor as a  reference point,  there exists an angle between the robot's heading and  range sensor's ray is termed as avoiding angle. To have a constant distance to the obstacle, we need a constant avoiding  angle $\phi_0$   satisfying  $d_0=r_s\sin\phi_0$.

\begin{figure} [bt]
  \centering
  \includegraphics[trim=3cm 11cm 3cm 0cm ,width=9 cm, height=8 cm]{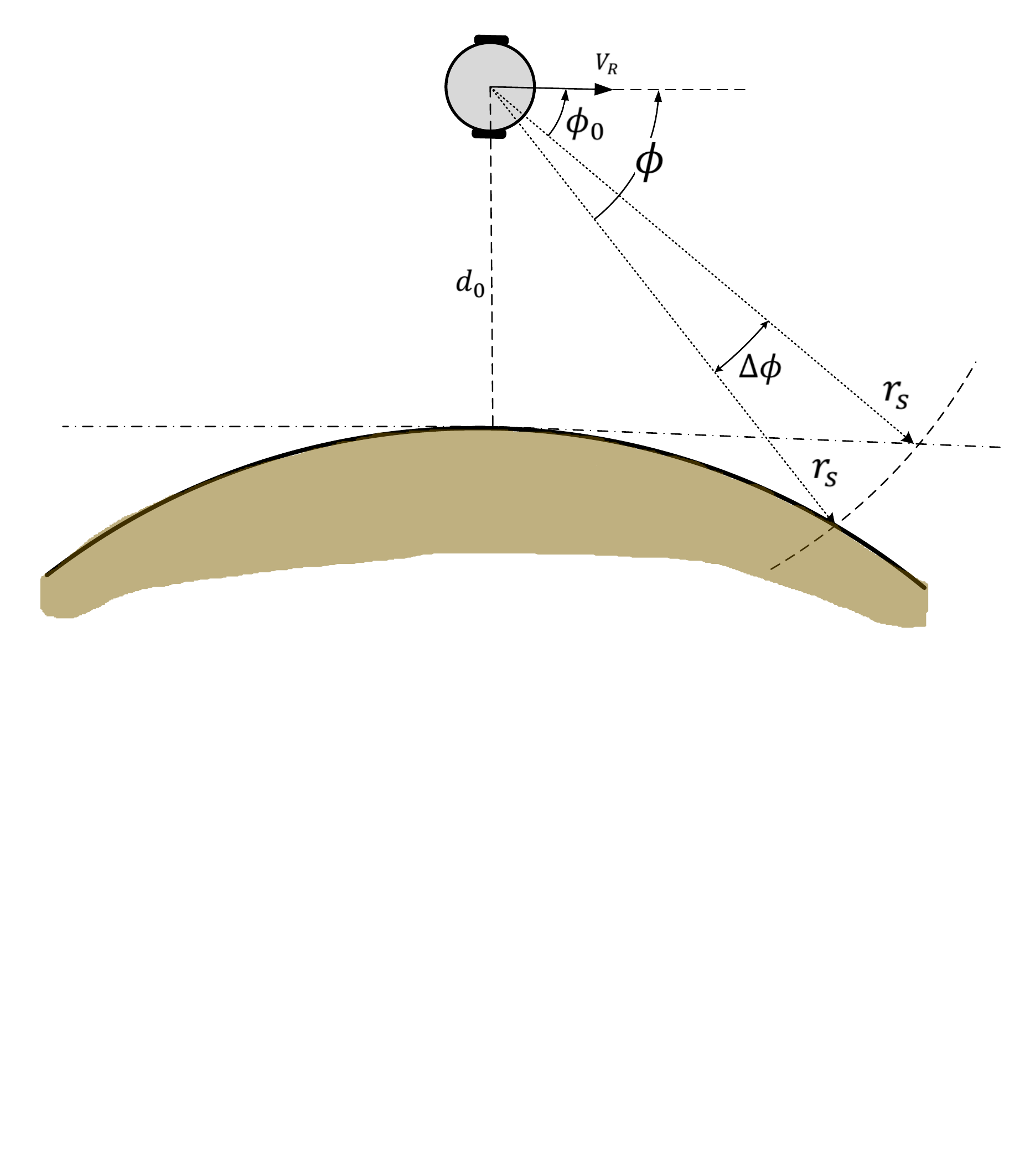}\\
  \caption{Detecting the curvature of an obstacle}\label{FB:curvedObstacle}
\end{figure}

Now, consider  the robot encounters a curved obstacle (see Fig. \ref{FB:curvedObstacle}). Therefore, the robot must follow a trajectory preserving the given distance of $d_0$ to the obstacle surface. For instance, as shown in Fig. \ref{FB:curvedObstacle}, the robot's distance to the obstacle is $d_0$ but the range sensor detects that the avoiding angle $\phi$ is greater than $\phi_0$ and their difference is $\Delta\phi=|\phi-\phi_0|$. Thus, the robot must turn in order to remove this gap; by turning equal to  $\Delta\phi$ to the right in this case.

\begin{figure} [bt]
\centering
\includegraphics[trim=4cm 10cm 4cm 0cm ,width=10 cm, height=10 cm]{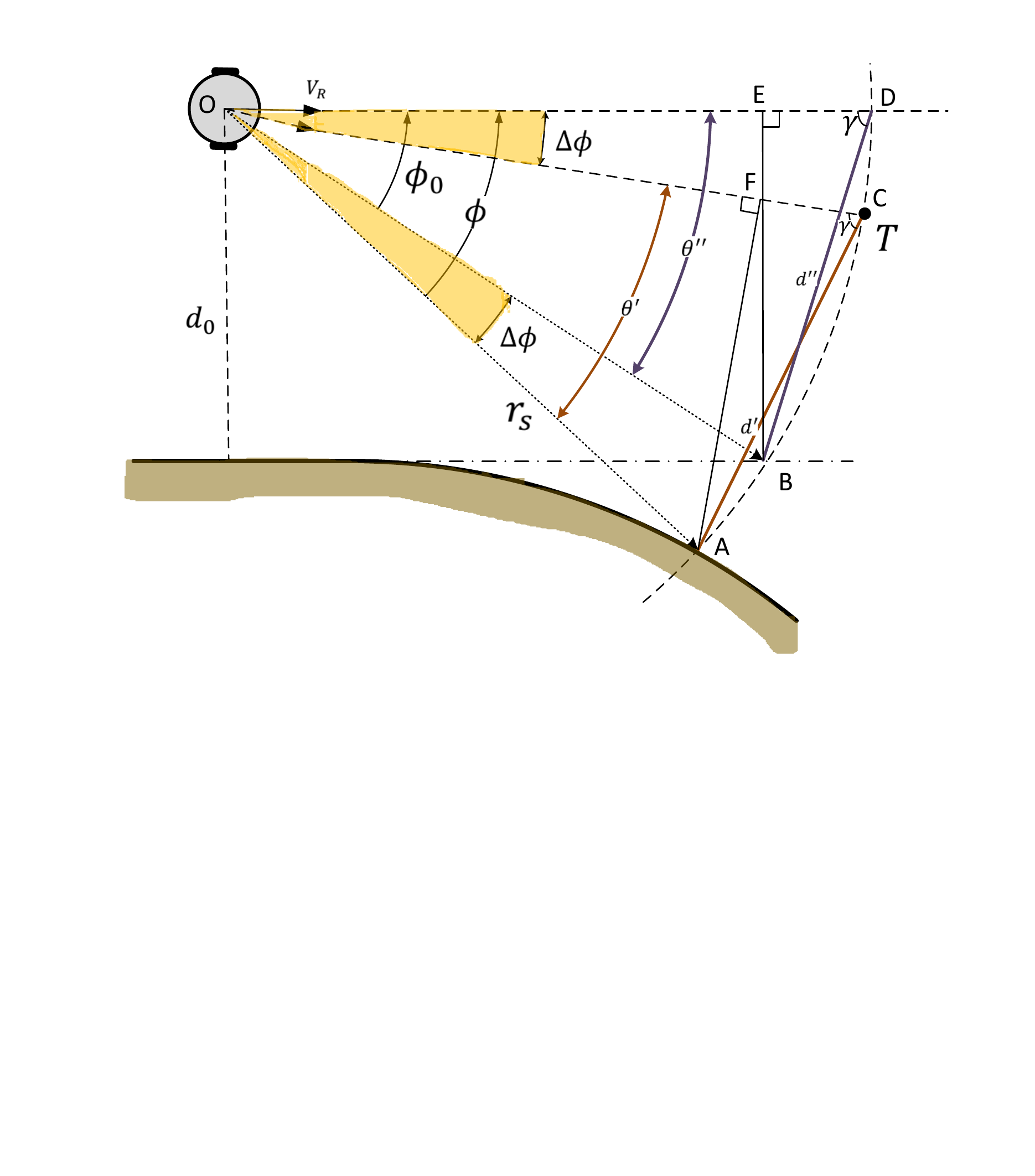}\\
\caption{A convex obstacle}\label{FB:posCurvature}
\end{figure}

Fig. \ref{FB:posCurvature} shows the details of obstacle avoidance approach when the obstacle is convex. As shown in Fig. \ref{FB:posCurvature}, the robot is moving along the surface of the obstacle with avoiding distance of $d_0$. As previously mentioned and shown in Fig. \ref{FB:anglePhi0}, there is an angle of $\phi_0$ between the robot's heading and  the range sensor's ray for a flat surface. However,  in order to keep moving with the avoiding distance of  $d_0$, the robot must turn by $\Delta\phi$ toward the obstacle. Assume a fictitious  target $T$, a point with a distance of  $r_s$ to the robot and angle of $\Delta\phi$ respect to  the robot's heading toward the obstacle (see Fig .\ref{FB:posCurvature}). As depicted in  Fig. \ref{FB:posCurvature},  angles $\theta'$ and $\theta''$ are equal thus the line segments $d'$ and $d''$ will be equal. In addition, since $\widehat{AB}=\widehat{CD}$ thus $\measuredangle{ODB}=\measuredangle{OCD}=\gamma$ which satisfies that triangles BED and AFC are equal. Therefore, line segment AF, which is the distance to the obstacle at F, will be equal to BE$=d_{0}$. It means that if point C is selected as the fictitious target, the distance to the obstacle will be maintained  to a given constant.

\begin{figure} [bt]
\centering
\includegraphics[trim=4cm 9cm 4cm 0cm ,width=10 cm, height=10 cm]{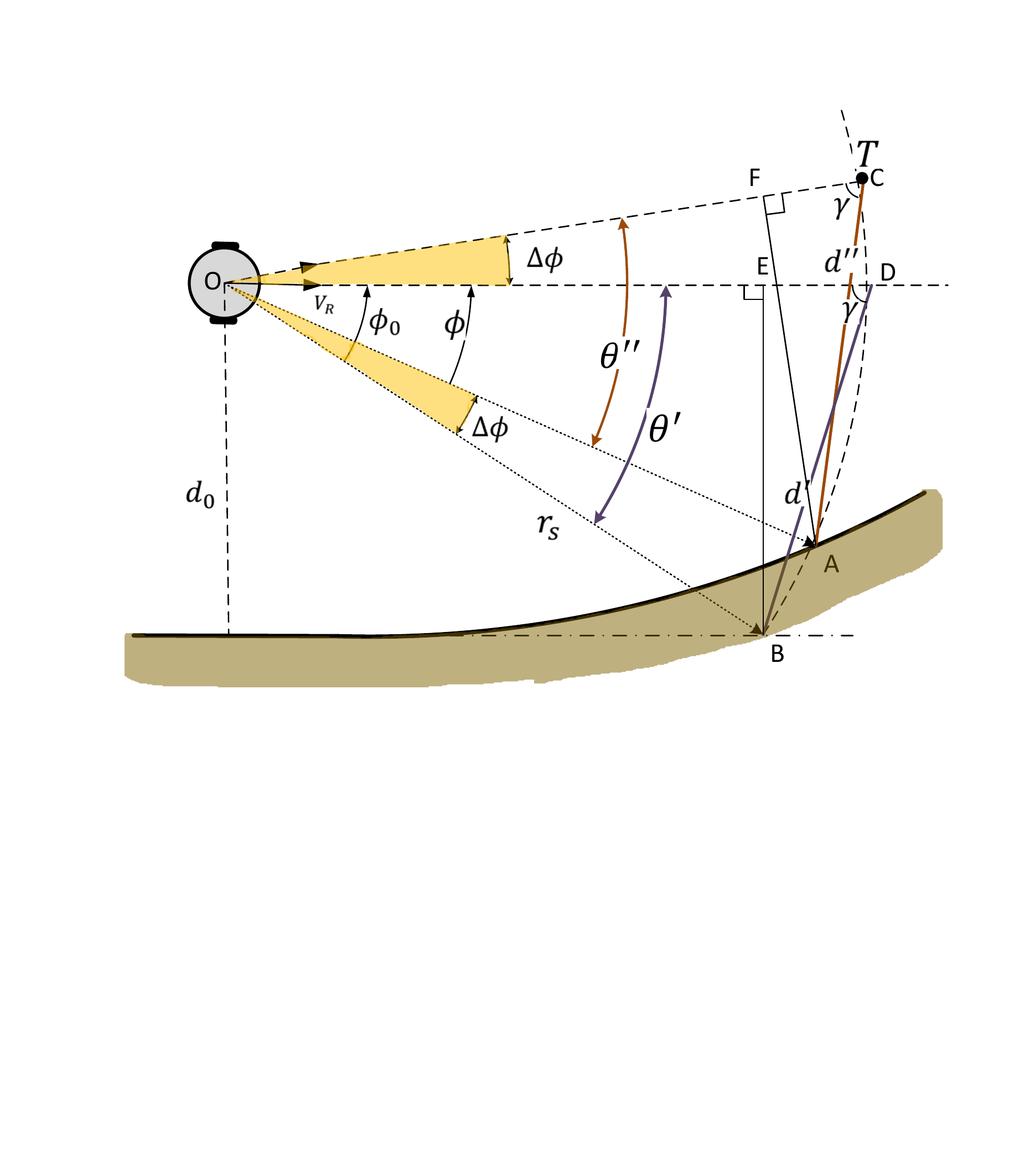}\\
\caption{A concave obstacle}\label{FB:negCurvature}
\end{figure}

Fig. \ref{FB:negCurvature} shows the case that the obstacle is concave. This case is  similar to the convex case except the fictitious target that is away from the obstacle; therefore, the robot must turn by $\Delta\phi$ away from the obstacle.

As a result, we propose the following control law that enables robots to avoid a collision by calculating  a smooth path around the obstacles.

\begin{align}\label{FB:controlLawObstacleAvoid}
\begin{split}
&v_{i}(t)=V^M  \\
& \omega_{i}(t)=\omega^{max}sign(\psi_{i}(t))
\end{split}
\end{align}

for all $i=1,2,\ldots,n$, where
\begin{equation}\label{FB:psi}
\small
\psi_{i}(t)=\left\{\begin{array}{ll}
1 & \text{If  $\phi<\phi_{0}$ }\\
 0 & \text{If $\phi=\phi_{0}$}\\
-1 & \text{If  $\phi>\phi_{0}$}
\end{array}\right.
\end{equation}
also $V^M $, $\omega^{max}$ and $sign(.)$ are given in (\ref{FB:constrain1}), (\ref{FB:constrain2}) and (\ref{FB:sign}), respectively.

Now, we are in a position to present the main results of this chapter.

\begin{mytheorem}\label{FB:theoremFormObst}
Consider the autonomous mobile robots described by the equations (\ref{FB:kinematicEq}) and the constraints (\ref{FB:constrain1}), (\ref{FB:constrain2}). Let $\mathcal{C}=\{X_{1},\ X_{2},\ \ldots,\ X_{n},\ Y_{1},\ Y_{2},\ \ldots,\ Y_{n}\}$ be a given configuration. Suppose that Assumptions \ref{FB:timeIntervals}, \ref{FB:initialValues}, \ref{FB:c and C} and \ref{FB:initialRobotsSpeeds} hold, and $c$ is a constant satisfying (\ref{FB:cConstant}). Then, the distributed control law (\ref{FB:consensusVar}), (\ref{FB:controlLaw}), (\ref{FB:controlLawObstacleAvoid}) is globally stabilizing with any initial conditions and the configuration $\mathcal{C}$.
\end{mytheorem}

{\bf Proof of Theorem \ref{FB:theoremFormObst}:} proof of Theorem \ref{FB:theoremFormObst} is completely similar to the proof of Theorem \ref{FB:theoremForm}. Both control laws, (\ref{FB:controlLaw}) for formation building  and (\ref{FB:controlLawObstacleAvoid}) for obstacle avoidance  are the same. The main difference is that the fictitious target ${T}$  in this case is variable between (\ref{FB:gVector}) and what is defined in this section. In other words,  whenever a  robot encounters an obstacle, the fictitious target switches from (\ref{FB:gVector}) to a point with a distance of  $r_s$ to the robot and angle of $\Delta\phi$ respect to  the robot's heading toward the obstacle (point C in Fig. \ref{FB:posCurvature} and Fig.  \ref{FB:negCurvature}).

\section{Simulation Results}

 We present computer simulation results for all algorithms proposed in this chapter: obstacle avoidance, formation building with obstacle avoidance and anonymous formation building with obstacle avoidance. We use Mobotsim 1.0 simulator, a powerful 2D simulator of mobile robots that simulates robots' motion, environment and range sensors like sonar. Simulation parameters are given in Table \ref{FB:simulationParameters}.

\begin{table}[!hbt]
\centering
\caption{Simulation Paremeters}\label{FB:simulationParameters}
\resizebox{12cm}{!} {
\begin{tabular}{|l|l|l|}
\hline
\textbf{Parameter}                & \textbf{Value} & \textbf{Comment} \\ \hline
Sampling Intervals        & 0.1   &     s    \\ \hline
Robot's Platform Diameter & 0.5   & meter   \\ \hline
Distance Between Wheels   & 0.35  & meter   \\ \hline
Wheels Diameter           & 0.2   & meter   \\ \hline
Maximum Angular Velocity  & 2     & rad/s   \\ \hline
Maximum Linear Velocity   & 1.5   & m/s     \\ \hline
Minimum Linear Velocity   & .2    & m/s     \\ \hline
Sonar’s Maximum  Range    & 2     & meter   \\ \hline
Number of Ranging Sonars  & 12    &         \\ \hline
Sonars' Radiation Cone    & 15    & degree  \\ \hline
\end{tabular}
}
\end{table}

\subsection{Obstacle Avoidance}
\begin{figure}[!htb]
\centering
\subfigure[]{\includegraphics[width=6 cm, height=5.5 cm]{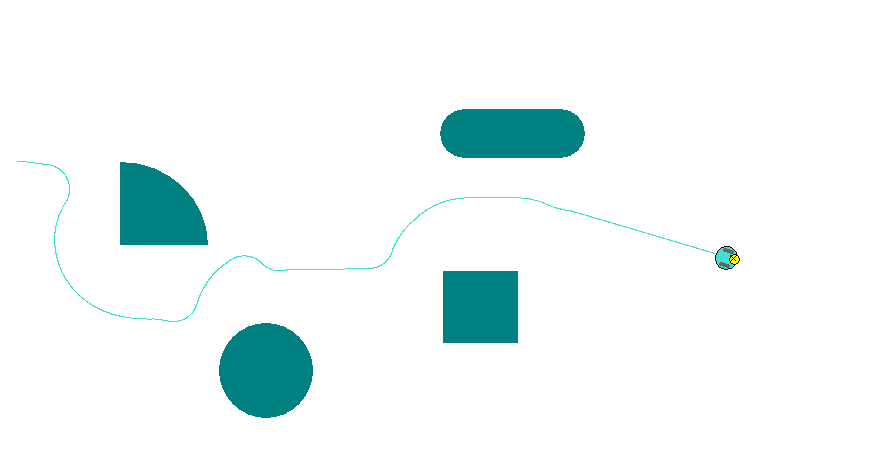}}\quad
\subfigure[]{\includegraphics[width=6 cm, height=5.5 cm]{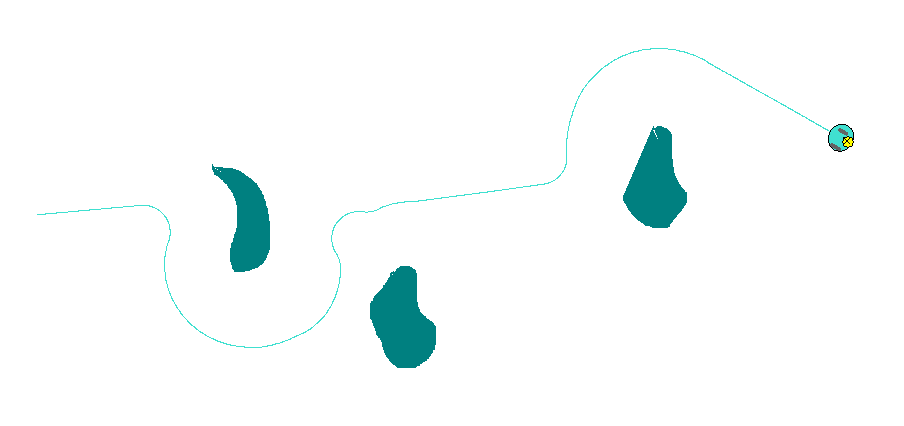}}\quad
\subfigure[]{\includegraphics[width=6 cm, height=5.5 cm]{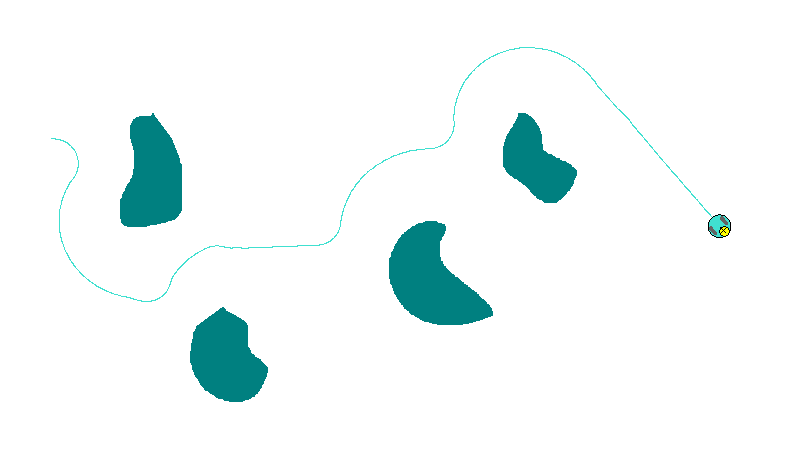}}\quad
\subfigure[]{\includegraphics[width=6 cm, height=5.5 cm]{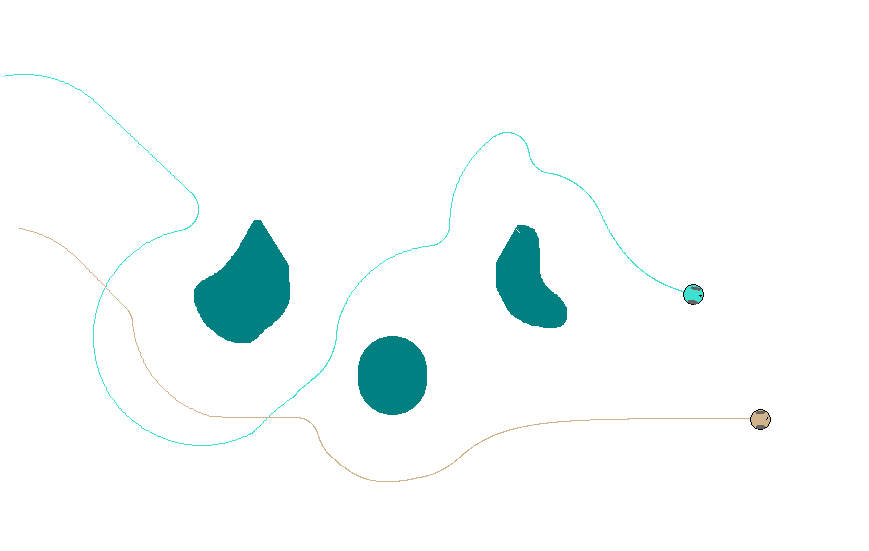}}
\caption{Applying the obstacle avoidance rule; robots pass the obstacles}
 \label{FB:obstAvoid}
\end{figure}

First, we present the simulation results for the proposed obstacle avoidance rule (\ref{FB:controlLawObstacleAvoid}). Fig. \ref{FB:obstAvoid} shows the simulation results for obstacle avoidance rule (\ref{FB:controlLawObstacleAvoid}) with some different obstacles. As Fig. \ref{FB:obstAvoid} displays, by applying the proposed obstacle avoidance rule, the robots successfully bypass the obstacles with different shapes and sizes.

\subsection{Formation Building with Obstacle Avoidance}\label{FB:simFormObstAvoid}
\begin{figure} [!bt]
  \centering
  \includegraphics[ width=8 cm, height=7 cm]{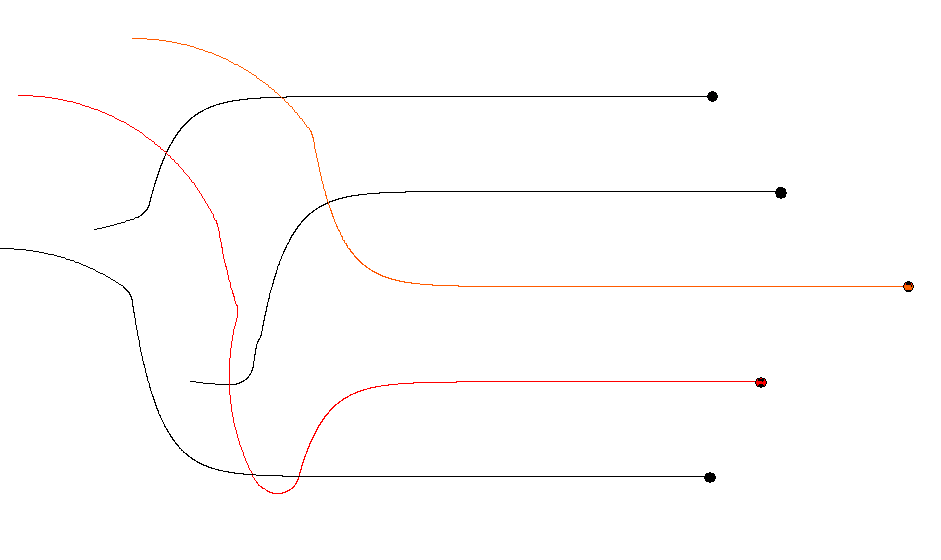}\\
  \caption{Robots form the desired pattern without any obstacles}\label{FB:formNoObst1}
\end{figure}

To simulate the algorithm of formation building with obstacle avoidance, we consider a team consisting of five robots  randomly located on the plane with different headings. The goal is to build a formation as well as  avoiding  the  obstacles that might obstruct robots'  movement.  The robots are to build the edge ( '$>$') by applying the proposed algorithm of formation building in Section \ref{FB:SecFormation} and the obstacle avoidance rule in Section \ref{FB:secObstAvoid}. First, we assume that there is not any obstacle; therefore, only the formation building rule of the proposed algorithm is used. As depicted in Fig. \ref{FB:formNoObst1}, the robots build the desired formation  ('$>$').

\begin{figure}[!tbp] 
\centering
\subfigure[]{\includegraphics[width=7 cm, height=5.5 cm]{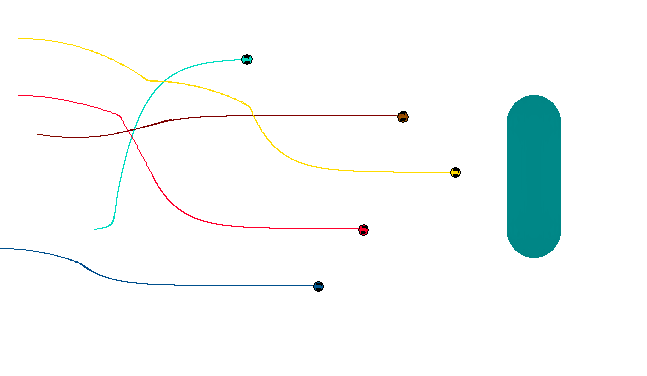}}\quad
\subfigure[]{\includegraphics[width=7 cm, height=5.5 cm]{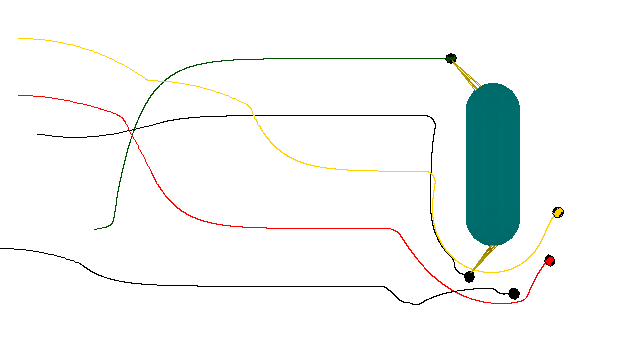}}\quad
\subfigure[]{\includegraphics[width=8 cm, height=5.5 cm]{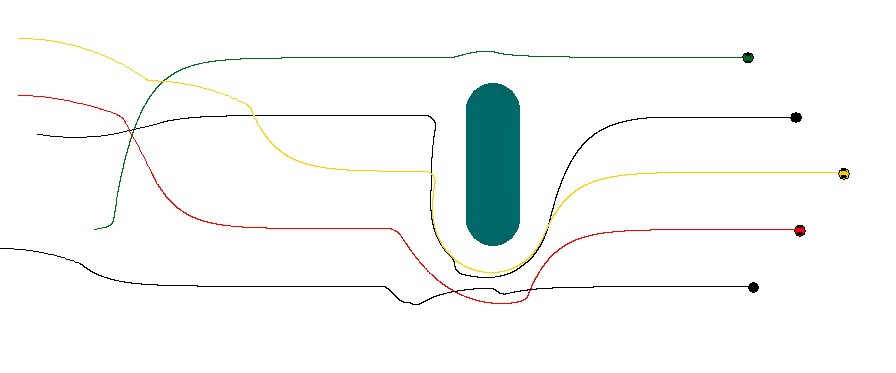}}
\caption{Robots pass the obstacle and build the desired form}
 \label{FB:form1Obst}
\end{figure}

\begin{figure} [!btp]
  \centering
  \subfigure[]{\includegraphics[   width=10 cm, height=5.5 cm  ]{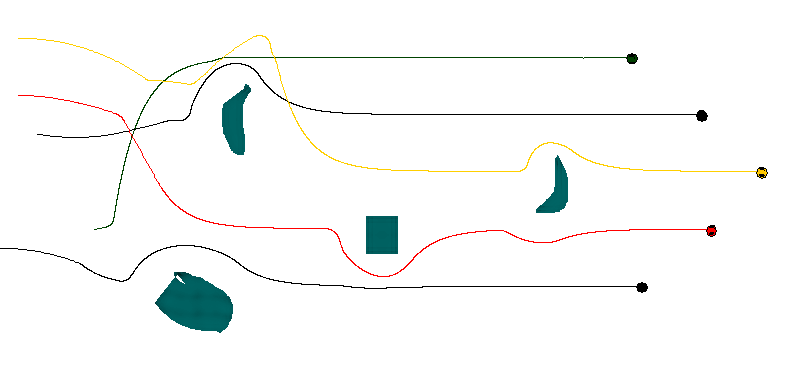}}\\
  \subfigure[]{\includegraphics[   width=10 cm, height=5.5 cm  ]{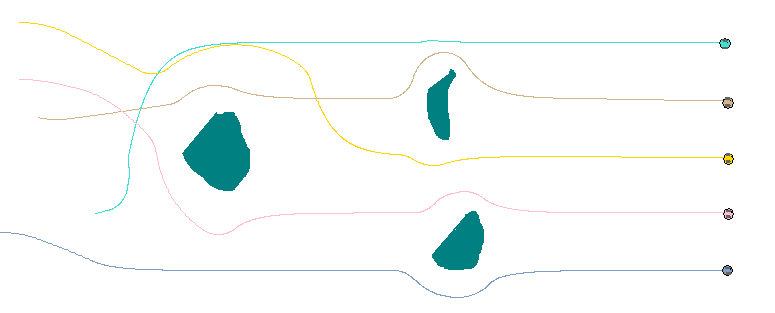}}\\
    \subfigure[]{\includegraphics[   width=10 cm, height=5.5 cm  ]{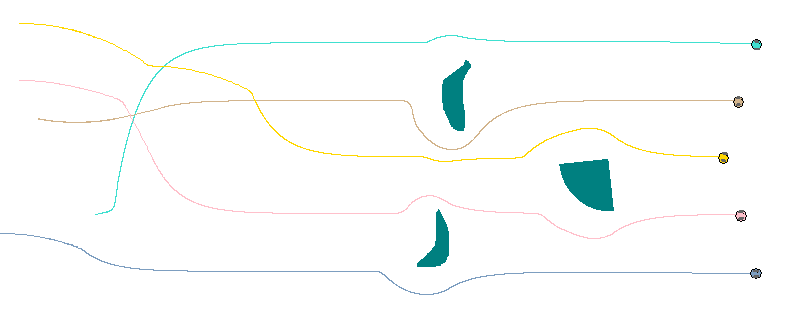}}\\
    \caption{Robots form the desired patterns and avoid  obstacles}\label{FB:form2Obst}
\end{figure}

 Second, we assume the same problem but this time with an obstacle. The simulation results of applying the proposed  algorithm are displayed in Fig. \ref{FB:form1Obst}. As shown in Fig. \ref{FB:form1Obst}(a), the robots build the desired formation  before they encounter the obstacle, and move  such that the formation configuration holds. When the robots detect an obstacle on their direction, they  avoid the obstacle by turning around. Fig. \ref{FB:form1Obst}(b) shows the snapshot of this phase. Passing the obstacle, the robots restart the formation building phase and as  Fig. \ref{FB:form1Obst}(c) shows, the desired formation is built again.  Note that as shown in Fig. \ref{FB:form1Obst}, it is not necessary for the robots to pass the obstacle all together and then start the formation building, e.g., while one robot is still in the obstacle avoidance phase, the other robots that have passed the obstacle begin  the formation building phase again.

To confirm that the proposed algorithm is effective even with any number of obstacles with different shapes and sizes, more simulations are fulfilled.
Fig. \ref{FB:form2Obst} shows the results of these simulations in which  more obstacles with different shapes and sizes are used, and the robots build various formations. In Fig. \ref{FB:form2Obst}(a), the robots build an edge ('$>$')  while avoiding the obstacles on their routes. In Fig. \ref{FB:form2Obst} (b) and (c), they form shapes of a line and an arc, respectively. The results confirm that the proposed algorithm is effective even with any number of obstacles with different shapes and sizes. It should be pointed out that the proposed obstacle avoidance algorithm prevents the collision between robots too; as a robot considers another robot in its sensing range as an obstacle.

\subsection{Formation Building with Anonymous Robots and Obstacle Avoidance}
\begin{figure} [!hbtp]
  \centering
 \mbox{ \subfigure[]{\includegraphics[   width=5 cm, height=5.5 cm  ]{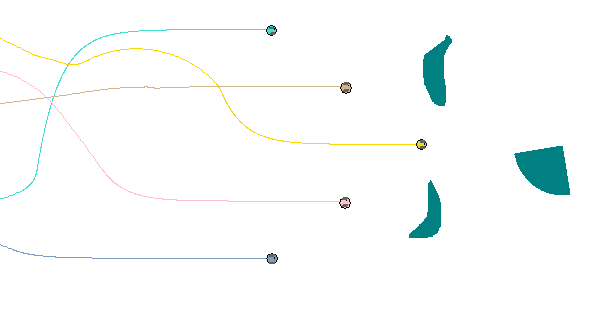}}
  \subfigure[]{\includegraphics[   width=8 cm, height=5.5 cm  ]{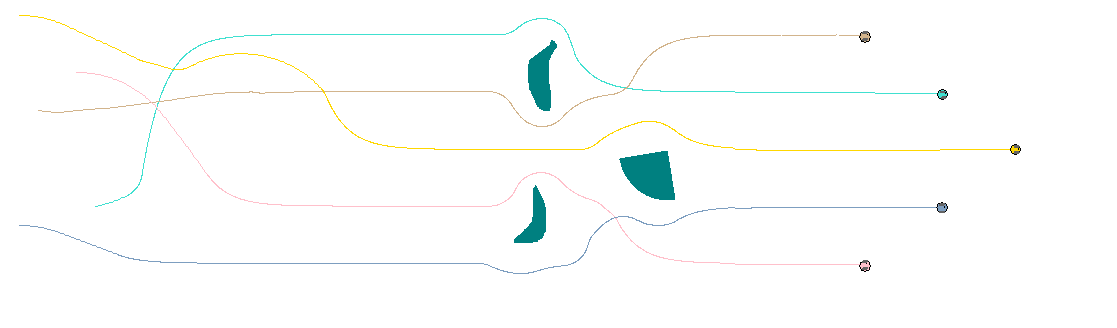}}}\\
  \mbox{ \subfigure[]{\includegraphics[   width=5 cm, height=5.5 cm  ]{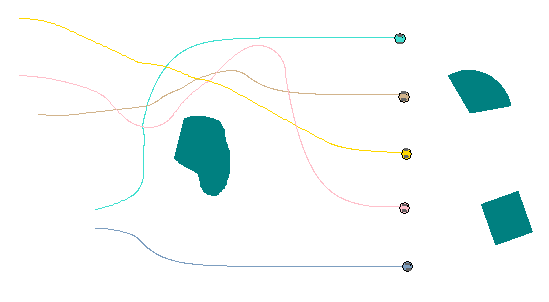}}
  \subfigure[]{\includegraphics[   width=8 cm, height=5.5 cm  ]{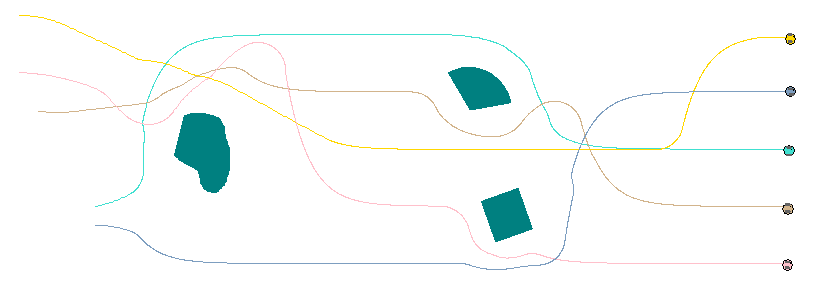}}}\\
  \mbox{ \subfigure[]{\includegraphics[   width=5 cm, height=5.5 cm  ]{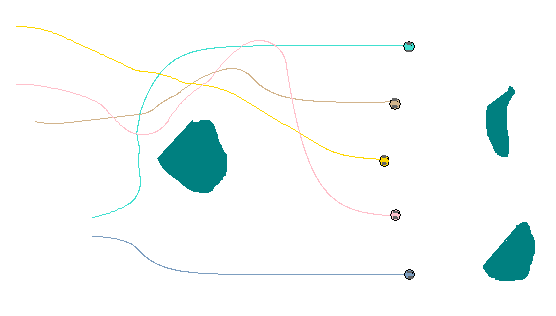}}
  \subfigure[]{\includegraphics[   width=8 cm, height=5.5 cm  ]{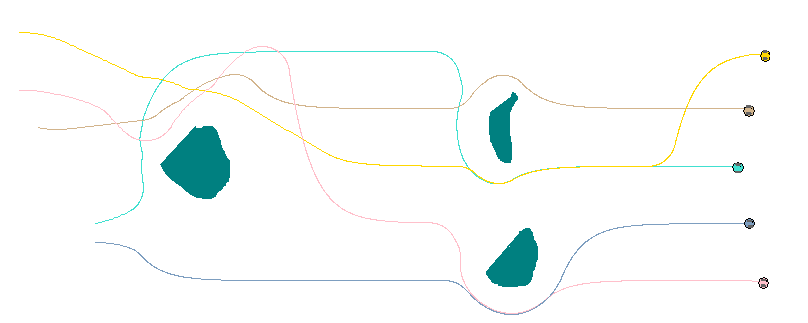}}}
    \caption{Robots form (a),(b) an edge; (c),(d) a line and (e),(f) an arc}\label{FB:formAnonObst1}
\end{figure}

\begin{figure} [!hbt]
  \centering
  \subfigure[Formation building without anonymous robots]{\includegraphics[   width=12 cm, height=6 cm  ]{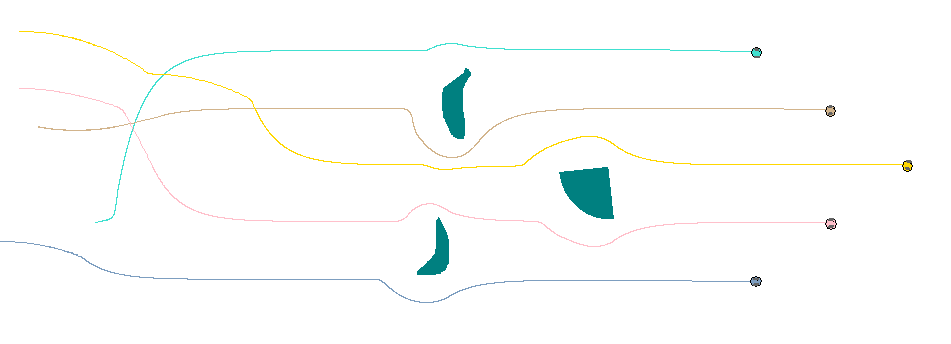}}\\
  \subfigure[Formation building with anonymous robots]{\includegraphics[   width=12 cm, height=6 cm  ]{Formation/FBformAnonEdge2.png}}\\
    \caption{A comparison between formation building  with and without anonymous robots}\label{FB:formAnonObst2}
\end{figure}

In Section \ref{FB:simFormObstAvoid}, simulation results for formation building with obstacle avoidance have been presented. As previously explained, in the algorithm of Section \ref{FB:SecFormation}, positions of all the robots are pre-assigned, i.e., each robot knows its position in the final formation building. Therefore, the positions of the robots do not change during the formation process; as it has been shown in Fig. \ref{FB:form2Obst} when positions of the robots are invariant before and after passing the obstacles. On the other hand, applying the algorithm of Section \ref{FB:SecAnonyFormation}, result in altering the position of the robots during the formation process.
Simulation results for the algorithm of formation building with anonymous robots presented in Section \ref{FB:SecAnonyFormation}, are given in Fig. \ref{FB:formAnonObst1}. In Fig. \ref{FB:formAnonObst1}(a), robots form an edge ('$>$') before encountering the obstacles, and  Fig. \ref{FB:formAnonObst1}(b) shows the positions of the robots after passing the obstacles. As shown in Fig. \ref{FB:formAnonObst1}(b), the positions of the robots after passing the obstacles are not the same as before; robots make the same formation building but with a different arrangement of the robots. Fig. \ref{FB:formAnonObst1}(c-f), show something like that for line '$|$' and arc '(' formation buildings. Fig. \ref{FB:formAnonObst2} shows a comparison between the formation building algorithms, with and without anonymous robots. As depicted in Fig. \ref{FB:formAnonObst2}(a), the positions of the robots in the formation building are the same before and after passing the obstacles while in Fig. \ref{FB:formAnonObst2}(b) where the algorithm of anonymous formation building is applied, the positions of the robots change.

\section{Summary}
The problem of formation building with obstacle avoidance for a team of mobile robots have been considered. The algorithm of global formation building has been  combined with a local obstacle avoidance algorithm. We have proposed a distributed motion coordination control algorithm  so that the robots collectively move in a desired geometric pattern from any initial position while  avoiding the obstacles on their way. We have considered unicycles with    standard kinematic equations and  hard constraints on their linear and angular velocities for the type of the robots. A consensus variables rule has been used for the formation building phase that is  based on the local information. Also, a novel technique based on the information from the range sensors have been employed for the obstacle avoidance phase. Furthermore, we propose a randomized algorithm for the anonymous robots which achieves the convergence to the desired configuration with probability 1.  Mathematically rigorous proofs of the proposed control algorithms have been given, and the  effectiveness of the algorithms have been confirmed  via computer simulations.

\chapter{Conclusions}\label{chap:Conclusions}

The main purpose of this report was to design some algorithms for search by multi-robot systems. We assumed an unknown area including some obstacles  to be searched by a team of autonomous mobile robots either partially for a given number of targets or entirely for the unknown number of targets. For that purpose,  we developed  three decentralized control algorithms to drive a multi-robot team to explore unknown environments.  We used a triangular grid pattern and  two-stage algorithms for the control law so that robots move through the vertices of the grid during the search procedure. In the first stage of the proposed search algorithms, using a consensus variables rule, the robots are located on the vertices of a triangular grid. For the second stage of the algorithm, three different scenarios were considered. First, a pure random grid-based algorithm described in Chapter \ref{chap:RandomSearch} was presented, by which the robots randomly move between the vertices of a common triangular grid so that in each step they only move to the one of the closest neighbouring vertices. Note that there are at most six closest neighbouring vertices for each vertex. In the second scenario, presented in Chapter \ref{chap:SemiRandomSearch}, we changed the pure random rule to a semi-random rule. In this case, the robots  still randomly move between the vertices of a common triangular grid so that in each step they move to the one of the closest neighbouring vertices; but,  only to those vertices which have not been  visited by the robots yet. If all the (at most) six neighbouring vertices have been visited already, one of  them will randomly be selected. Finally, a modified algorithm was proposed in Chapter \ref{chap:ModSearch}.  This algorithm does not confine the robots to move only to the closest neighbouring vertices; but, they move to the nearest unvisited vertex anywhere in the search area.

It has been shown that a triangular grid pattern is asymptotically optimal in terms of the minimum number of robots required for the complete coverage of an arbitrary bounded area. That is we employed a triangular grid pattern for the proposed algorithms, i.e., robots certainly go through the vertices of a triangular grid during the search operation. Therefore, using the vertices of a triangular grid coverage guarantees complete search of  the whole area as well as better performance in terms of search time. Furthermore, we presented a new kind of topological map which robots make and share during the search operation. Unlike many other hubristic algorithms in this area, we gave mathematically  rigorous proofs of convergence with probability 1 of the proposed algorithms.

The procedures of this study were approved by computer simulation results using a simulator of real robots and environments. To evaluate the performance of the algorithms,  we presented the experiment results with real Pioneer 3DX mobile robots for one of the algorithms with detailed descriptions and explanations. The results demonstrated the features of the proposed algorithms and their performance with  real systems. Moreover, we compared the  proposed algorithms with each other and also with three other algorithms from other researchers. The comparison showed the strength of our proposed algorithms over the other existing algorithms.

Also, a further study on networked multi-robot formation building algorithms was presented in this report. The problem of formation building for a group of mobile robots was considered. A decentralized formation building with obstacle avoidance algorithm for a group of mobile robots to move in a defined geometric configuration was proposed. Furthermore, we considered a more complicated formation problem with a group of anonymous robots where the robots are not aware of their position in the final configuration and have to reach a consensus during the formation process while avoiding obstacles. We proposed a randomized algorithm for the anonymous robots which achieves the convergence to the desired configuration with probability 1. Moreover, we presented a novel obstacle avoidance rule which was employed in the formation building algorithms. We demonstrated mathematically rigorous proofs of convergence of the presented algorithms. Also, we confirmed the performance and applicability of the proposed algorithms by computer simulation results.\\

\noindent  {\Large \textbf {Future Work}}

\noindent In terms of directions for future research, further work could be  as follows:

\begin{itemize}

\item It is an interesting direction for future research to apply the proposed search algorithms to swarm systems. In that case, it could also be conducted to determine the effectiveness of limited wireless communication and memory resources \cite{saleem2011swarm, dorigo2013swarmanoid}.

\item During the experiments, no visible or severe drift on wheel odometry was observed; but, it can be significant if the number of  vertices of the grid is large. Addressing this problem is a direction for our future work \cite{censi2013simultaneous, lee2010kinematic}.

\item We have assumed static obstacles in the workspace. For a real application, more challenges may appear with moving obstacles in the environment that would be a fruitful area for further work   \cite{matveev2015globally, savkin2015safe}.

\item In the proposed search algorithms, we have assumed static targets. It is recommended that further research be undertaken with moving targets \cite{hollinger2009efficient}.

\item In Chapter \ref{chap:Formation}, we proposed new strategies in formation control with obstacle avoidance of autonomous robots and presented  computer simulation results. It would be interesting to investigate and verify the effectiveness of the proposed algorithms through experiments with real robots.

\item Regarding the proposed formation building algorithms, the future work can be modifying the proposed algorithms so that the formation holds while passing the obstacles \cite{olfati2006flocking}.

\item Another possible area of future research would be to consider the problem of  environmental extremum seeking by multi-robot teams using the algorithms presented in this report for search and formation \cite{matveev2011navigation, ghods2012multiagent, cochran2009source, liu2010stochastic}.

\end{itemize}


\bibliographystyle{IEEEtran}
\bibliography{ref}

\end{document}